\pdfoutput=1

\documentclass[11pt]{article}

\usepackage{ACL2023}

\usepackage{times}
\usepackage{latexsym}

\usepackage[T1]{fontenc}

\usepackage[utf8]{inputenc}

\usepackage{microtype}
\usepackage{multirow}
\usepackage{graphicx}
\usepackage{arydshln}
\usepackage{tablefootnote}
\usepackage{CJKutf8}
\usepackage{xcolor}
\usepackage{soul}
\usepackage{stmaryrd}
\usepackage{enumitem}
\usepackage{multirow}
\usepackage{booktabs} 
\usepackage{colortbl}
\usepackage{xspace}
\usepackage{amsmath}
\usepackage[nodisplayskipstretch]{setspace}
\usepackage{amsfonts}
\usepackage{caption}
\usepackage{subcaption}

\usepackage{inconsolata}

\newcommand{\Sref}[1]{\S\ref{#1}}

\newcommand{\Fref}[1]{Figure~\ref{#1}}

\newcommand{\Tref}[1]{Table~\ref{#1}}
\newcommand{\Aref}[1]{Appendix~\ref{#1}}

\newcommand{\methodname}{\textsc{ValueScope}\xspace}

\usepackage{enumitem}
\usepackage{colortbl}
\newcommand{\graycell}[1]{\cellcolor{gray!10}#1}
\newcommand{\redcell}[1]{\cellcolor{red!10}#1}
\newcommand{\greencell}[1]{\cellcolor{green!10}#1}

\newcommand\revision[1]{{\textcolor{black}{#1}}}

\title{
\textsc{ValueScope}: Unveiling Implicit Norms and Values\\via Return Potential Model of Social Interactions}

\author{Chan Young Park\textsuperscript{*1} \ \ \ \ \ \ \ Shuyue Stella Li\textsuperscript{*1} \ \ \ \ \ \ \ Hayoung Jung\textsuperscript{*1} \\
\textbf{Svitlana Volkova\textsuperscript{2}} \ \ \ \textbf{Tanushree Mitra\textsuperscript{1}} \ \ \ \textbf{David Jurgens\textsuperscript{3}} \ \ \ \textbf{Yulia Tsvetkov\textsuperscript{1}} \\
\textsuperscript{1}University of Washington \  
\textsuperscript{2}Aptima \
\textsuperscript{3}University of Michigan \\
\texttt{\{chanpark, stelli, hjung10\}@cs.washington.edu}
}

\begin{document}
\maketitle
\def\thefootnote{*}\footnotetext{Equal contribution.}\def\thefootnote{\arabic{footnote}}

\begin{abstract}

This study introduces \methodname, a framework leveraging language models to quantify social norms and values within online communities, grounded in social science perspectives on normative structures. We employ \methodname to dissect and analyze linguistic and stylistic expressions across 13 Reddit communities categorized under gender, politics, science, and finance. Our analysis provides a quantitative foundation showing that even closely related communities exhibit remarkably diverse norms. This diversity supports existing theories and adds a new dimension---community preference---to understanding community interactions. \methodname not only delineates differing social norms among communities but also effectively traces their evolution and the influence of significant external events like the U.S.~presidential elections and the emergence of new sub-communities. 
The framework thus highlights the pivotal role of social norms in shaping online interactions, presenting a substantial advance in both the theory and application of social norm studies in digital spaces.\footnote{\url{https://github.com/stellali7/valueScope}}

\end{abstract}

\section{Introduction}
Social norms---the perceived, informal, and mostly unwritten rules that govern acceptable behaviors within a community---are foundational to understanding the dynamics of social interactions and shaping the community’s identity \cite{unicef}. 
Social values, in turn, are the deeper ideals and principles that a community aspires to uphold, guiding the creation and enforcement of these norms \citep{mcclintock1978social}. 
Social norms and values emerge organically through the interplay of behaviors \cite{stanford:2023} and are difficult to grasp without gaining experience of the community firsthand. 
This complexity poses challenges for new users to assimilate \cite{Lampe2014CrowdsourcingCA} and makes it difficult for automatic community moderation systems \citep{park-etal-2021-detecting-community-custom}.

\begin{figure}[t!]
    \centering
    \includegraphics[width=\linewidth]{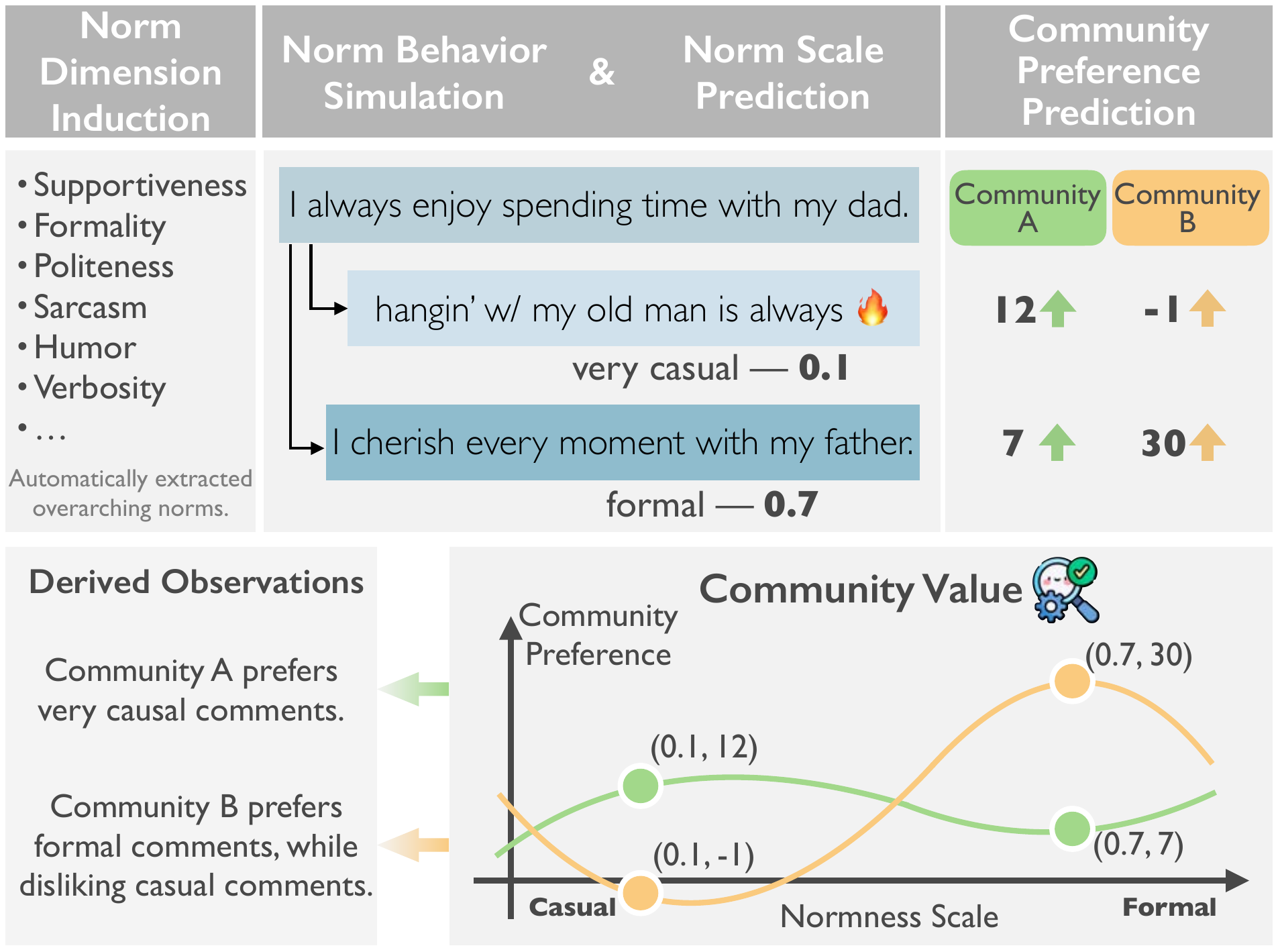}\vspace{-3mm}
    \caption{
    \textbf{The \methodname framework}. 
    We characterize a comment along a norm dimension (e.g., formality), outputting the \emph{normness scale} (e.g., a very casual comment has a formality scale of 0.1). 
    Then, we predict the \emph{return potential}, reflecting community preference (e.g., the number of upvotes).
    Finally, we plot the return potential against the normness scale using the Return Potential Model (RPM) to visualize community values.
    }
    \label{fig:main}\vspace{-5mm}
\end{figure}

Previous studies have focused on a small subset of norms outlined by explicit rules, known as \emph{active norms}, to examine active moderation and governance \citep{fiesler2018reddit, chandrasekharan2018internet, park-etal-2021-detecting-community-custom, neuman2023ai}. However, most social norms remain \emph{implicit}, subtly revealed through social interactions and reinforced by the community, presenting significant challenges for computational modeling. 
Most current methods either rely on qualitative analysis and case studies \citep{shen2022tale, chancellor2018norms, kasunic2018least} or analyze lexical variations, which offer limited explanatory power and generalizability \citep{snoswell2023measuring}. 
Consequently, we ask (RQ1): \emph{How can we identify and measure implicit social norms ingrained in community interactions?} We posit that social norms should not be categorical but understood on a spectrum, reflecting the diversity of human behavior and social groups \citep{jackson1966conceptual}, thereby defining the notion of \emph{normness scale}---the degree of conformity to a norm dimension inspired by \citet{labovitz1973measuring}.

To answer RQ1, we draw inspiration from social science, particularly the \textbf{Return Potential Model} \citep[RPM;][]{jackson1966conceptual}, which views norms as dynamic elements shaped by interactions. We propose a theoretically-grounded computational framework---\textbf{\methodname} (\Fref{fig:main})---to quantify behaviors along social norm dimensions and investigate the interplay of normness scale and community preference to study the formation and evolution of \emph{values}. This leads to our second research question (RQ2): \emph{Can we predict the change in community norms based on observed normative behaviors?} To address this question, we extend \methodname along the temporal axis to capture the shifts in community norms. We examine whether the magnitude and variance of community preferences can help predict future changes in norms.

\methodname offers a scalable framework applicable to diverse online communities and norm dimensions, facilitating large-scale analysis of social norm dynamics. 
Our contributions include:
\begin{enumerate}
[noitemsep,topsep=0pt,leftmargin=12pt]
    \item We introduce \methodname ---a theoretically-grounded framework based on the Return Potential Model (RPM)---to analyze social norms and values within online communities. 
    \item To operationalize the framework, we develop an innovative modeling pipeline consisting of a \textbf{Normness Scale Predictor} to measure the scale of social norms in text and a \textbf{Community Preference Predictor} to quantify community reactions to these variations. We also introduce novel evaluation methods to validate both individual components and the pipeline holistically.
    \item We offer new insights into social dynamics, especially how they evolve over time.
    These findings have important scientific and practical implications for social scientists and community moderators, helping them identify norms that are likely to change and enabling proactive intervention. 
\end{enumerate}

\section{Related Works}
\noindent\textbf{Social Science Literature on Social Norms}\hspace{2mm} 
A \emph{community} represents a collective of individuals united by shared interests \citep{wenger:2015} that develop unique norms, linguistic practices, and identities,  
cultivating specific in-group languages and norms over time \citep{eckert:1989, eckert:1999, eckert:2013, govindarajan2023counterfactual}.
To analyze these norms, \citet{jackson1966conceptual} introduced the Return Potential Model (RPM), viewing social norms as dynamic processes influenced by community members' (dis)approval of behaviors \citep{jackson1975normative}. 
While previous studies have applied RPM through qualitative methods in areas like communication and leadership \citep{glynn2007opinions, nolan:2015, torres1999leadership, henry2004return}, our work diverges as we use computationally analyze implicit norms and values in online communities at scale, focusing on the interplay between community preference and behaviors. 

\noindent\textbf{Norms and Values in Online Communities}\hspace{2mm} 
Computational studies have examined linguistic norms and semantic changes in online communities \cite{lucy-bamman-2021, del-tredici-fernandez-2018-road-custom, 10.1145/2835776.2835784, DanescuNiculescuMizil2013NoCF, 10.1145/2145204.2145254, del-tredici-fernandez-2017-semantic-custom, snoswell2023measuring, chancellor2018norms}. However, these often focus narrowly on language use and neologisms, neglecting the broader spectrum of community values influenced by feedback. \revision{Prior research has utilized Schwartz's Theory of Human Values to estimate values of online communities \cite{van-der-meer-etal-2023-differences, borenstein2024investigatinghumanvaluesonline}. \citet{Weld_Zhang_Althoff_2024} has employed survey methods to create a taxonomy of online community values.}
While some research has addressed explicit governance \cite{chandrasekharan2018internet, fiesler2018reddit, park-etal-2021-detecting-community-custom} or qualitatively studied implicit norms \cite{kasunic2018least, shen2022tale}, our approach fills the gap by (1) focusing on a range of implicit norms (e.g., formality and sarcasm) automatically selected through a generalizable norm induction process, and (2) analyzing collective community preference
over behaviors along the selected norm dimensions to capture a comprehensive spectrum of community values, which can provide a more fine-grained and objective measurement for alignment \cite{bergman2024stela, findeis2024inverse}.

\section{Methodology}

We introduce \methodname---a theoretically-grounded framework to model social norms and values in online communities (\Sref{sec:method:framework}). 
This framework is operationalized through a modeling pipeline consisting of a Normness Scale Predictor (\Sref{sec:method:norm_pred}) and a Community Preference Predictor (\Sref{sec:method:upvote_pred}) to capture two interwoven dimensions of community values. 

\subsection{The \methodname Framework}\label{sec:method:framework}

\noindent\textbf{Theoretical Background}\hspace{2mm}
Community members acquire social adeptness by learning unwritten rules, or implicit norms with feedback from others to guide their behaviors \citep{coutu1951role, zhang2023we}. 
The Return Potential Model \citep[RPM]{jackson1966conceptual} quantifies these norms by mapping the \emph{return potential}---expected (dis)approval---across different behaviors. Individuals in a community adjust their actions based on the learned mental model of return potential. 
We propose \textbf{\methodname}, a computational framework that adapts RPM to analyze the expected community preference to behaviors with varying \emph{normness scales} (i.e., conforming to a norm dimension to different extents), offering scalable insights into community values.

\noindent\textbf{Problem Definition}\hspace{2mm}
Let $\mathcal{C}$ be communities, $\mathcal{A}$ be comments, and $\mathcal{D}$ be norm dimensions (e.g., sarcasm).
For an arbitrary community $c\in\mathcal{C}$ and norm dimension $d\in\mathcal{D}$, \methodname measures the \emph{normness scale} \textbf{$\Phi$} via the Normness Scale Predictor, $\Phi_d:\mathcal{A}\rightarrow\mathbb{R}$, 
and the \emph{community preference} \textbf{$\Psi$} via the Community Preference Predictor, $\Psi_c:\mathcal{A}\rightarrow\mathbb{R}$,
of all $N$ comments in $c$: $\mathcal{A}_c$.
\footnote{Empirically, we perform a distillation step to mitigate confounding factors and distill scores as derived in \Sref{sec:method:norm_pred} and \Sref{sec:method:upvote_pred}---we simply take the delta between two comments $(a_i, a_i')$ to get 
$\nabla\Phi_d:(\mathcal{A}\times\mathcal{A})\rightarrow\mathbb{R}=\Phi_d(a_i')-\Phi_d(a_i)$ and 
$\nabla\Psi_c:(\mathcal{A}\times\mathcal{A})\rightarrow\mathbb{R}=\Psi_c(a_i')-\Psi_c(a_i)$.} 
For an arbitrary range of normness scales $\Phi_d^i:=[\phi_d',\phi_d'')$ (e.g., ``somewhat sarcastic''), we take the set of comments $\mathcal{A}_{c,d}^i:=\{a_i|\Phi_d(a_i)\in\Phi_d^i\}$ with normness scales in the given range, and let $N_{c,d}^i:=||\mathcal{A}_{c,d}^i||$ be the number of comments in this subset. We compute the community preference of these comments:
\vspace{-4mm}{
\setlength{\abovedisplayskip}{0pt}
\setlength{\belowdisplayskip}{0pt}
\setlength{\abovedisplayshortskip}{0pt}
\setlength{\belowdisplayshortskip}{0pt}
\begin{align*}    &\Psi_{c,d}^i:=\Psi_c(\mathcal{A}_{c,d}^i)\\
    &=\{\psi_1,\ldots,\psi_{N_{c,d}^i}|\psi_i=\Psi_c(a_i), a_i\in\mathcal{A}_{c,d}^i\},\vspace{-2mm}
\end{align*}
}
and the estimated community preference of the given normness scale range:
$\widehat{\psi_{c,d}^i} =\frac1{N_{c,d}^i}\sum_{j=1}^{N_{c,d}^i}\psi_j$.
Finally, we obtain $(\Phi_d^i, \widehat{\psi_{c,d}^i})$ as one point on the return potential curve\footnote{Alternatively, $(\nabla\Phi_d^i, \Delta\widehat{\psi_{c,d}^i})$ for the distilled RPM plot.} representing community preferences for comments of varying normness scales. 
For instance, we later show that \texttt{r/askscience} strongly prefers ``very supportive'' comments compared to its spin-off \texttt{r/shittyaskscience} (\Sref{sec:results}).

Differing from the social-science RPM theory, our work proposes \emph{bidirectional continuous normness dimensions} to capture behaviors at both ends of a spectrum, such as identifying both rude and polite comments rather than just measuring politeness. 
This bidirectionality broadens the representational span of our analysis, empirically reduces cases where a comment is orthogonal to the norm dimension, and leads to easier generalization.

\noindent\textbf{Interpreting \methodname}\hspace{2mm}
Via \methodname, we quantitatively observe a number of features of the RPM model proposed in social science literature \cite{jackson1966conceptual, nolan:2015, linnan2005norms}. 
Specifically, we use the \textbf{point of maximum return}---the highest point on the RPM curve---to locate the ideal normative behavior one should follow to maximize community preference, and the \textbf{potential return difference}---total positive feedback minus total negative feedback---to discover norm regulation strategies; i.e., whether the community tends to use reward or punishment to guide the formation and adaptation of its values.

\subsection{Normness Scale Predictor (NSP)}\label{sec:method:norm_pred}

The Normness Scale Predictor (NSP) quantifies the extent to which a comment exhibits a specified social norm and is decomposed into two stages: normness measurement and normness distillation.

\noindent\textbf{Normness Measurement}\hspace{2mm}
The measurement module should map a comment to a numerical score that represents the scale of normness in the comment. We describe the challenges we tackle to construct a robust norms measurement pipeline. 
First, the intricacy and complexity of social norms make them extremely difficult to learn using a small regression model with limited expressive power and scarce data. Yet, it is not ideal either 
to use an LLM to score the comments directly; although LLMs can perform tasks with few labeled data, they are computationally expensive or rely on external APIs, posing security risks \citep{greshake2023not}. 
To address this, we reformulate the regression task into a binary classification task inspired by \citet{lee2022neural}. Instead of assigning a numerical normness label to a comment, the model only learns the relative normness of comments. 
Then, we obtain numerical normness scales using win-rates 
and mathematically show that this reformulation is equivalent to a regression task given that we are only interested in relative differences in normness scales (\Aref{app:binary_to_numerical}). 

The second challenge is the lack of labeled data; to the best of our knowledge, there is no oracle dataset with normness scale labels. To this end, we automatically label comment pairs in terms of their \emph{relative normness scale} using an LLM with high utility \citep{zheng2023judging} to train a student model \cite{rao-etal-2023-makes, Sorensen2023ValueKE}. 
To summarize, we operationalize the NSP via training a \emph{lightweight binary classifier} using high-quality synthetic labels and evaluate both the synthetic labels and the trained classifier with human annotations.

\noindent\textbf{Normness Distillation}\hspace{2mm}
The normness distillation stage addresses two key challenges. 
First, unlike survey-based social science studies, our approach observes normative behaviors \emph{post-hoc}, lacking the opportunity to explore ``alternative behaviors.'' We attempt to recreate the ``hypothetical conditions'' proposed in \citet{jackson1966conceptual}, in which the individual considers alternative options to maximize return \citep{zhang2023we}. 
We achieve this with a \textbf{Community Language Simulation} \revision{(CLS)} module, which generates comments identical to the original, except for \textit{controlled} variations in one norm dimension. \revision{This design ensures that any confounding factors are controlled, as the generated comment remains identical to the original except for the intended variation.}
We then apply the normness measurement module to quantify the normness scales of the transformed comments. 
E.g., for an original comment, ``\textit{ty!},'' we generate ``\textit{thank you}'' by varying formality, and obtain formality scales of $0.2$ and $0.4$, respectively.

Second, the unconstrained nature of language brings a myriad of potential confounding factors biasing the predictions of the NSP, such as content variations and personal linguistic habits. 
By varying only one norm dimension and comparing the original and rewritten comments, the norm distillation stage aims to mitigate these confounding factors.
In the above example, comparing ``\textit{ty!}'' and ``\textit{thank you}'' eliminates gratitude as a potential confounder for formality.
We use a series of filters to ensure the quality of the generated text, including fluency and content preservation, and evaluate with annotations from in-community members.

\subsection{Community Preference Predictor (CPP)}\label{sec:method:upvote_pred}
The Community Preference Predictor (CPP) estimates community reactions to comments, thereby serving as an indicator of prevailing community norms that govern behavior within online communities. Similar to the NSP, the CPP also consists of a measurement stage and a distillation stage.

\noindent\textbf{Community Preference Measurement} \hspace{2mm}
The measurement stage of the CPP focuses on estimating community preference, which is quantified using net preference scores computed as the number of upvotes minus the number of downvotes of each comment. Unlike the NSP, which requires synthetic labeling, the CPP leverages real-world data for training. To capture the nuances of community approval, the CPP accounts for various contextual factors---post titles and time metadata---in addition to the comments as inputs, and outputs the predicted net community preference score. 

\noindent\textbf{Community Preference Distillation} \hspace{2mm}
Is a comment receiving more upvotes because of its timing, its content, or because the amount of sarcasm is just right? 
To answer such questions, the distillation stage of the CPP aims to isolate the effects of specific norm dimensions on community reactions by calculating the difference in predicted preference between the original comment and its rewrite (which vary only in one norm dimension), and comparing it with the change in normness. 
Returning to the ``\textit{ty!}'' and ``\textit{thank you}'' example (\Sref{sec:method:norm_pred}), the CPP uses identical contextual information and produces community preference scores of $2$ and $5$; thus, a preference increase of $3$ can be attributed to a formality increase of $0.2$.
Overall, this approach addresses confounders such as temporal dynamics and content differences, by constraining variations to a single norm dimension and comparing the preference predictions with the original comments.

\section{Experiments}\vspace{-1mm}
We outline our data curation process (\Sref{sec:exp:datasets}) and describe experiments done to thoroughly validate the Normness Scale Predictor (\Sref{sec:exp:norm_pred}) and the Community Preference Predictor (\Sref{sec:exp:upvote_pred}).

\subsection{Datasets}\label{sec:exp:datasets}\vspace{-1mm}
We obtain data from the Reddit Dump via Academic Torrents, which includes posts, comments, and their metadata. 
Our analysis primarily focuses on first-order comments directly responding to posts from the time period 2019 to 2023.

\noindent\textbf{Inductive Norm Identification}\hspace{2mm}
Given the flexibility of \methodname, we can select any norm dimensions that describe the comments (aka behaviors) in the community. We employ an inductive norm identification process to surface the overarching norms in Reddit communities to use in our experiments as a proof of concept.
First, we assume familiarity of GPT-4 with the top 5,000 subreddits \citep{dignan2024reddit}, and instruct it to categorize them into 30 broad thematic topical groups such as finance or politics. Then, we identify the prominent norm dimensions within each category; for instance, the politics subreddits often consist of \emph{argumentative} discussions. Consultations with subreddit experts help prioritize the six most significant norms based on their prevalence and relevance: Politeness, Supportiveness, Sarcasm, Humor, Formality, and Verbosity.

\noindent\textbf{Subreddit Selection}\hspace{2mm}
We select the subreddit topics of gender, politics, finance, and science based on their relevance and on prior work discussing their norms \cite{nyt-wallstreetbets, Hessel2016ScienceAA, Rajadesingan2020QuickCL, eckert2013language}. For each topic, we select the most active, related subreddits to ensure data scale. See dataset details and sizes in \Aref{app:subreddit_selection}.

\subsection{Normness Scale Predictor (NSP)}\label{sec:exp:norm_pred}\vspace{-1mm}

\subsubsection{Normness Measurement}\label{sec:exp:norm_measure}

\noindent\textbf{Data Preprocessing}\hspace{2mm} 
Each topical group and norm dimension except for the verbosity dimension
\footnote{Instead of training a verbosity scale classifier, we measure verbosity using character count and compute winrates in the range [0-1] based on the count to align with other dimensions.} has a dedicated classifier model, enabling comparisons across similar subreddits. 
Normness measurement relies on synthetic labels generated through stratified sampling and automatic labeling. During the sampling stage, comments are rated on a 5-point Likert scale by GPT-3.5 \citep{brown2020language} to gauge normness (see \Aref{app:norm-scale} for the Likert scale details; \Aref{app:gpt-3.5-eval} for GPT-3.5 rating evaluation details). Then, 10 comments are sampled per scale point per subreddit, resulting in 150 comments per topic (200 for finance with 4 subreddits included). 
From these, 1,250 comment pairs are randomly selected to create binary synthetic labels using GPT-4
\footnote{\revision{We used GPT-3.5 for stratified sampling to save costs, as perfect precision was unnecessary. GPT-4, which performed best in our evaluation (Table \ref{tab:prompt-tuning-eval}), assigned high-quality labels to pairwise comments. See \Aref{app:cost-estimation} for GPT cost estimations.}} \citep{openai2024gpt4}; we detail the GPT-4 prompt tuning and synthetic label evaluations in \Aref{app:gpt-4-eval}.
We train DeBERTa-base \cite{he2020deberta} with the synthetic labels for each of the 4 topic groups and 5 norm dimensions with training details in \Aref{app:norm_training}.

\noindent\textbf{Evaluation}\hspace{2mm}
To evaluate the quality of GPT-4 generated training labels and the NSP models, we curate a high-quality human annotation set of 450 samples for each norm dimension, where each sample is annotated by 3 annotators with an average inter-annotator agreement, measured by Fleiss's kappa, of 0.56 (see \Aref{app:norm-dim-anno} for annotation details). \revision{We then compare the GPT-4 generated labels against the human annotations and present the evaluation results in \Aref{app:gpt-4-human-eval}, with the evaluation of the NSP models detailed in \Aref{app:norm_eval}. Overall, we found that GPT-4 achieved average F1 scores ranging from 75.2-82.4 across the topical groups. In comparison, the NSP models obtained average F1 scores ranging from 74.2-83.0, further validating the quality of the NSP models.}

\subsubsection{Community Language Simulation}\label{sec:cls}
The norm distillation stage of NSP employs a \textbf{community language simulation} module to synthesize comments and control for norm variations.\footnote{\revision{We include confounding factor baselines where only original comments with real upvotes are plotted in Appendix \ref{appendix:rpm-plots-original}. We found that original comments are unevenly distributed across the normness scales in different subreddits (e.g., r/shittyaskscience is mostly sarcastic, r/askscience is mostly serious), making direct comparison challenging and thus further justifying the need to use the CLS module.}}

\noindent\textbf{Data Generation}\hspace{2mm}
To simulate community language, we instruct Llama-3-8B-Instruct \citep{touvron2023llama} to perform linguistic style transfer while preserving the original content and context. The model takes post titles and comment content as input and generates five variations of each comment representing different normness scales, such as: ``\texttt{Very Toxic},'' ``\texttt{Somewhat Toxic},'' ``\texttt{Neutral},'' ``\texttt{Somewhat Supportive},'' ``\texttt{Very Supportive}'' for the Toxic--Supportive dimension. 
See Appendix \ref{app:style_transfer_prompts} for the prompts used for each norm dimension.

\begin{figure}[t!]
    \centering    \includegraphics[width=\linewidth]{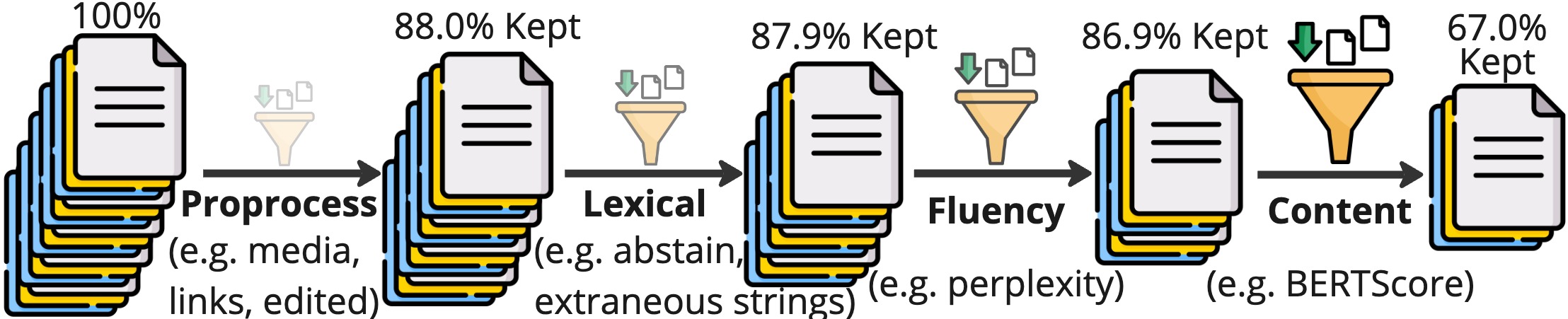}\vspace{-2mm}
    \caption{\textbf{Data filtering pipeline}, including preprocessing, lexical, fluency, and content preservation filters to ensure data quality, keeps 67\% data after filtering.}
    \label{fig:st_filters}\vspace{-3mm}
\end{figure}

\noindent\textbf{Data Processing}\hspace{2mm}
We sample 50K comments per subreddit\footnote{The data is sampled from the subset \emph{not} used to train the community preference predictor, which ensures that the trained CPP model does not perform any inference on its training data in the community preference distillation stage.}
to use as the seed comments for community language simulation. 
To ensure the synthetic data quality, we apply preprocessing, lexical, fluency, and content preservation filters (\Fref{fig:st_filters}) inspired by prior works in style transfer evaluation \cite{briakou-etal-2021-evaluating-custom, mir-etal-2019-evaluating-custom}, removing 33\% of the synthetic comments (\Aref{app:st_filters}).

\noindent\textbf{Evaluation}\hspace{2mm}
Three expert annotators familiar with each topical group evaluated 5 original--synthetic comment pairs per subreddit, resulting in 195 annotated samples. 
The annotators assessed (1) content similarity of the pair, (2) fluency, (3) authorship (LLM or human), and (4) overall quality (i.e., whether the comment is suitable to be posted in the subreddit) of each comment. \Tref{tab:st_annotation} shows that synthetic data fluently preserves content, and is of good overall quality. Expert annotators \emph{failed} to identify synthetic data as machine-generated 50\% of the time. Moreover, postmortem interviews revealed that being ``politically correct'' is a strong identifier for machine-ness, and authorship is indistinguishable in science and finance topics. 
Overall, these results validate the quality of the filtered data. Further details are in \Aref{app:synthetic-data}.

\begin{table}[t!]
    \setlength{\tabcolsep}{3pt}
    \centering
    \resizebox{\linewidth}{!}{
    \begin{tabular}{ccccc}
    \toprule
        \textbf{Metric}    & \textbf{Cont. Sim.}   & \textbf{Fluency}  & \textbf{Authorship}   & \textbf{Holistic}\\
        Threshold          & \footnotesize{roughly similar}\normalsize & \footnotesize{somewhat fluent}\normalsize  
                                                   & \footnotesize{human-written}\normalsize   & \footnotesize{suitable}\normalsize \\ \midrule
        \textbf{Original}      & \multirow{2}{*}{86.0} & 94.0          &  81.0                 & 91.0 \\
        \textbf{Synthetic} &                       & 95.9              &  50.0                 & 71.3 \\
        \bottomrule
    \end{tabular}}\vspace{-2.3mm}
    \caption{\textbf{Human evaluation results} of community language simulation. Numbers indicate the \% of original/synthetic comments rated at/above the threshold.}
    \label{tab:st_annotation}\vspace{-3.5mm}
\end{table}

\subsection{Community Preference Predictor (CPP)}\label{sec:exp:upvote_pred}

\noindent\textbf{Data Preprocessing}\hspace{2mm}  
We take all first-level comments and their associated up-/down-vote counts. We exclude comments deleted, edited, created after 1 day of the post creation time, or created within 1 day of data scraping to obtainthe true preference.

\noindent\textbf{Models}\hspace{2mm} 
CPP is fine-tuned on the DialogRPT model---a dialog response ranking GPT-2 based model trained on 133M data from Reddit \citep{gao2020dialogrpt}. Initializing CPP with DialogRPT weights enhances its understanding of general dialogue dynamics and community preferences.
We train a distinct CPP model for each selected subreddit; 
the fine-tuning process customizes the model to better predict the preference habits of the specific community. 
See \Aref{app:upvote_training} for training details.

\noindent\textbf{Baselines}\hspace{2mm} 
We investigate the effect of contextual data with $4$ input format variants: \textbf{comment only}, \textbf{comment+post}, \textbf{comment+post+timestamp}, and \textbf{comment+post+timestamp+author}.

\noindent\textbf{Evaluation}\hspace{2mm} Following \citet{gao2020dialogrpt}, model performance is evaluated using binary accuracy: whether the relative relations between the predictions and ground truth labels of comment pairs align. 
We found that including contextual information such as the post title and time of the post significantly improves the accuracy, while adding author information only helps in certain subreddits such as \texttt{r/libertarian}. 
The most performant setup, \textbf{comment+post+timestamp}, achieved an accuracy of 73.9\% (±4.1), suggesting reliable prediction performance. 
See detailed results in \Aref{app:upvote_eval}.

\section{Results}\label{sec:results}\vspace{-1mm}
Using the validated NSP and CPP, we explore prevailing norms and values of online communities by modeling return potentials and analyzing the \emph{point of maximum return} (PMR) and 
\emph{potential return difference} (PRD) to corroborate our findings with existing work on similar communities and then uncover additional insights at scale.

\paragraph{Return Potential Modeling (RPM)}
\begin{figure}[t!]
    \centering
    \includegraphics[width=0.9\linewidth, height=5.83cm]{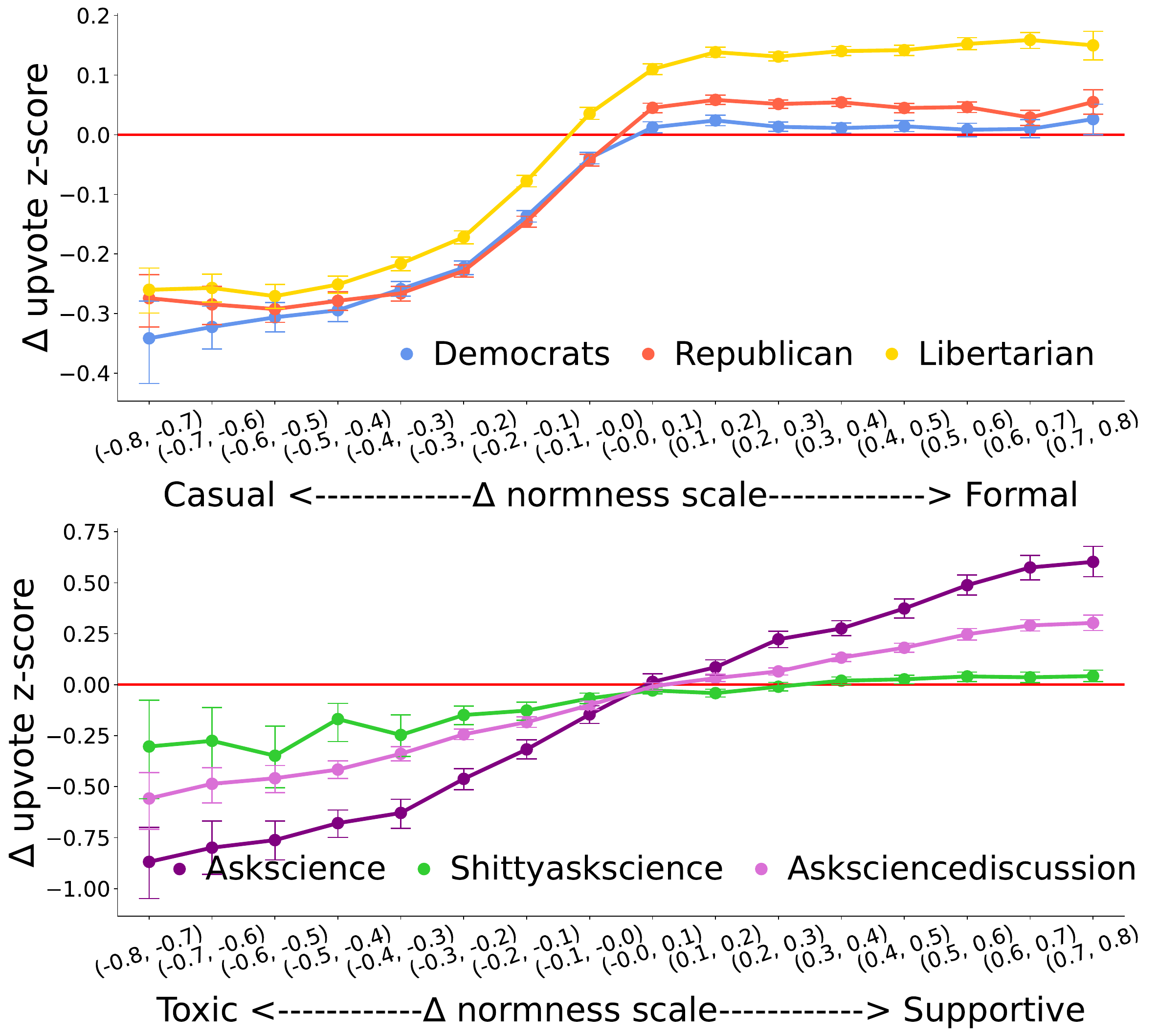}\vspace{-2.3mm}
    \caption{Estimated return potential over normness scales. Formality preferences in politics subreddits (top) and supportiveness preferences in science subreddits both corroborate prior findings about the communities.
    }\vspace{-3mm}
    \label{fig:RPM}
\end{figure}
Our RPM results demonstrate how a community's preferences varies with the scale of normness.
We highlight two key RPM plots---formality preferences in politics subreddits and supportiveness preferences in science subreddits---to validate \methodname in \Fref{fig:RPM} (with full results in \Aref{appendix:rpm-plots}). 

In the politics subreddits, community preference for formal to neutral comments is nearly invariant, but as comments become progressively more casual, there is a steep decrease in preference across all subreddits. These patterns align with community rules that encourage more formal interactions (e.g., ``quality control'' and ``no disinformation'') and denounce casual behaviors (e.g., ``no trolling'' and ``no spamming''). 
Higher preferences toward formal comments in \texttt{r/libertarian} is consistent with its strict guidelines encouraging detailed explanations and references to policies.

The RPM results of science subreddits show a general disapproval for toxic behaviors that gradually changes to approval as the comments become supportive. 
\texttt{r/askscience} and \texttt{r/asksciencedi-} \texttt{scussion}---subreddits designated for scientific discussion with guidelines discouraging offensive language and encouraging helpful answers---show a stronger preference for supportive comments than \texttt{r/shittyaskscience}, which is a parody created to mock \texttt{r/askscience} \cite{Hessel2016ScienceAA}.
Overall, \methodname effectively surfaces community norms shaped by guidelines and core premises.

\paragraph{What Are the Ideal Norm Behaviors?}

\begin{figure}[t!]
    \centering
    \includegraphics[width=0.92\linewidth, height=4.5cm]{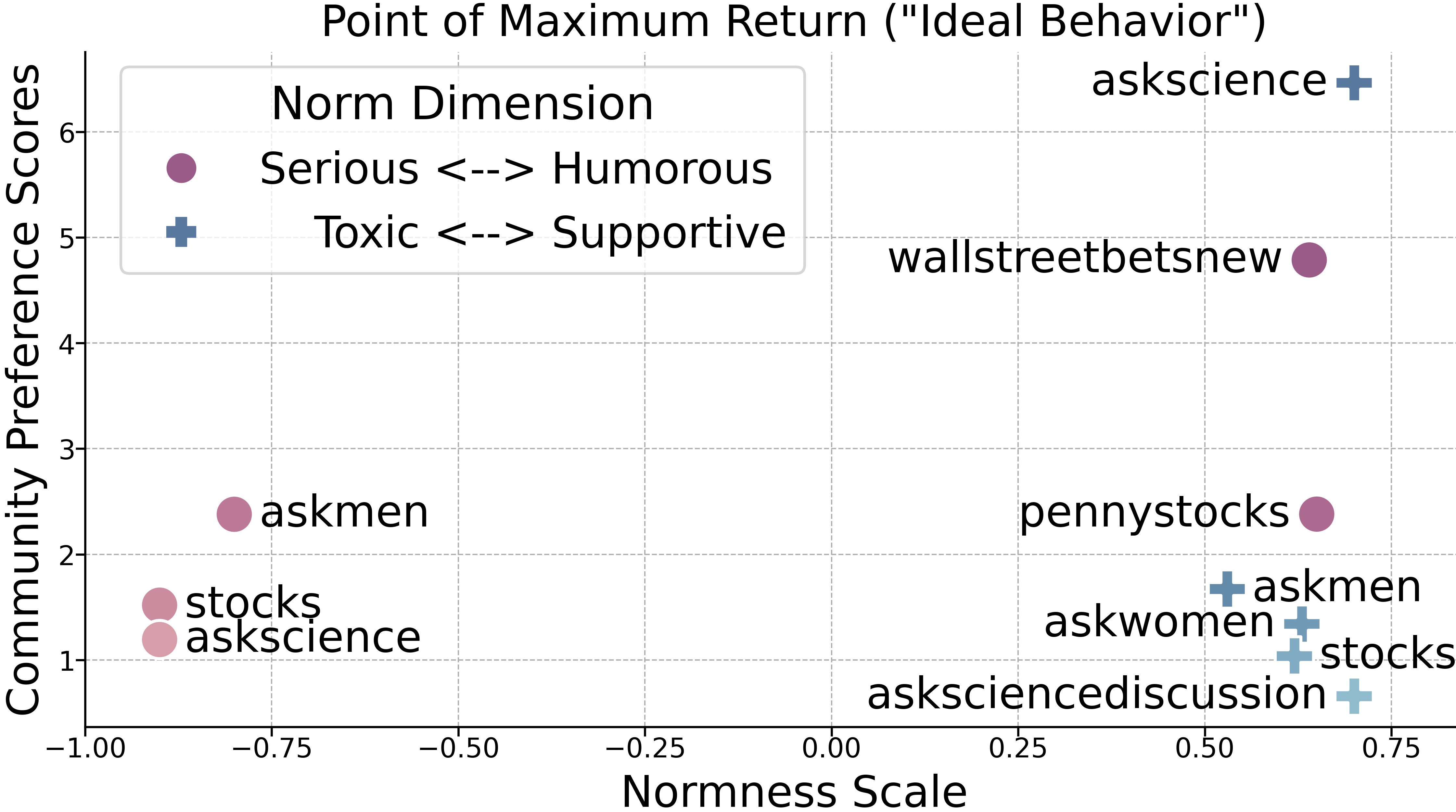}\vspace{-2mm}
    \caption{\textbf{PMR of the top five subreddits for Serious--Humorous and Toxic--Supportive}. The point of maximum return on an RPM curve describes the ``ideal'' behavior that would maximize community preference. For instance, these results show that \texttt{r/askscience} strongly prefers supportive comments.}
    \label{fig:max_return_potential}\vspace{-3mm}
\end{figure}

The point of maximum return (PMR) signifies the behaviors most favored by each community. 
\Fref{fig:max_return_potential} illustrates the PMR for the top $5$ subreddits across humor and supportiveness dimensions. 
For instance, \texttt{r/askscience} prefers supportive comments, as discussed above, and serious comments, which is in line with its explicit community rules (e.g., "memes or jokes are not allowed") and  implicit rules identified in prior work; e.g., ``no personal anecdotes'' \citep{chandrasekharan2018internet}. 
Additionally, all subreddits show a preference for supportiveness over toxicity to varying degrees, which aligns with Redditquette, which are informal values held by most redditors \citep{fiesler2018reddit}. 
See \Aref{app:max_return} for PMR results in all dimensions.

\paragraph{Inferring Norm Regulation Strategies}
\begin{figure}[t!]
    \centering
    \includegraphics[width=0.92\linewidth] {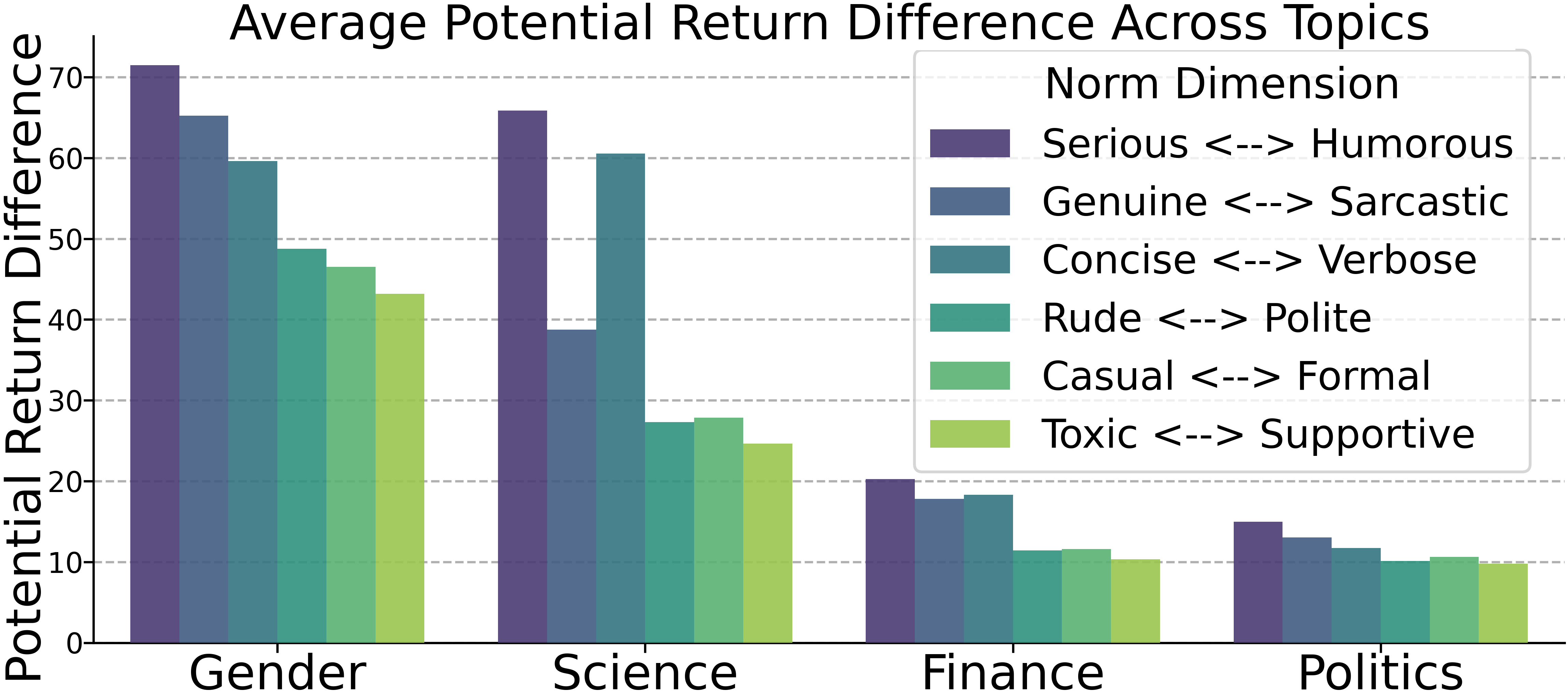}\vspace{-2.2mm}
    \caption{\textbf{PRD across topical groups}, reflecting the feedback strategy used by the community to regulate certain norms. All studied communities tend to use positive feedback: the gender related subreddits extensively reward behaviors aligned with their values, while the politics subreddits reward much more conservatively.}\vspace{-3mm}
    \label{fig:potential_return_difference}
\end{figure}

Potential return differences (PRD) in \Fref{fig:potential_return_difference} reveal how much communities emphasize rewards (PRD$>$0) or punishments (PRD$<$0) to enforce norms.
All communities significantly favor positive reinforcement, indicating a generally supportive atmosphere \citep{jackson1966conceptual}, echoing calls for positivity in Redditquette \citep{fiesler2018reddit}. 
Moreover, punitive measures are ineffective in maintaining prosocial communities \citep{mulder2008difference, de2019people, shen2022tale}. 

Feedback intensity distinctly varies across topics. Gender-related subreddits extensively reward behaviors aligned with their values, suggesting a strong preference for promoting norms that enhance inclusivity and respect. Politics subreddits are more conservative with rewards, possibly due to explicit rules against ``disproportionate upvoting'' and ``brigading,'' which aim to prevent bias. These regulations may contribute to more measured rewards.
Lastly, PRD variations across norm dimensions reveal which normative behaviors are most regulated. The serious--humorous, genuine--sarcastic and concise--verbose dimensions witness the most intense regulation in all groups, suggesting the importance of tone and authenticity of interactions in cultivating social identity \citep{brown2022identities}.

Findings in this section validate \methodname and, more importantly, allude to the impact of moderation on social norms and potential applications of \methodname: if undesirable behaviors are detected to rise, moderation strategies should be updated to maintain healthy community norms.

\section{Analysis}\label{sec:analysis}
To address RQ2---\emph{Can we predict the change in norms based on observed normative behaviors?}---we study the fluidity and stability of social norms and its implications using \methodname and social science theories, specifically norm intensity and crystallization \cite{jackson1966conceptual, nolan:2015}, then analyze their temporal changes in the context of external events and internal community conflicts. 
\paragraph{Norm Crystallization}

\begin{table}[t]
\centering
\resizebox{0.85\columnwidth}{!}{
\begin{tabular}{lccccc}
\toprule
\textbf{} & \multicolumn{2}{c}{{$NI$-only}} & \multicolumn{3}{c}{{$NI$+$CR$}} \\
\cmidrule(r){2-3} \cmidrule(r){4-6}
\textbf{} & \textbf{c$_{NI}$} & \textbf{R$^2$} & \textbf{c$_{NI}$} & \textbf{c$_{CR}$} & \textbf{R$^2$} \\
\midrule
Politeness & 0.26 & 0.17 & 0.16 & -0.14 & 0.23 \\
Supportiveness & 0.16 & 0.04 & 0.05 & -0.13 & 0.10 \\
Sarcasm & 0.42 & 0.13 & 0.45 & -0.13 & 0.14 \\
Humor & 0.50 & 0.27 & 0.50 & -0.13 & 0.28 \\
Formality & 0.40 & 0.17 & 0.27 & -0.07 & 0.18 \\
Verbosity & 2.57 & 0.09 & 2.57 & -0.35 & 0.09 \\
\bottomrule
\end{tabular}
}\vspace{-2mm}
\caption{\textbf{Coefficients of $NI$ and $CR$, and $R^2$} of two linear regression models ($NI$-only and $NI$+$CR$).}
\label{tab:crystallization}\vspace{-4mm}
\end{table}

Social norms are constantly evolving. Understanding such changes and their predictive features can help community moderators respond effectively.
\citet{jackson1966conceptual} introduces the concepts of \emph{norm intensity} ($NI$) and \emph{crystallization} ($CR$). $NI$ measures the magnitude of community (dis)approval of behaviors at a given normness scale, indicating how strongly the community cares about the norm, while $CR$ represents the level of consensus on the preference.

Taking the year 2021 as a cutoff, we test the predictive power of $NI$ and $CR$ on upcoming temporal changes ($TC:=\Delta NI$) with a linear regression model. We use results from \methodname predictions and follow implementation defined in \citet{linnan2005norms} (details in \Aref{app:crystallization}). 
Our results in \Tref{tab:crystallization} show that $NI$ and $NI+CR$ are both significant predictors of $TC$, while adding $CR$ increases the coefficient of determination $R^2$ significantly. 
Additionally, higher norm intensity and less crystallization (i.e., community members have strong opinions but less agreement) are correlated with larger shifts in norm intensity. 
Our findings support \citet{jackson1975normative}'s hypothesis that these volatile instances are more likely to generate conflicts and trigger changes in norms. This demonstrates \methodname's potential to help moderators identify and proactively address norms likely to change by setting explicit community rules.

\begin{figure}[t!]
    \centering
    \includegraphics[width=0.91\linewidth, height=7.3cm]{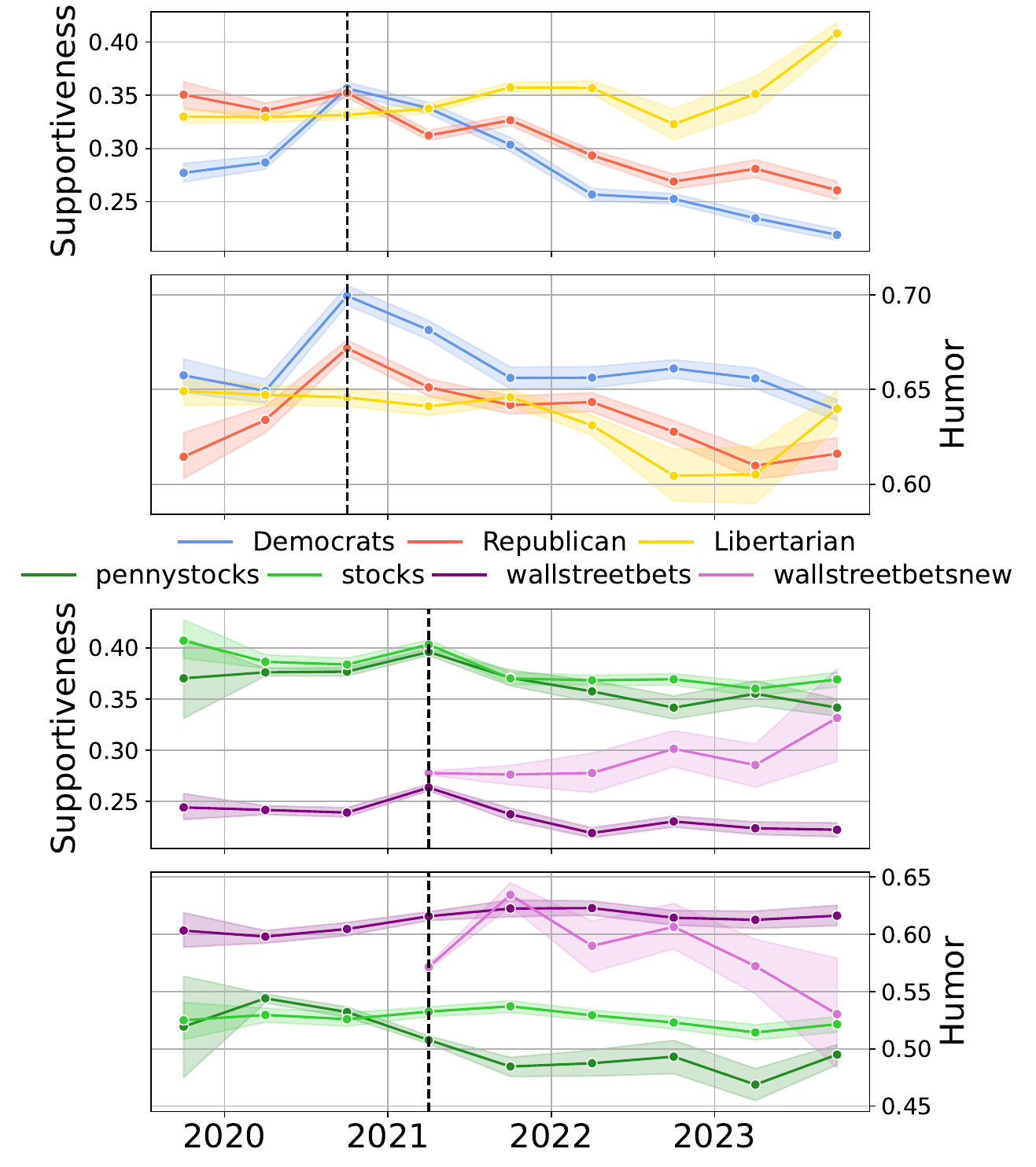}\vspace{-2.5mm}
    \caption{\textbf{Temporal changes in average norm intensity} for politics and finance subreddits. Comments were binned by 6 month intervals based on their posting date. For instance, a point for 2020.25 represents the average norm intensity of comments posted from January to June 2020. The vertical lines mark two events: the U.S. presidential election and the creation of \texttt{r/wallstreetbetsnew}, highlighting changes before and after these events.}
    \label{fig:temporal_change}\vspace{-3mm}
\end{figure}

\begin{table*}[ht]
\centering
\resizebox{0.8\linewidth}{!}{
\begin{tabular}{@{}lccccc@{}}
\toprule
\textbf{community shift / norm dimension} & \textbf{politeness} & \textbf{supportiveness} & \textbf{sarcasm} & \textbf{humor} & \textbf{formality} \\ 
\midrule
\texttt{r/wallstreetbets} $\rightarrow$ \texttt{r/wallstreetbetsnew} (925.6) & \graycell{-0.003} & \graycell{0.013} & \graycell{0.003} & \graycell{0.005} & \greencell{0.018} \\ 
\texttt{r/wallstreetbets} $\rightarrow$ \texttt{r/stocks} (2157.6) & \greencell{0.084} & \greencell{0.092} & \redcell{-0.044} & \redcell{-0.062} & \greencell{0.131} \\ 
\texttt{r/wallstreetbets} $\rightarrow$ \texttt{r/pennystocks} (1052.0) & \greencell{0.091} & \greencell{0.094} & \redcell{-0.023} & \redcell{-0.084} & \greencell{0.063} \\ 
\texttt{r/askwomen} $\rightarrow$ \texttt{r/askmen} (717.4) & \graycell{-0.015} & \graycell{-0.022} & \graycell{0.026} & \greencell{0.036} & \graycell{0.004} \\ 
\texttt{r/republican} $\rightarrow$ \texttt{r/democrats} (223.8) & \graycell{0.026} & \graycell{0.016} & \greencell{0.036} & \graycell{0.018} & \graycell{-0.008} \\ 
\bottomrule
\end{tabular}
}\vspace{-3mm}
\caption{User behavior shifts in select subreddit transition pairs. Gray cells indicate changes that are insignificant (p > 0.05); red and green~cells represent significant negative and positive changes.}\label{tab:norm-diff-user-partial}\vspace{-2mm}
\end{table*}

\paragraph{Temporal Change in Norm Intensity}

We further investigate how $NI$ changes over time, particularly in relation to external events. \Fref{fig:temporal_change} shows $NI$ of the humor and supportiveness dimensions from 2019-2023 in politics and finance subreddits.

For politics, a significant event during this period is the 2020 U.S. presidential election, represented by the vertical line in the plot (corresponding to July-December 2020). Our results reveal highly similar patterns of norm shifts in \texttt{r/republican} and \texttt{r/democrats}, characterized by a steep increase of community preference of humor and supportiveness during the election period. Following this peak, both dimensions experienced a continuous decline until 2023. On the other hand, \texttt{r/libertarian} bears a notable increase in supportiveness over time and was not impacted as much by the election. 
These results suggest that external events, such as elections, could potentially shape the overall norms in online communities. 

For finance subreddits, a notable event was the creation of \texttt{r/wallstreetbetsnew}---a spinoff from \texttt{r/wallstreetbets}---in 2021 by members dissatisfied with the culture of \texttt{r/wallstreetbets} in an attempt to create a less toxic environment focused on serious trading strategies on risky stocks.%
\footnote{As one user noted: ``The moderators in the original \texttt{r/wallstreetbets} are driving the narrative away from \$GME and \$AMC and the vibe is very negative/toxic over there'' (paraphrased from a subreddit post in \texttt{r/wallstreetbetsnew}).} 
Among the finance subreddits, our results show that the $NI$ of \texttt{r/wallstreetbetsnew} starts diverging from \texttt{r/wallstreetbets} and begins to resemble the $NI$ of \texttt{r/stocks} and \texttt{r/pennystocks}, becoming more supportive and less humorous over time. 
This finding aligns with \citet{zhang2021understanding} in showing that new communities establish their own identities and norms over time. 
Additionally, after the creation of \texttt{r/wallstreetbetsnew}, the $NI$ of \texttt{r/wallstreetbets} also shifts, becoming less supportive and more humorous. This suggests that the culture of the original community may be influenced when some members leave to form a new spinoff community as explored below.

\paragraph{Community Norm Adaptation by Users}\label{sec:analysis_user}

Social norms can influence the behavior of community members \citep{ mcdonald2015social}, so we examine how individual users modify their language and interaction styles based on the subreddit they are participating in. 
We define user-level norm behavior in a community as the average $NI$ of comments left by the specific user in that community. 
For related subreddits with shared users, we compute the change in normative behavior of these users when they switch from subreddit A to subreddit B using a paired two-tailed t-test (\Tref{tab:norm-diff-user-partial}), with experimental details and full results in \Aref{appendix:user-norm}. 

Our results reveal significant variability in user normative behaviors between the selected subreddit pairs. For example, users in \texttt{r/wallstreetbets}, known for its usage of profane jargon and aggressive trading strategies \cite{nyt-wallstreetbets}, significantly modify their behaviors in \texttt{r/stocks} and \texttt{r/pennystocks}, but adapt much less in the spinoff subreddit \texttt{r/wallstreetbetsnew}. Additionally, user behaviors tend to remain consistent in identity-related subreddits (e.g., \texttt{r/askwomen}, \texttt{r/askmen}) or those with competing relationships (\texttt{r/republican}, \texttt{r/democrats}), highlighting the context-specific nature of community norm adaptations by users. We also observe that users are more likely to change their formality to fit different subreddit contexts than other dimensions, such as humor, indicating that certain norms are more malleable and adaptable than others. 

Different extents to which users adapt their language to the audience suggest that digital identities are fluid and context-dependent. 
This can inform the development of tailored moderation tools to align with the behavioral norms of specific communities, potentially improving user experience and engagement on a more fine-grained level.

\section{Conclusion \& Future Directions}
We introduced \methodname, a novel framework based on the RPM theory from social science, to quantify social norms and values at scale. 
We comprehensively validated the effectiveness of \methodname to assess the normness of behaviors and predict community preferences while controlling for confounders. 
\methodname enables numerous quantitative analyses, including predicting norm shifts and contextualizing temporal changes with external events, providing a deeper understanding of social norm dynamics in online communities.

Our work contributes a robust and generalizable method that can be easily extended to various norms and communities. It opens up many exciting possibilities for applications and future research:

\noindent\textbf{Computational Modeling Applications}\hspace{2mm}
Our framework can enhance community moderation tools by integrating theoretically grounded insights, such as maximum return potential, to refine toxicity detectors. It can also guide generation models to produce contextually appropriate responses specialized to each community's unique norms.

\noindent\textbf{Applications for Social Scientists}\hspace{2mm}
Our method empowers the development of new hypotheses about social norms, by providing social scientists with enhanced tools to explore how norms form and influence social interactions within communities.

\noindent\textbf{Support Tools for Communities}\hspace{2mm}
\methodname can enhance community management by enabling moderators to  monitor and address norm shifts in real-time. It can help transform widely accepted but informal norms into explicit rules,  clarifying guidelines and easing new member integration. This approach is applicable in various settings (e.g., workplaces) where it can guide individuals on appropriate cultural expressions, improving their integration and acceptance. Platform developers can use this method to refine community recommendation engines, aligning users with groups that match their preferences and values, thereby enhancing user engagement and community growth.

\section*{Limitations}

\textbf{Return Potential Model} In this work, we introduce \methodname, a novel framework based on the RPM theory in social science. However, the RPM specifically measures the potential approval by other community members, representing only one dimension of broader norm structures in a community. Prior cross-sectional survey work employed the RPM and expanded towards the descriptive dimension of norms\footnote{Descriptive norms represents the beliefs of common or typical behaviors.} \cite{extending_RPM}. Future works can expand our current computational model of RPM, incorporating the broader norms and values within online communities. 

\paragraph{Platform and Language Scope}
While \methodname is not limited to any specific platform or language, our work focused on English comments on Reddit. We believe interesting future directions include extending our framework to various other platforms that provide similar community preference signals, such as YouTube comments. Additionally, expanding to other languages would enable more in-depth cross-cultural analyses of community norms.

\paragraph{Role of Other Stakeholders} To understand the implicit norms in communities, we focus on the interactions between community members through comments and their upvotes. However, stakeholders such as users, moderators, and other interested parties constantly negotiate norms in online communities \cite{kim2006community}. Thus, future works should explore the role of moderators and other stakeholders in potentially shaping the implicit norms in online communities.

\paragraph{Dynamic Nature of Norms} 
Our study quantifies and predicts the community norms and values at scale. However, as shown in \Sref{sec:analysis}, norms are dynamic and constantly changing over time \cite{bicchieri2005grammar}. Our methodology, such as the RPM and the experimental setup, are compatible with future temporal analyses. 

\paragraph{Predictions on Synthetic Comments} In our work, we employ synthetic comments to simulate community preference for comments with varying normness scale. Predicting the community approval of synthetic comments may potentially add noise to our results. However, we aimed to address this limitation by employing an extensive filtering process based on prior works \cite{briakou-etal-2021-evaluating-custom, mir-etal-2019-evaluating-custom} and validating the quality of the filtered data using expert human annotations (See \Sref{sec:cls}). 

\paragraph{Investigating deeper and beyond norm dimensions and community topics.} 
In \Sref{sec:exp:datasets}, we employ an inductive norm identification process to surface six overarching norm dimensions and select subreddit topics based on prior works. However, there are several other dimensions to explore beyond these six, such as optimism, empathy, and confidence. Meanwhile, there are several other relevant and interesting subreddit topics, such as ones based on cultures and nations (\texttt{r/korea} and \texttt{r/southafrica}). \methodname can facilitate future analyses on different norm dimensions and topics of communities. 

\paragraph{Model Error Cascades} We train small local models as the normness and preference predictors. Despite extensive model training and experimentation,the error rates in our \methodname pipeline may potentially influence our downstream analysis. Thus, we designed our pipeline to mitigate as much noise as possible (for example, ``Community Preference Distillation'' in \Sref{sec:method:upvote_pred}) and validate our findings with prior work and existing community guidelines.

\section*{Ethical Considerations}
We use publicly accessible LLMs to conduct our research, which includes generating more toxic versions of comments.
In our investigation to understand the implicit norms of online communities, our experiments inevitably produced toxic content to measure how communities react to toxicity. 
However, we believe the benefits of our research outweigh the risks, as community moderators and platform developers can use our framework to understand the implicit norms in various communities, especially in response to toxic content, and self-assess and monitor their culture.
The generated toxic content was only used to compute aggregated metrics to identify high-level patterns, and it will not be released to the public. To ensure reproducibility while protecting the rights of Reddit users, we will only release the IDs of the comments used in our analysis. Using these provided IDs, practitioners will need to independently fetch the comments from the publicly accessible Reddit Dump.

\section*{Acknowledgement}
This work was supported by the National Science Foundation (Grant No. IIS-2143529) and the National Institutes of Health (Grant Nos. 5R21DA056725-02, R21DA056725-01A1). 
We gratefully acknowledge support from the National Science Foundation under CAREER Grant No.~IIS2142739, and NSF grants No.~IIS2125201 and IIS2203097. 

\bibliography{custom}
\bibliographystyle{acl_natbib}

\appendix

\section{Nomenclature \& Definition References}\label{app:definition}
\begin{itemize}[leftmargin=*, topsep=0pt, noitemsep]
\item \textbf{Norm:} Informally agreed-upon rules governing community behavior, such as the expectation of toxicity or politeness in interactions.
\item \textbf{Value:} The deeper ideals and principles that a community aspires to embody and promote. Values are fundamental in shaping and guiding the development of norms.
\item \textbf{Behavior:} The observable actions taken by community members, such as the comments they post in a subreddit.
\item \textbf{Norm Dimension:} Attributes or characteristics of behaviors that can be measured along a (bidirectional) continuum, serving as a quantitative axis for analyzing norm adherence.
\item \textbf{Normative Behavior:} Actions that align with a specific norm dimension, such as expressions of support or aggression in user comments.
\item \textbf{Normness Scale:} A metric indicating the extent to which a behavior conforms to a particular norm dimension.
\item \textbf{Community Preference:} The collective judgment expressed by community members through mechanisms of approval or disapproval, quantified by the net balance of upvotes and downvotes a comment receives.

\end{itemize}

\section{Converting Binary Classification to Continuous Normness Scale}\label{app:binary_to_numerical}

We reformulate the normness scale measurement module from a regression task to a binary classification task. After getting the binary labels of pairs of comments, we convert the binary labels into numerical scores as follows:

Given comments $\mathcal{A}=\{a_1,\ldots,a_n\}$ with ground truth normness scales $\Phi_d(\mathcal{S})=\{\phi_1\ldots,\phi_n\}$, we have binary labels $\mathcal{B}_d=\{\beta_{ij}|1\leq i<j\leq n, \beta_{ij}=
\begin{cases}1\,\,\,\text{if }\phi_i<\phi_j,\\
0\,\,\,\text{otherwise},\end{cases}$ 
as target labels of the classifier $\mathcal{M}_d:\mathcal{A}\times\mathcal{A}\rightarrow\{0,1\}$. 
For any comment $a_k\in\mathcal{A}$, its adjusted normness scale, $\phi_k'=\Phi_{d}'(a_k)$, is defined as the win-rate of $a_k$ compared against all other comments in $\mathcal{A}$: \vspace{-1.5mm}
\begin{equation}
    \Phi_{d}'(a_k):=\frac1{n-1}(\sum_{i=1}^{k-1}\beta_{ik}+\sum_{i=k+1}^{n}(1-\beta_{ki})),\vspace{-1.5mm}
\end{equation}
which is the percentage of times that $a_k$ is labeled as having a higher normness degree, when compared with other comments in the set of comments $\mathcal{A}$.

\subsection{Monotonicity of Binary Win-rate as Normness Scale}\label{sec:app:binary_proof}
We now should that if we are only interested in the relative normness scales of comments, the binary win-rate and the normness scale are monotonic. 

Recall that for a set of comments $\mathcal{A}=\{a_1,\ldots,a_n\}$ with ground truth normness scales $\Phi_d(\mathcal{S})=\{\phi_1\ldots,\phi_n\}$, we have binary label set $\mathcal{B}_d=\{\beta_{ij}|1\leq i<j\leq n\}$ defined above, from which we obtain $\Phi_{d}'(a_k)$. 

We prove that the two metrics $\Phi_d$ and $\Phi_d'$ are monotonic with respect to each other by showing that, 
$\forall i,j\in[1,n]$ s.t. $\,1\leq i<j\leq n\in\mathbb{R}$ s.t. if $\phi_i\leq\phi_j$, then $\phi_i'\leq\phi_j'$ and if $\phi_i\geq\phi_j$, then $\phi_i'\geq\phi_j'$.

First, let $\mathcal{A}^{-}$, $\mathcal{A}^{+}$, $\mathcal{A}^*$ be subsets of $\mathcal{A}$ such that \vspace{-2mm}
\begin{align}
    \mathcal{A}^{-}&:=\{a'|\Phi_d(a')<\phi_i\},\\
    \mathcal{A}^{+}&:=\{a''|\Phi_d(a'')\geq\phi_j\}\text{, and}\\
    \mathcal{A}^*&:=\{a^*|\Phi_d(a^*)\geq\phi_i, \Phi_d(a^*)<\phi_j\}. 
\end{align}

Let $p=||\mathcal{A}^{-}||,\,q=||\mathcal{A}^{+}||,\,r=||\mathcal{A}^*||$ and $s=\mathcal{I}_{\{\phi_i<\phi_j\}}(i,j)$ (the indicator function where $s=1$ if $\phi_i<\phi_j$ and $s=0$ if $\phi_i=\phi_j$). Then, we can compute win-rates $\phi_i'$ and $\phi_j'$ as: \vspace{-2mm}
\begin{align}
    \phi_i' &= \frac{1}{n-1}(p\cdot1+q\cdot0+r\cdot0+s\cdot0)\\
            &=\frac{p}{n-1}\\
    \phi_j' &= \frac{1}{n-1}(p\cdot1+q\cdot0+r\cdot1+s\cdot1)\\
            &=\frac{p+r+s}{n-1}.
\end{align}

Since $r\geq0$ and $s\geq0$, we have $\phi_j'-\phi_i'=\frac{1}{n-1}(r+s)\geq0$ and $\phi_i'-\phi_j'\leq0$. 
Thus, we have $\phi_i'\leq\phi_j'$ for arbitrary $i$ and $j$. Similarly, we can show that if $\phi_i\geq\phi_j$, then $\phi_i'\geq\phi_j'$. Therefore, we proved that the two metric are monotonic.

\section{Subreddit Selection Details}\label{app:subreddit_selection}
To form the dataset used in this study, we first select subreddit topics based on relevance and prior work, obtaining gender, politics, finance, and science. Then, for each topic, we take the most representative subreddits out of the top 5,000 SFW (safe-for-work) subreddits based on the size of the subreddit. For the gender topical group, we have \texttt{r/askmen}, \texttt{r/askwomen} and \texttt{r/asktransgender}; for the politics topical group, we have \texttt{r/republican}, \texttt{r/demcorats} and \texttt{r/libertarian}. For the science topical groups, we select \texttt{r/askscience}, its spinoff subreddit \texttt{r/shittyaskscience} which was created to mock \texttt{r/askscience}, and a more open variant \texttt{r/asksciencediscussion} that discusses topics \emph{in} science and \emph{related} to science, such as academia \citep{Hessel2016ScienceAA}. Lastly, for the finance-related topics, we selected the most popular three subreddits from the top 5,000: \texttt{wallstreetbets}, \texttt{stocks}, \texttt{pennystocks}, and additionally consider \texttt{r/wallstreetbetsnew}, which is the spinoff subreddit of \texttt{r/wallstreetbets}. 
Table \ref{tab:datasets} summarizes the topics, subreddits, and dataset sizes examined in this study.

 \begin{table}[h!]
    \centering
    \setlength{\tabcolsep}{4pt}
    \resizebox{\linewidth}{!}{
    \begin{tabular}{c@{\hskip -1mm}c@{\hskip -1mm}cc}
    \toprule
        \textbf{Topic} & \textbf{Subreddit} & \textbf{Raw Data} & \textbf{Synthetic Data} \\ \midrule
        \multirow{3}{*}{\textbf{Gender}} 
        & \texttt{r/askmen} & 4.56M & 1.08M \\
        & \texttt{r/askwomen} & 2.13M & 1.21M\\
        & \texttt{r/asktransgender} & 1.61M & 1.01M \\ \midrule
        \multirow{3}{*}{\textbf{Politics}} 
        & \texttt{r/libertarian} & 3.66M & 1.00M\\ 
        & \texttt{r/democrats} & 534K & 922K \\
        & \texttt{r/republican} & 502K & 1.01M\\ \midrule
        \multirow{3}{*}{\textbf{Science}} 
        & \texttt{r/askscience} & 426K & 1.23M\\
        & \texttt{r/shittyaskscience} & 185K & 761K\\
        & \texttt{r/asksciencediscussion} & 141K & 1.10M\\ \midrule
        \multirow{4}{*}{\textbf{Finance}} 
        & \texttt{r/stocks} & 3.51M & 1.05M\\
        & \texttt{r/pennystocks} & 1.23M & 1.04M\\
        & \texttt{r/wallstreetbets} & 49.3M & 864K \\
        & \texttt{r/wallstreetbetsnew} & 655K & 784K \\ 
        \bottomrule
    \end{tabular}}\vspace{-2mm}
    \caption{Selected online communities (subreddits) across various topics. For each subreddit, we show the number of existing comments within the community (column ``Raw Data'') and the number of synthetic comments remaining after applying filters to ensure the quality of the simulated comments (column ``Synthetic Data'').}
    \label{tab:datasets}
\end{table}

\section{Grounding 5-point Scale for Normness Ratings}\label{app:norm-scale}

In \Sref{sec:exp:norm_pred}, we employ a 5-point Likert scale using GPT-3.5 to rate comments and sample them to gauge their normness. Additionally, in \Sref{sec:cls}, we generate five variations of each original seed comment based on the 5 different scales of normness. Thus, for each norm dimension, we created a 5-point Likert scale and grounded their definitions in prior works \cite{dementieva2022detecting, 10.1145/3038912.3052591, 10.1145/3124420, goffman1955face, brown1987politeness, lakoff1973logic}. For example, we define formality based on using abbreviations, slang, colloquialisms, non-standard capitalizations, complete sentences, contractions, punctuations, and opening expressions of sentences \cite{dementieva2022detecting}. Meanwhile, we define politeness as a set of strategies for conducting face-threatening acts while minimizing the chance that we or others will lose our positive or negative faces. \cite{brown1987politeness}. The 5-point Likert scale across the norm dimensions can be found in Figures \ref{fig:ha-guideline}-\ref{fig:ha-guideline3} as well as Figure \ref{appendix:st_ratings}.

\section{GPT Evaluations}\label{app:prompt-tuning}
Recall in \Sref{sec:exp:norm_measure} that we employ GPT-3.5 to sample and rate comments on a 5-point Likert scale (defined in \Aref{app:norm-scale}) for a particular norm dimension and subsequently use GPT-4 to generate binary synthetic labels comparing a pair of comments. In \Aref{app:norm-dim-anno}, we describe the process of curating human annotations. In \Aref{app:gpt-3.5-eval}, we evaluate the quality of GPT-3.5 rating. In \Aref{app:gpt-4-eval}, we describe our prompt design considerations and prompt tuning results. In \Aref{app:gpt-4-human-eval}, we evaluate the final GPT-4 automatic pairwise labeling pipeline using the human annotations. 

\subsection{Normness Scale Annotation}\label{app:norm-dim-anno}
To evaluate the NSP models and the quality of GPT-4 generated labels for student models, we curate a high-quality human annotation set of 450 samples for each norm dimension. \revision{The human annotations of norms are challenging due to subjectivity. To reduce subjectivity, we conducted training sessions with annotators and iteratively improved our annotation guidelines, grounding the definitions of various norms based on prior works (see \Aref{app:norm-scale}).} Each sample was annotated by three volunteer annotators, who are graduate students in NLP and Linguistics at a US-based institution \revision{and familiar with the subreddits in our study.} We did not provide payment, but we obtained consent to use their annotations for AI model evaluation.

For each topic, we use stratified random sampling to select two comments from various subreddits, creating pairs of comments. We then ask three human annotators to make binary judgments on which comment exhibits a higher normness scale for five norm dimensions (e.g., which one is more formal/less casual?). For each annotation, we chose the binary judgment with at least a majority agreement among three annotators\footnote{We discarded annotated samples whose final labels were ``hard-to-tell'' or ``media-needed'' as these samples could not be properly annotated with the given context.}. 

Across the four topics, we collected human annotations for 450 samples\footnote{For all topics except ``Gender,'' we annotated 100 randomly-sampled pairwise comments. For ``Gender'' topic, we annotated 150 pairwise comments, in which 100 pairwise comments came from \texttt{r/askmen} and \texttt{r/askwomen} while the remaining 50 pairwise comments came from comparisons with one of the gender subreddits (including \texttt{r/asktransgender}) and \texttt{r/asktransgender}.}. Each sample was annotated for five norm dimensions, resulting in a total of 2,250 annotations per human annotator.  

The average inter-annotator agreement, measured by Fleiss's $\kappa$, was 0.56, considered a moderate agreement \cite{landis1977measurement}. \revision{Due to the nuance and subtlety of norms, Fleiss's $\kappa = 0.56$ provides a solid foundation for our annotation labels. For instance, \citet{passonneau-carpenter-2014-benefits} reported scores as low as 0.2 in subjective tasks such as word sense annotations.} Refer to Table \ref{tab:nnotator-agreement} for the full agreement scores across 4 topics and 5 norm dimensions. 

\begin{figure*}
    \centering
    \includegraphics[width=\textwidth]{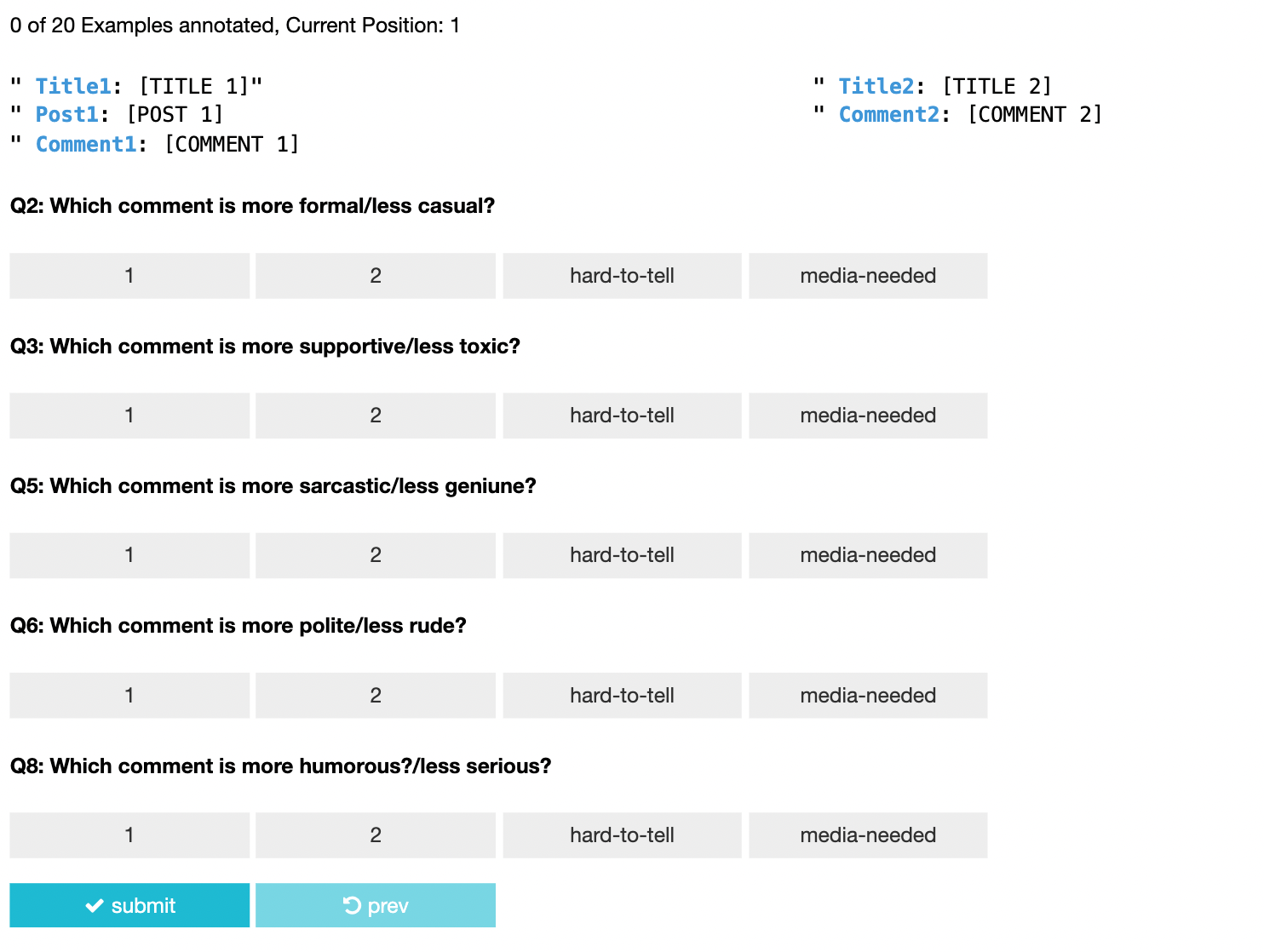}
    \caption{Human annotation UI for the binary norm dimension classification task. For each question, two options (1, 2) were provided without a tie option. Additionally, there were two extra options to mark samples that could not be properly annotated with the given context (hard-to-tell, media-needed).}
    \label{fig:ha-ui}
\end{figure*}

\begin{figure*}
    \centering
    \includegraphics[width=\textwidth]{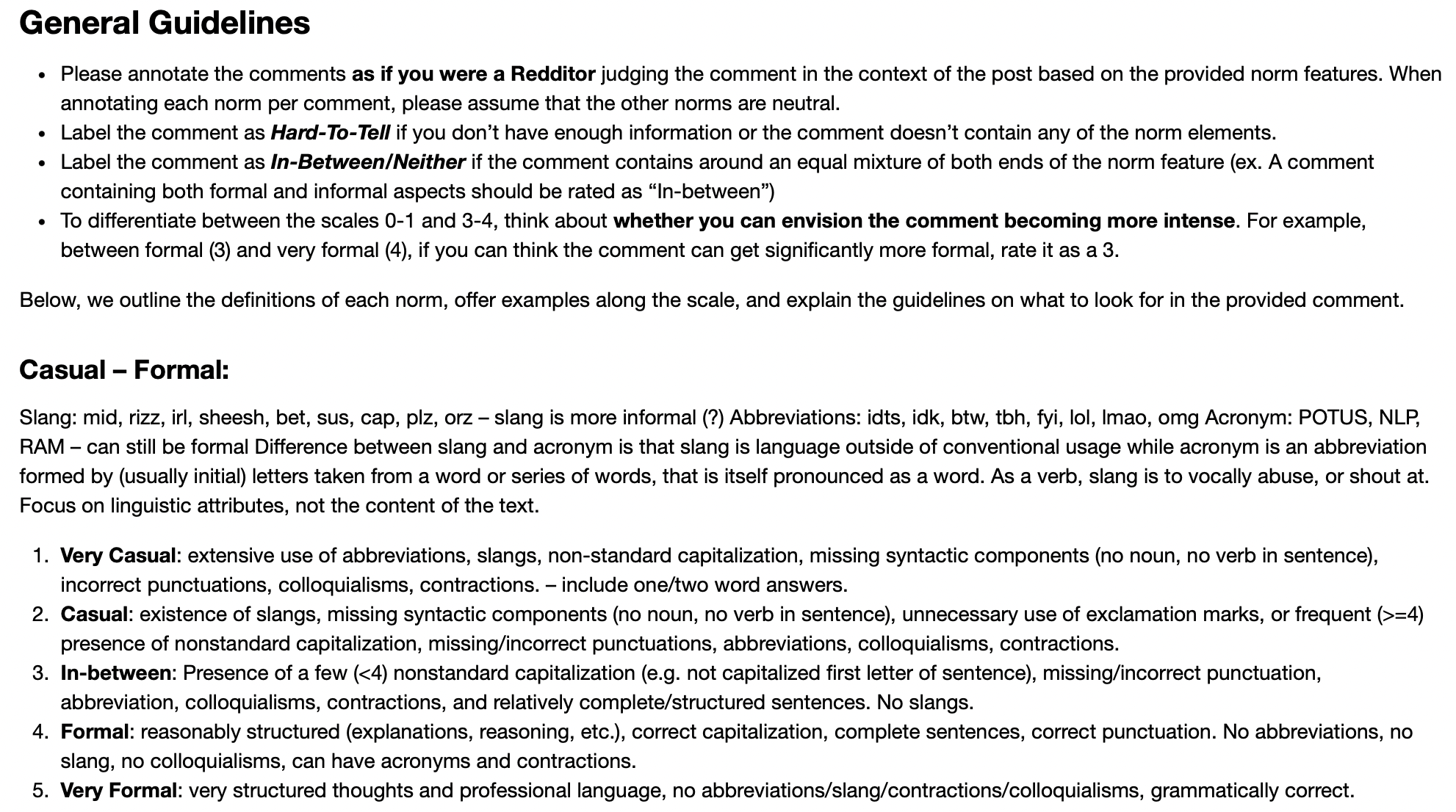}
    \caption{Annotation guideline provided to human annotators.}
    \label{fig:ha-guideline}
\end{figure*}

\begin{figure*}
    \centering
    \includegraphics[width=\textwidth]{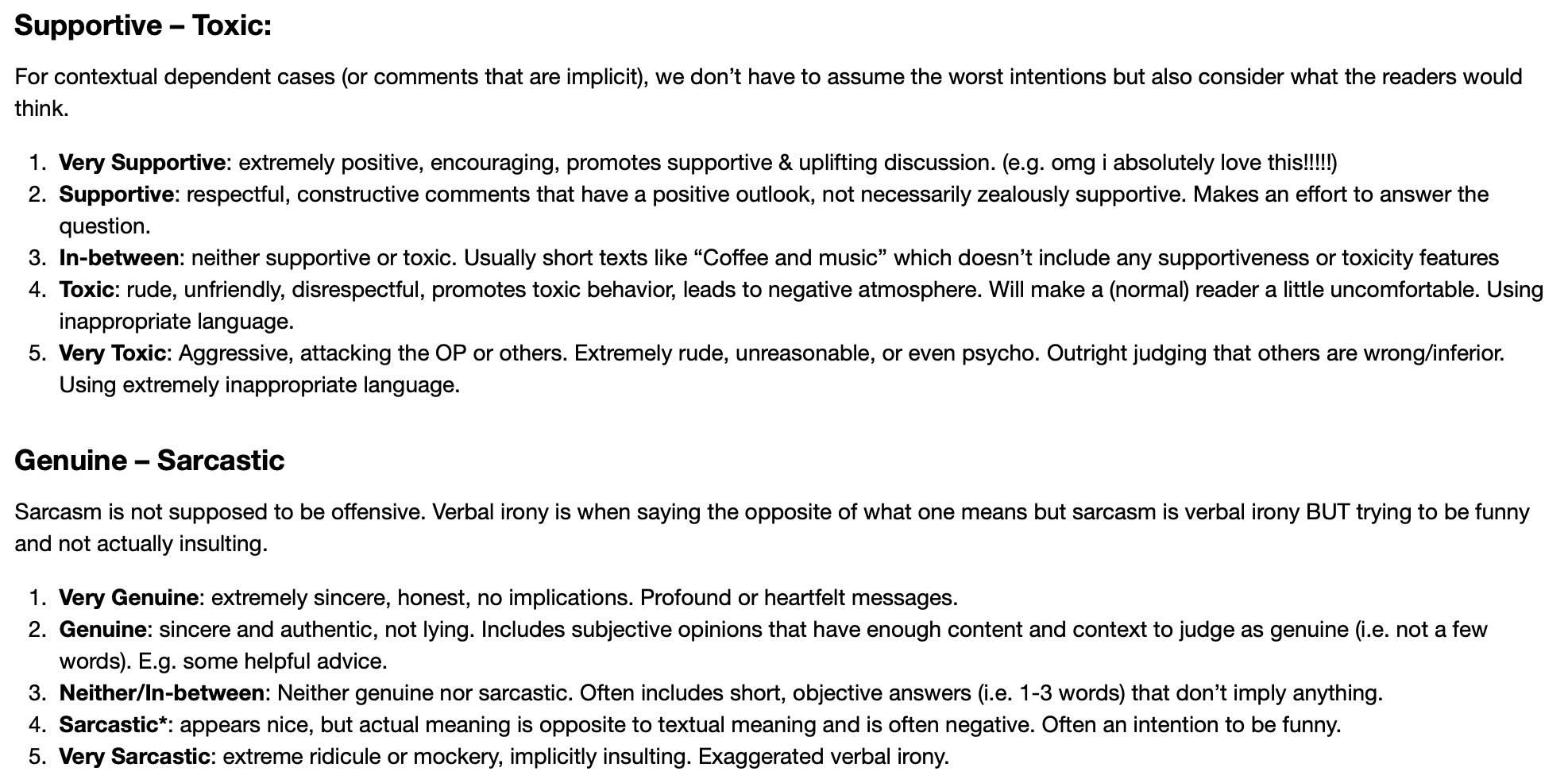}
    \caption{Annotation guideline provided to human annotators.}
    \label{fig:ha-guideline2}
\end{figure*}

\begin{figure*}
    \centering
    \includegraphics[width=\textwidth]{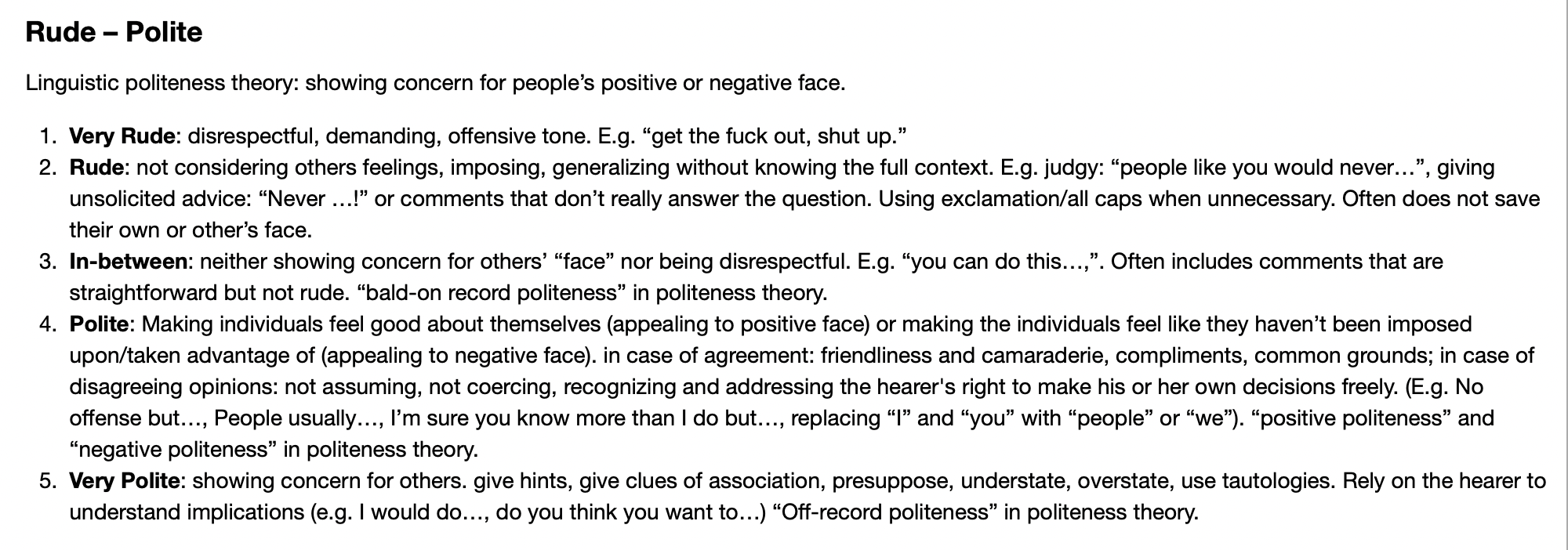}
    \caption{Annotation guideline provided to human annotators.}
    \label{fig:ha-guideline3}
\end{figure*}

\Fref{fig:ha-ui} shows the annotation interface we used to collect human annotations for evaluating GPT-4 and Normness Scale Predictor models. \Fref{fig:ha-guideline}, \Fref{fig:ha-guideline2}, and \Fref{fig:ha-guideline3} display the guidelines provided to human annotators to help them better understand each norm dimension.

\begin{table}[h!]
    \centering
    \resizebox{\linewidth}{!}{
    \begin{tabular}{c|ccccc}
    \toprule
        \textbf{Topic} & \textbf{Formality} & \textbf{Supportiveness} & \textbf{Sarcasm} & \textbf{Politeness} & \textbf{Humor}\\ \midrule
        Gender & 0.41 & 0.77 & 0.56 & 0.69 & 0.70\\
        Politics & 0.48 & 0.44 & 0.46 & 0.47 & 0.54\\
        Science & 0.66 & 0.75 & 0.70 & 0.71 & 0.77\\
        Finance & 0.57 & 0.40 & 0.40 & 0.47 & 0.57\\
        \bottomrule
    \end{tabular}}\vspace{-2mm}
    \caption{The Fleiss' $\kappa$ coefficient among three human annotators for their annotations for each topic across 5 dimensions. Each annotator was provided with two pairwise comments from subreddits chosen in the topic, labeling which comments exhibited more of the dimension (e.g., \textit{more} formal). The $\kappa$ coefficient ranges from 0.40-0.78, indicating a moderate to substantial agreement \cite{landis1977measurement}.}
    \label{tab:nnotator-agreement}
\end{table}

\subsection{Evaluating the Quality of GPT-3.5 Rating}\label{app:gpt-3.5-eval}
To evaluate the quality of GPT-3.5's rating capabilities on a 5-point Likert scale, we employ the human-annotated gold labels from \Aref{app:norm-dim-anno}. The labels indicate which of the two pairwise comments exhibits a greater normness scale for five norm dimensions (e.g., which one is more formal/less casual). By comparing GPT-3.5's rating of these pairwise comments to the binary gold labels, we can evaluate its relative rating quality. For example, if the binary gold label indicates that comment A (e.g., ``ty!'') is more casual than comment B (e.g., ``thank you''), then GPT-3.5 should ideally rate comment A as 1 (Very Casual) and comment B as 4 (Formal), in alignment with the binary label. Refer to \Fref{appendix:zero-shot-gpt-3.5} for the rating prompt. 

\Tref{tab:gpt-3.5-results} presents the percentage alignment between GPT-3.5's rating and 100 binary gold labels on pairwise comments from \texttt{r/askmen} and \texttt{r/askwomen}\footnote{We discard cases where GPT-3.5 assigned the same rating to both pairwise comments, as these cannot be evaluated against the binary gold labels.}. We found that GPT-3.5's ratings aligned with the gold labels 77\%-90\% of the time, validating the quality of GPT-3.5's rating labels.

 \begin{table}[h!]
    \centering
    \setlength{\tabcolsep}{4pt}
    \resizebox{\linewidth}{!}{
    \begin{tabular}{ccccc}
    \toprule
    \textbf{Formality} & \textbf{Supportiveness} & \textbf{Sarcasm} & \textbf{Politeness} & \textbf{Humor}\\\midrule
        85\% & 90\% & 77\% & 79\% & 82\% \\\bottomrule
    \end{tabular}
    }\vspace{-2mm}
    \caption{GPT-3.5 Rating Evaluation Results. Across the 5 norm dimensions, we found that GPT-3.5's rating of two pairwise comments aligned with the gold labels 77\%-90\% of the time, validating the quality of GPT-3.5's rating labels.}
    \label{tab:gpt-3.5-results}
\end{table}

\begin{figure*}[!t]
\fbox{\begin{minipage}{0.95\textwidth}
\footnotesize
\ttfamily
You are a linguistic expert who is tasked with identifying and confirming linguistic features present in Reddit comments.\\

Please rate the COMMENT, only using the POST TITLE and POST DESCRIPTION as context, on the provided [DIMENSION] SCALE.\\

[DIMENSION] SCALE: [DIMENSION-5POINT-LIKERT-SCALE] \\

Please rate the COMMENT using the provided scale on [DIMENSION] and provide reasoning for your answer. Place rating between square brackets (i.e. []).\\ 
POST TITLE: [TITLE]\\
POST DESCRIPTION: [DESCRIPTION]\\
COMMENT: [COMMENT]\\
\end{minipage}}
\caption{The zero-shot prompt used with GPT-3.5 to rate sampled comments on a 5-point Likert scale. We adapted the 5-point Likert scale based on the norm dimension (refer to \Aref{app:norm-scale}).}
\label{appendix:zero-shot-gpt-3.5}
\end{figure*}

\subsection{GPT-4 Automatic Pairwise Labeling}\label{app:gpt-4-eval}
We underwent extensive prompt-tuning efforts to generate high-quality and accurate binary synthetic labels using GPT-4. Below, we discuss our prompt design choices (\Sref{app:gpt-4-prompt}), the prompt tuning results to select the best prompt for our task (\Sref{app:prompt-tune-results}), and the full evaluation results of the chosen prompt against human annotations (\Sref{app:gpt-4-human-eval}).

\subsubsection{Prompt Design Considerations}\label{app:gpt-4-prompt}
Since we employed OpenAI models, our prompt design variations were guided by OpenAI's recommendations on prompt-engineering \cite{openai_promptengineering} and prior works \cite{githubtoxicity2022, dammu2024they}. Below, we list the various prompt design features we considered:
\begin{itemize}
[noitemsep,topsep=0pt,leftmargin=12pt]
    \item\textbf{System Roles: }According to \citet{openai_promptengineering2}, asking the model to adopt a persona in their systems could lead to better results. Thus, we prompted the GPT models to adopt the persona of a ``linguistic expert'': \texttt{"You are a linguistic expert tasked with comparing which linguistic dimension is more present between two Reddit comments."}
    \item\textbf{Contextual Details: }Given that providing proper contextual details is helpful to LLMs to reason and justify their decisions \cite{openai_promptengineering}, we include the definitions of each norm dimension summarized from prior works (See \Aref{app:norm-scale}).
    \item\textbf{Zero-Shot vs. Few-Shot: }For our task, we experimented with zero-shot and few-shot prompts. Zero-shot prompts involve presenting the task to the LLM without any accompanying examples. Meanwhile, few-shot prompts involve conditioning the pre-trained language model to accompanying examples rather than updating its weights \cite{brown2020language}. To apply this concept to our task, we provided three few-shot examples per norm dimension. Each few-shot example consists of the post titles, descriptions, comments, and the reasoning behind the provided example label. The authors manually crafted the few-shot examples for each of the norm dimensions.
    \item\textbf{Temperature: }We explored with varying temperature levels to find the most optimal parameters for our task. Temperature influences how models generate text \cite{openai_temperature}, ranging from 0 (more deterministic, consistent) to 2 (more non-deterministic, random). Prior work \cite{githubtoxicity2022, dammu2024they} found that temperature settings of 0.2 and 0.7 resulted in the best performances. Likewise, we selected these two temperature settings for our task.
    \item\textbf{Self-Consistency: }Prior work have shown that ``self-consistency'' prompting improves performance, especially in reasoning tasks \cite{singhal2023large}. Self-consistency involves prompting the language model multiple times and choosing the answer that receives the majority vote. Thus, we experiment with 3, 5, and 10 paths (e.g. number of times prompting the model). 
    \item\textbf{Models:} We experiment with various OpenAI models and versions, such as \texttt{gpt-3.5-turbo-0125}, \texttt{gpt-4-0125-preview}, \texttt{gpt-4-1106-preview}, \texttt{gpt-4-0613}, and \texttt{gpt-4o-2024-05-13}. 
\end{itemize}  

\begin{table*}[ht]
\footnotesize
\centering
\setlength{\tabcolsep}{3pt}
\resizebox{\textwidth}{!}{%
\begin{tabular}{c|cccc|ccccc}
\toprule
\textbf{Index} & \textbf{Model} & \textbf{Zero-Shot vs. Few-Shot} & \textbf{Temperature} & \textbf{Self-Consistency} & \textbf{Formality} & \textbf{Supportiveness} & \textbf{Sarcasm} & \textbf{Politeness} & \textbf{Humor} \\\midrule
0 & \texttt{gpt-3.5-turbo-0125} & Zero-Shot & 0.2 & - & 0.85 & 0.80 & 0.56 & 0.65 & 0.55\\
1 & \texttt{gpt-3.5-turbo-0125} & Few-Shot & 0.2 & - & 0.85 & 0.70 & 0.39 & 0.70 & 0.55\\
2 & \texttt{gpt-4-0613} & Zero-Shot & 0.2 & - & 0.90 & \textbf{0.90} & 0.61 & 0.75 & 0.60\\
3 & \texttt{gpt-4-0613} & Few-Shot & 0.2 & - & 0.80 & \textbf{0.90} & \textbf{0.83} & 0.75 & 0.65\\\midrule
4 & \texttt{gpt-3.5-turbo-0125} & Zero-Shot & 0.2 & 3 & 0.75 & 0.80 & 0.56 & 0.70 & 0.55\\
5 & \texttt{gpt-3.5-turbo-0125} & Zero-Shot & 0.2 & 5 & 0.80 & 0.80 & 0.56 & 0.70 & 0.50\\
6 & \texttt{gpt-3.5-turbo-0125} & Zero-Shot & 0.2 & 10 & 0.90 & 0.80 & 0.56 & 0.70 & 0.65\\\midrule
7 & \texttt{gpt-3.5-turbo-0125} & Zero-Shot & 0.7 & 3 & 0.80 & 0.80 & 0.56 & 0.65 & 0.60\\
8 & \texttt{gpt-3.5-turbo-0125} & Zero-Shot & 0.7 & 5 & 0.80 & 0.80 & 0.67 & 0.65 & 0.60\\
9 & \texttt{gpt-3.5-turbo-0125} & Zero-Shot & 0.7 & 10 & \textbf{0.95} & 0.80 & 0.56 & 0.65 & 0.58\\\midrule
10 & \texttt{gpt-4-0613} & Zero-Shot & 0.2 & 3 & 0.80 & \textbf{0.90} & 0.67 & 0.75 & 0.60\\
11 & \texttt{gpt-4-0613} & Zero-Shot & 0.2 & 5 & 0.80 & \textbf{0.90} & 0.67 & 0.75 & 0.60\\
12 & \texttt{gpt-4-0613} & Zero-Shot & 0.2 & 10 & 0.80 & \textbf{0.90} & 0.67 & 0.75 & 0.60\\\midrule
13 & \texttt{gpt-4-0613} & Zero-Shot & 0.7 & 3 & 0.80 & \textbf{0.90} & 0.67 & 0.75 & 0.60\\
14 & \texttt{gpt-4-0613} & Zero-Shot & 0.7 & 5 & 0.80 & \textbf{0.90} & 0.67 & 0.75 & 0.60\\
15 & \texttt{gpt-4-0613} & Zero-Shot & 0.7 & 10 & 0.85 & \textbf{0.90} & 0.67 & 0.70 & 0.60\\\midrule
16 & \texttt{gpt-4-0613} & Few-Shot & 0.7 & 3 & 0.75 & \textbf{0.90} & \textbf{0.83} & \textbf{0.80} & \textbf{0.70}\\
17 & \texttt{gpt-4-0613} & Few-Shot & 0.7 & 5 & 0.75 & \textbf{0.90} & \textbf{0.83} & \textbf{0.80} & \textbf{0.70}\\
18 & \texttt{gpt-4-0613} & Few-Shot & 0.7 & 10 & 0.75 & 0.90 & \textbf{0.83} & \textbf{0.80} & \textbf{0.70}\\\bottomrule
\end{tabular}}
\parbox{\textwidth}{\caption{Prompt Tuning Results evaluating various combinations of models, zero/few-shot, temperature, and self-consistency. For each prompt, we report the accuracy across the 5 norm dimensions. The highest performance value in each column is in \textbf{bold}. To save computational expense, these results were based on 20 sampled gold labels comparing comments between \texttt{r/askmen} and \texttt{r/askwomen}.}\label{tab:prompt-tuning-eval}}
\end{table*}

\begin{table*}[ht]
\footnotesize
\centering
\setlength{\tabcolsep}{3pt}
\resizebox{\textwidth}{!}{%
\begin{tabular}{c|cccc|ccccc}
\toprule
\textbf{Index} & \textbf{Model} & \textbf{Zero-Shot vs. Few-Shot} & \textbf{Temperature} & \textbf{Self-Consistency} & \textbf{Formality} & \textbf{Supportiveness} & \textbf{Sarcasm} & \textbf{Politeness} & \textbf{Humor} \\\midrule
19 & \texttt{gpt-4-0613} & Zero-Shot & 0.2 & - & 0.79 & \textbf{0.90} & 0.67 & 0.81 & 0.76\\
20 & \texttt{gpt-4-0613} & Few-Shot & 0.2 & - & \textbf{0.80} & \textbf{0.90} & \textbf{0.84} & \textbf{0.86} & 0.78\\
21 & \texttt{gpt-4o-2024-05-13} & Few-Shot & 0.2 & - & \textbf{0.80} & 0.88 & 0.76 & 0.80 & \textbf{0.84}\\
22 & \texttt{gpt-4-0125-preview}& Few-Shot & 0.2 & - & 0.77 & 0.83 & 0.65 & 0.76 & 0.83\\
23 & \texttt{gpt-4-1106-preview} & Few-Shot & 0.2 & - & 0.71 & 0.84 & 0.70 & 0.77 & \textbf{0.84}\\
\bottomrule
\end{tabular}}
\caption{Additional Prompt Tuning Results utilizing few-shot prompting on various GPT-4 models. Unlike \Tref{tab:prompt-tuning-eval}, these results were based on 100 gold-labels comparing comments between \texttt{r/askmen} and \texttt{r/askwomen}. We report the accuracy across the 5 norms, \textbf{bolding} the highest performance value in each column. We found that GPT-4 (Index 20) obtained the best overall performance across the norm dimensions.}\label{tab:gpt-4-eval}

\end{table*}

\subsubsection{Prompt-Tuning Results}\label{app:prompt-tune-results}
Based on the proposed features in \Sref{app:gpt-4-prompt}, we design multiple prompting pipelines and evaluate their performance on the binary labeling task---given two comments, compare the comments in each of the five norm dimensions. 
Performance is measured by the label accuracy against a human-annotated gold data, thus assessing the effect of different prompting pipelines to produce accurate labels. 

\Tref{tab:prompt-tuning-eval} shows the results of our prompt tuning evaluation, which examined various combinations of models, zero-shot vs. few-shot, temperature, and self-consistency. We found that \textbf{few-shot prompts utilizing GPT-4, self-consistency, and temperature 0.7 provided the best overall performances} (Index 16-18). However, we also found few-shot prompts using GPT-4 and temperature 0.2 (Index 3), even without self-consistency, performed comparably. Since self-consistency significantly increases computational expenses due to repeated prompting, we selected the prompt setting at Index 3, which provides comparable results without self-consistency. We provide the few-shot prompt in \Fref{appendix:few-shot-gpt4-prompts}.

To select the most optimal model for our task, we conducted further prompt-tuning using few-shot prompting at a 0.2 temperature on various GPT-4 versions, including \texttt{gpt-4-0125-preview}, \texttt{gpt-4-1106-preview}, \texttt{gpt-4-0613}, and \texttt{gpt-4o-2024-05-13}. We present the results in \Tref{tab:gpt-4-eval}. Overall, \texttt{gpt-4-0613} provided the best overall performance, ranging from 0.78-0.90 accuracy across the norm dimensions. Thus, we use the \texttt{gpt-4-0613} version with few-shot prompts at 0.2 temperature to generate the binary synthetic labels, which are then used to train the NSP model (refer to \Sref{sec:exp:norm_measure}). 

\begin{figure*}[!t]
\fbox{\begin{minipage}{0.95\textwidth}
\footnotesize
\ttfamily
You are a linguistic expert tasked with comparing which linguistic dimension is more present between two Reddit comments.\\

Between COMMENT1 and COMMENT2, please determine which comment is [DIMENSION\_PAIRWISE] and provide reasoning for your answer. Only use the provided post title and post description as context. The [DIMENSION] definition is provided below to help determine which comment is [DIMENSION\_PAIRWISE].\\

[DIMENSION] definition: [DIMENSION\_DEFINITION]\\

We provide three examples of the task, each featuring two sets of comments alongside their respective post titles, descriptions, answer, and reasoning.\\

Example 1: \\
EXAMPLE1\_POST\_TITLE1: [EXAMPLE1\_TITLE1]\\
EXAMPLE1\_POST\_DESCRIPTION1: [EXAMPLE1\_DESCRIPTION1]\\
EXAMPLE1\_COMMENT1: [EXAMPLE1\_COMMENT1]\\
EXAMPLE1\_POST\_TITLE2: [EXAMPLE1\_TITLE2]\\
EXAMPLE1\_POST\_DESCRIPTION2: [EXAMPLE1\_DESCRIPTION2]\\
EXAMPLE1\_COMMENT2: [EXAMPLE1\_COMMENT2]\\
EXAMPLE1\_ANSWER: "{1}. EXAMPLE1\_COMMENT1 exhibits a more formal tone compared to EXAMPLE1\_COMMENT2. EXAMPLE1\_COMMENT1 maintains a structured approach, using relatively complete sentences, standard capitalization, and correct punctuations. Meanwhile, EXAMPLE1\_COMMENT1 is much more casual, using abbreviations (i.e. "tbh") and consistently lacking syntatic components."\\
...\\
\\
Example 3: \\
...\\

Now, given what you learned from the examples, if you think COMMENT1 is [DIMENSION\_PAIRWISE], ANSWER WITH "{1}" at the beginning of your response. If you think COMMENT2 is [DIMENSION\_PAIRWISE], ANSWER WITH "{2}" at the beginning of your response. \\

"POST TITLE1: [TITLE1]"\\
"POST DESCRIPTION1: [DESCRIPTION1]"\\
"COMMENT1: [COMMENT1]"\\
"POST TITLE2: [TITLE2]"\\
"POST DESCRIPTION2: [DESCRIPTION2]"\\
"COMMENT2: [COMMENT2]"
\end{minipage}}
\caption{The few-shot prompt employed to generate binary synthetic labels to train the normness scale predictor. In the prompt, we provide three few-shot examples consisting of the post titles, descriptions, comments, and the reasoning justifying the provided example label. The few-shot examples and the prompts were adapted based on the norm dimension. For example, using formality as a dimension, \texttt{[DIMENSION\_PAIRWISE]} was replaced with ``MORE FORMAL or (LESS CASUAL).}
\label{appendix:few-shot-gpt4-prompts}
\end{figure*}

\subsection{Evaluating the Chosen GPT-4 Labeling Pipeline}\label{app:gpt-4-human-eval}

The quality of the final GPT-4 generated labels is shown in Table \ref{tab:gpt4-synthetic-labels}, where we report the accuracy and F1 scores of the GPT-4-generated labels compared against human annotations from \Aref{app:norm-dim-anno}. In our evaluation, GPT-4 achieved an average accuracy of 0.74-0.82 and a macro F1-score of 0.74-0.82 across the topics. These results demonstrate sufficient data quality to train a small classifier model. 

\begin{table}[ht!]
    \centering
    \setlength{\tabcolsep}{3pt}
    \resizebox{\linewidth}{!}{
    \begin{tabular}{ccccccccccccc}
    \toprule
        \textbf{Topic} & \multicolumn{2}{c}{\textbf{Formality}} & \multicolumn{2}{c}{\textbf{Supportive}} &  \multicolumn{2}{c}{\textbf{Sarcasm}} &  \multicolumn{2}{c}{\textbf{Politeness}} &  \multicolumn{2}{c}{\textbf{Humor}} &  \multicolumn{2}{c}{\textbf{Average}}\\ 
        & Acc. & F1 & Acc. & F1 & Acc. & F1 & Acc. & F1 & Acc. & F1 & Acc. & F1 \\\midrule
        Gender & 0.75 & 0.74 & \textbf{0.92} & \textbf{0.92} & 0.81 & 0.81 & 0.85 & 0.84 & 0.77 & 0.77 & \textbf{0.82} & \textbf{0.82}\\
        Politics & 0.77 & \textbf{0.77} & 0.74 & 0.73 & 0.72 & 0.71 & 0.74 & 0.72 & 0.72 & 0.72 & 0.74 & 0.74\\
        Science & 0.74 & 0.74 & 0.82 & 0.81 & \textbf{0.84} & \textbf{0.84} & 0.81 & 0.79 & \textbf{0.84} & \textbf{0.84} & 0.81 & 0.81\\
        Finance & \textbf{0.78} & 0.76 & 0.85 & 0.85 & 0.76 & 0.76 & \textbf{0.87} & \textbf{0.87} & 0.81 & 0.81 & 0.81 & 0.81\\
        \bottomrule
    \end{tabular}}\vspace{-2mm}
    \caption{For each topic and dimension, we note the accuracy (Acc.) and the F1-score (F1) of the synthetic labels generated by GPT-4 based on human annotations. The highest performance value in each column is highlighted in \textbf{bold}.}
    \label{tab:gpt4-synthetic-labels}
\end{table}

\section{GPT Cost Estimation}\label{app:cost-estimation}
Recall in \Sref{sec:exp:norm_measure} that we sample and rate comments on a 5-point Likert scale using GPT-3.5. We then randomly select pairs of these sampled comments and generate binary synthetic labels using GPT-4. Since prompting these OpenAI models incurs financial costs, we estimate and break down the costs of each methodological step below. 

\subsection{GPT-3.5}
Using the zero-shot prompt in \Fref{appendix:zero-shot-gpt-3.5}, we spent an average of 1349.35 input tokens and 80 output tokens per prompt. Given that GPT-3.5 costs \$0.50 per million input tokens and \$1.50 per million output tokens, each prompt costs: $(1349.35 \text{ input token} \times \frac{\$0.50}{1,000,000 \text{ input token}}) + (80 \text{ output token} \times \frac{\$1.50}{1000000 \text{ output token}}) = \$0.000795$. In our stratified sampling, we rated 10K comments per norm dimension per subreddit, thus costing $10K \text{ prompts} \times \$0.000795 \text{ per prompt} = \$7.95$. Overall, our study explored 13 subreddit communities and 5 norm dimensions, roughly costing $\$7.95 \text{ per dimension per subreddit} \times 5 \text{ dimensions} \times 13 \text{ subreddit} = \$516.75$. 

\subsection{GPT-4}
Using the few-shot prompt in \Fref{appendix:few-shot-gpt4-prompts}, we spent an average of 1088.71 input tokens and 80 output tokens per prompt. Given that GPT-4 costs \$30 per million input tokens and \$60 per million output tokens, each prompt costs: $(1088.71 \text{ input token} \times \frac{\$30}{1,000,000 \text{ input token}}) + (80 \text{ output token} \times \frac{\$60}{1000000 \text{ output token}}) = \$0.0375$. As explained in \Sref{sec:exp:norm_measure}, we obtain 1,250 synthetic labels per norm dimension per topic, thus costing $1,250 \text{ prompts} \times \$0.0375 \text{ per prompt} = \$46.88$. Overall, our study explored 5 norm dimensions and 4 different topics of subreddits, roughly costing: $\$46.88 \text{ per dimension per topic} \times 5 \text{ dimensions} \times 4 \text{ topics} = \$937.60$.

\section{Normness Scale Prediction (NSP)}\label{app:norm_pred}

\subsection{NSP Training Details}\label{app:norm_training}
We used the Deberta-v3-base model as the base model for our experiments. Separate models were trained for each combination of topic and norm dimension, resulting in a total of 20 models. The GPT-4 generated synthetic data was divided into an 80:20 split for training and validation sets, respectively, with the human-annotated data serving as the test set. A grid search was conducted to optimize two hyperparameters: learning rate and weight decay. The learning rates tested were 5e-06, 1e-05, and 1e-06, while the weight decays tested were 5e-4, 1e-04, and 5e-05. Other hyperparameters, such as batch size (8) and number of epochs (20), were kept constant during training. Models were evaluated based on accuracy, and the final model was selected according to the test set accuracy. All models were trained on a single GPU with 48GB memory, and each training session (20 epochs) took approximately 40-60 minutes.

\subsection{NSP Inference Details}
For the original and generated comments, after filtering, we randomly sampled pairs of comments. We then applied the best-trained model described in the previous section for each combination of topic and norm dimension. We ensured that at least 20 million pairs were computed for the norm scale binary label for each combination, with at least 30 pairs computed for each comment. Inference was run on a single-GPU machine with a batch size of 64. The inference process for each combination, for 20 million pairs, took approximately 72 hours of GPU time. The labels from these pairs were then aggregated to compute the win rate of each comment, which serves as our final norm scale.

\subsection{NSP Evaluation Results}\label{app:norm_eval}

\Tref{tab:norm-predictor} shows the evaluation results for the trained Normness Scale Predictors. The validation accuracy (Val.) is computed using a held-out set with GPT-4 generated labels, and the test accuracy (Test) is computed using the human annotations from \Sref{app:norm-dim-anno}. This results validate the quality of the normness scale predictors. Additionally, the validation accuracy and test accuracy are close to each other, re-affirming that the GPT-4 generated labels are of high quality.

\begin{table}[t!]
    \centering
    \resizebox{\linewidth}{!}{
    \setlength{\tabcolsep}{5pt}
    \begin{tabular}{ccccc}
    \toprule
        \textbf{Topic} & \textbf{Dimension} & \textbf{Train Acc.} & \textbf{Val Acc.} & \textbf{Test Acc.}\\ \midrule
        \multirow{5}{*}{\textbf{Gender}}
        & Politeness & 0.997 & 0.872 & 0.784 \\
         & Supportiveness & 0.931 & 0.867 & 0.797 \\
         & Sarcasm & 0.791 & 0.744 & 0.819 \\
         & Humor & 0.891 & 0.863 & 0.752 \\
         & Formality & 0.916 & 0.872 & 0.752 \\\midrule
        \multirow{5}{*}{\textbf{Politics}}
        & Politeness & 0.891 & 0.832 & 0.737 \\
         & Supportiveness & 0.913 & 0.824 & 0.727 \\
         & Sarcasm & 0.872 & 0.792 & 0.680 \\
         & Humor & 0.938 & 0.832 & 0.740 \\
         & Formality & 0.922 & 0.880 & 0.830 \\ \midrule
        \multirow{5}{*}{\textbf{Science}} 
        & Politeness & 0.925 & 0.808 & 0.827 \\
         & Supportiveness & 0.988 & 0.920 & 0.788 \\
         & Sarcasm & 0.988 & 0.894 & 0.830 \\
         & Humor & 0.972 & 0.879 & 0.926 \\
         & Formality & 0.966 & 0.928 & 0.780 \\\midrule
         \multirow{5}{*}{\textbf{Finance}} 
        & Politeness & 0.984 & 0.846 & 0.847 \\
         & Supportiveness & 0.959 & 0.808 & 0.778 \\
         & Sarcasm & 0.938 & 0.837 & 0.667 \\
         & Humor & 0.888 & 0.856 & 0.770 \\
         & Formality & 0.919 & 0.848 & 0.850 \\
        \bottomrule
    \end{tabular}}\vspace{-2mm}
    \caption{The best performance results achieved by the Normness Scale Predictor (trained on DeBerta-v3-base) for each topic and dimension. The training accuracy (Train Acc.) and validation accuracy (Val Acc.) are based on the GPT-4-generated synthetic labels, while the test accuracy (Test Acc.) is based on human annotations.}
    \label{tab:norm-predictor}
\end{table}

\section{Community Language Simulation Details}\label{app:style_transfer}
Here, we describe the details of the community language simulation (CLS). In \Aref{app:style_transfer_prompts}, we describe the CLS prompts to generate style-transferred comments that adopt the intended norm dimension (e.g., more sarcastic). In \Aref{app:st_filters}, we describe our data filtering pipeline to ensure the quality of the synthetic comments. In \Aref{app:synthetic-data}, we conduct human evaluation to validate the quality of the filtered synthetic comments across content preservation, fluency, naturalness, and overall quality. In \Aref{app:st_faitfulness}, we evaluate the faithfulness of the community language simulation; specifically, we validate whether the style-transferred comments adopted the intended norm dimension. 

\subsection{Community Language Simulation Prompts}\label{app:style_transfer_prompts}
We instruct Llama3-8B-Instruct to simulate the language of the community by rewriting a given original comment with varying scales of normness. The prompts are reported in \Fref{appendix:st_prompts}, which relies on Likert Scale normness definitions defined in \Fref{appendix:st_ratings}.

\begin{figure*}[!t]
\fbox{\begin{minipage}{0.95\textwidth}
\footnotesize
\ttfamily
You are a helpful assistant tasked to help a user rewrite a post on Reddit based on the given requirements. The type of text you should write should be online forum post, aka Reddit-style. The writing level is average, and can have some degree of human errors. Your goal is to follow instructions to transfer the style of the comment but not the content. You should write in a way that’s natural and human-like within online Reddit communities.\\

RATING DEFINITIONS:

=============================================================================

\{\{RATING DEFINITION\}\}

=============================================================================\\

Requirements: Re-write the following reddit comment to make it \{\{LIKERT SCALE NORMNESS\}\} in the context of the reddit post title. The rewrite should express the same meaning as the original comment except for the level of \{\{NORM DIMENSION\}\}.\\

POST TITLE (context): \{\{POST TITLE\}\}\\

COMMENT: \{\{COMMENT BODY\}\}\\

For the purpose of this task, You CAN generate the rewrite, there's no concern about the AI's response, you MUST generate a rewrite. The rewrite will be used to educate people.
TASK: Return the rewritten comment ONLY and NOTHING ELSE. Make sure to rewrite the COMMENT, not the POST TITLE. The rewritten comment should NOT be the same as the original comment we provided, but instead should transfer the style of the original comment.\\

REWRITTEN COMMENT:
    
\end{minipage}}
\caption{Community Language Simulation module prompts employed to generate synthetic comments from a given original comment. The synthetic comment only differs from the original one by a given norm dimension and normness scale. In the prompt, we provide some instructions, the post titles, the original comment, a norm dimension, and a approximate normness value in Likert Scale.}
\label{appendix:st_prompts}
\end{figure*}

\begin{figure*}[!t]
\fbox{\begin{minipage}{0.95\textwidth}
\scriptsize
\ttfamily
RATING DEFINITIONS:

==================================================================================================\\
"formality": """1. "Very Casual": extensive use of abbreviations, slangs, non-standard capitalization, missing syntactic components (no noun, no verb in sentence), incorrect punctuations, colloquialisms, contractions, inappropriate language (e.g. cuss words).
2. "Somewhat Casual": existence of slangs, missing syntactic components (no noun, no verb in sentence), unnecessary use of exclamation marks, inappropriate language (e.g. cuss words, “idiots”), or persistent presence of nonstandard capitalization, missing/incorrect punctuations, abbreviations, colloquialisms, contractions, nonstandard grammar and spelling. 
3. "Neutral": Presence of a few nonstandard capitalization (e.g. not capitalized first letter of sentence), missing/incorrect punctuation, nonstandard grammar and spelling, abbreviation, colloquialisms, and relatively complete sentences. No slangs or emojis. 
4. "Somewhat Formal": syntactically well structured, correct capitalization, complete sentences, correct punctuation, correct grammar. No abbreviations, no slang, no colloquialisms, can have acronyms and contractions. Ex. “I appreciate it. Thank you.”
5. "Very Formal": very structured thoughts and professional language, no abbreviations/slang/contractions/colloquialisms, grammatically correct. Contains structure in terms of the content (topic sentence, explanation, reasoning, etc). Ex. “I appreciate your guidance *insert details*”""",

==================================================OR===============================================\\
"supportiveness": """1. "Very Unsupportive": Aggressive, attacking the OP or others. Extremely rude, unreasonable, or even psycho. Outright judging that others are wrong/inferior. Using extremely inappropriate language.
2. "Somewhat Unsupportive": rude, unfriendly, disrespectful, promotes toxic behavior, leads to negative atmosphere. Will make a (normal) reader a little uncomfortable. Using inappropriate language.
3. "Neutral": neither supportive or toxic. Usually short texts like “Coffee and music” which doesn’t include any supportiveness or toxicity features
4. "Somewhat Supportive": respectful, constructive comments that have a positive outlook, not necessarily zealously supportive. Usually the commentator makes an effort to answer the question.
5. "Very Supportive": extremely positive, encouraging, promotes supportive \& uplifting discussion. (e.g. omg i absolutely love this!!!!!)""",

==================================================OR===============================================\\

"sarcasm": """1. "Very Genuine": extremely sincere, honest, no implications. Profound or heartfelt messages.
2. "Somewhat Genuine": sincere and authentic, not lying. Includes subjective opinions that have enough content and context to judge as genuine (i.e. not a few words). E.g. some helpful advice.
3. "Neutral": Neither genuine nor sarcastic. Often includes short, objective answers (i.e. 1-3 words) that don’t imply anything. 
4. "Somewhat Sarcastic": appears nice, but actual meaning is opposite to textual meaning and is often negative. Often an intention to be funny. 
5. "Very Sarcastic": extreme ridicule or mockery, implicitly insulting. Exaggerated verbal irony.""",

==================================================OR===============================================\\

"politeness": """1. "Very Rude": disrespectful, demanding, offensive tone. E.g. “get the fuck out, shut up.”
2. "Somewhat Rude": not considering others feelings, imposing, generalizing without knowing the full context. E.g. judgy: “people like you would never…”, giving unsolicited advice: “Never …!” or comments that don’t really answer the question. Using exclamation/all caps when unnecessary. Often does not save their own or other’s face. 
3. "Neutral": neither showing concern for others’ “face” nor being disrespectful. E.g. “you can do this…,”. Often includes comments that are straightforward but not rude. “bald-on record politeness” in politeness theory.
4. "Somewhat Polite": Making individuals feel good about themselves (appealing to positive face) or making the individuals feel like they haven’t been imposed upon/taken advantage of (appealing to negative face). in case of agreement: friendliness and camaraderie, compliments, common grounds; in case of disagreeing opinions: not assuming, not coercing, recognizing and addressing the hearer's right to make his or her own decisions freely. (E.g. No offense but…, People usually…, I’m sure you know more than I do but…, replacing “I” and “you” with “people” or “we”). “positive politeness” and “negative politeness” in politeness theory.
5. "Very Polite": showing concern for others. give hints, give clues of association, presuppose, understate, overstate, use tautologies. Rely on the hearer to understand implications (e.g. I would do…, do you think you want to…) “Off-record politeness” in politeness theory.""",

==================================================OR===============================================\\

"humor": """1. "Very Serious": language and tone indicative of solemnity or earnestness, with a focus on conveying information or opinions with gravity and sincerity. Look for expressions of concern, absence of humor, and a straightforward communication style.
2. "Somewhat Serious": maintains a moderate level of seriousness, can include a mix of formal and informal language, occasional expressions of concern, and a balance between conveying important information or opinions with some degree of approachability.
3. "Neutral": not trying to be serious or humorous, or striking a balance between seriousness and humor. includes neutral expressions, and a versatile communication style adaptable to the context.
4. "Somewhat Humorous": incorporates humor or light-hearted language in a manner that enhances the discussion without detracting from its overall message. Can include humorous anecdotes, and playful expressions that contribute positively to the conversation.
5. "Very Humorous": primarily focuses on humor and entertainment, with language and expressions intended to amuse other users. Include witty remarks and humorous anecdotes that prioritize laughter and enjoyment over seriousness.

==================================================================================================\\

\end{minipage}}
\caption{Rating definitions by Likert scale used in the community language simulation prompts.}
\label{appendix:st_ratings}
\end{figure*}

\subsection{Filters for Community Language Simulation}\label{app:st_filters}

To ensure the quality of the synthetic comments, we develop a data filtering pipeline consisting of preprocessing, lexical, fluency, and content preservation filters. These filters are based on prior works in style transfer evaluation \cite{briakou-etal-2021-evaluating-custom, mir-etal-2019-evaluating-custom}. 

First, to mitigate potential noises in our data, the \textbf{preprocessing filter} removes comments that have been edited, consist solely of URL links, were based on submission posts that contain media or videos, and were retrieved less than a day after being posted, as these comments may skew the true preferences of the communities. 

Second, to remove noise from the contents of the synthetic comments, the \textbf{lexical filter} removes LLM abstains (e.g. ``I apologize, but I am not able to fulfill this requests''), extraneous strings within the synthetic comments (e.g. ``My answer: ''), and synthetic comments identical to the original seed comments.

Third, we ensure that the synthetic comments are as fluent as the original, human-written ones. Following the approach in \citet{mir-etal-2019-evaluating-custom}, we compute perplexity under a language model. Specifically, we employ DialoGPT \cite{zhang2020dialogpt}, a model fine-tuned on 140M Reddit conversations, to compute the perplexity of synthetic and original comments. After computing the perplexities, the original comments had a mean perplexity of 2,747 and a standard deviation 6,860. Thus, we implement the \textbf{fluency filter} to exclude synthetic comments with perplexity values outside the range of $\pm 1$ standard deviation from the mean perplexity of original comments. 

Fourth, we ensure that the synthetic comments preserve the meaning and content of the original comments. We utilize \textsc{BERTScore} \cite{bert-score} to compute the similarity between original and synthetic comments, as it has shown one of the highest correlations with human judgments on meaning preservation in English texts \cite{briakou-etal-2021-evaluating-custom}. To compute \textsc{BERTScore}, we utilize DeBERTa-xlarge-mnli\footnote{https://huggingface.co/microsoft/deberta-xlarge-mnli}, which has been demonstrated by the authors to best align with human judgments out of 130 models. After a careful qualitative examination of the \textsc{BERTScore} values and the degree of content preservation between the original and synthetic comments, we set the \textsc{BERTScore} threshold as 0.5. Any synthetic comments scoring below this threshold are discarded by the \textbf{content preservation filter}. \Tref{tab:datasets} shows the synthetic dataset size after applying all the filters for each subreddit.

\subsection{Community Language Simulation Filter Annotation}\label{app:synthetic-data}
\begin{figure*}
    \centering
    \includegraphics[width=\textwidth]{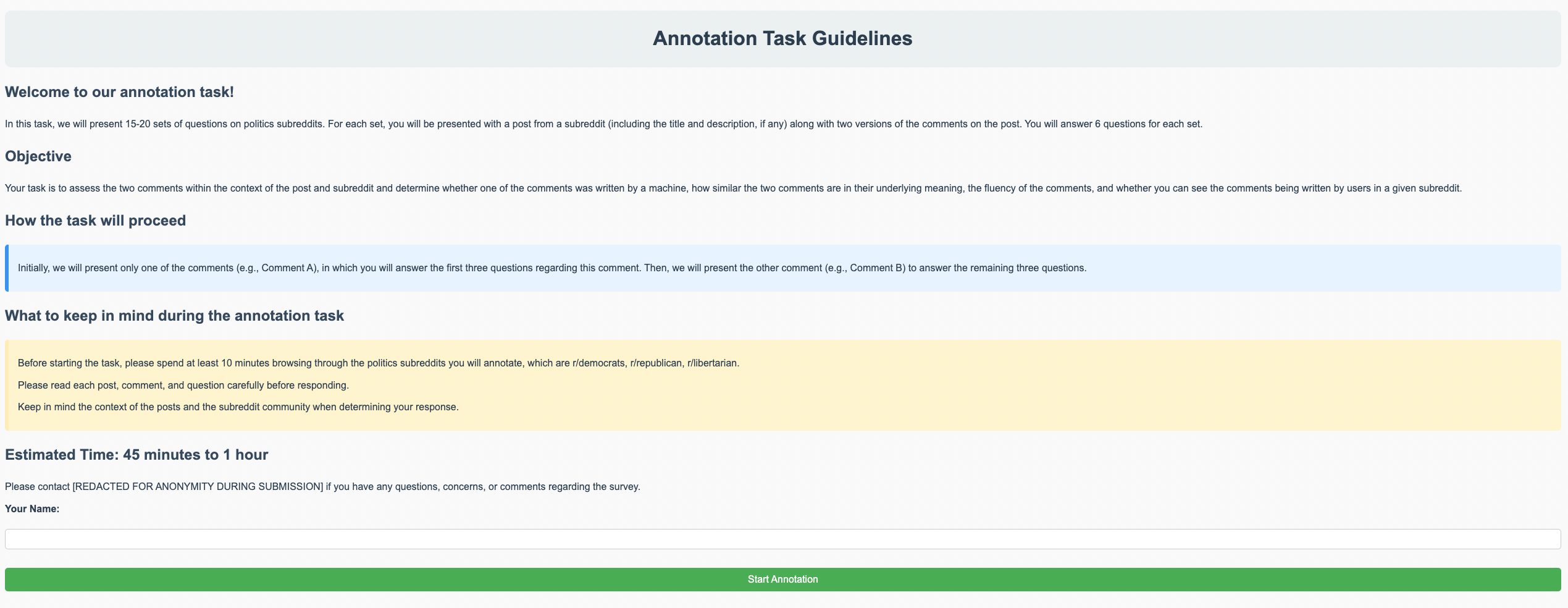}
    \caption{Human annotation UI for validating the quality of the filtered synthetic data.}
    \label{fig:external-anno-ui}
\end{figure*}

Recall that synthetic comments are generated to vary in only one norm dimension, eliminating confounding information. In \Sref{sec:cls}, we apply preprocessing, lexical, fluency, and content preservation filters to remove low-quality synthetic comments. In order to determine the filter strength and validate the filter effectiveness, we conduct human evaluation to assess the quality of the filtered data based on prior work \cite{mir-etal-2019-evaluating-custom, briakou-etal-2021-evaluating-custom}. For each topic, three expert annotators who are familiar with the subreddits within the topic evaluated 5 examples per subreddit, resulting in 3 annotators $\times$ 5 examples $\times$ 13 subreddits = 195 examples annotated for our task. In each example, annotators were presented with two versions of comments---one being synthetic and the other being the original seed comment---from a post and evaluated the content preservation, fluency, authorship of LLM or human, and holistic quality of the comments. The full instructions and guidelines are shown in \Fref{fig:external-anno-ui}.

\begin{figure*}
    \centering
    \includegraphics[width=\textwidth]{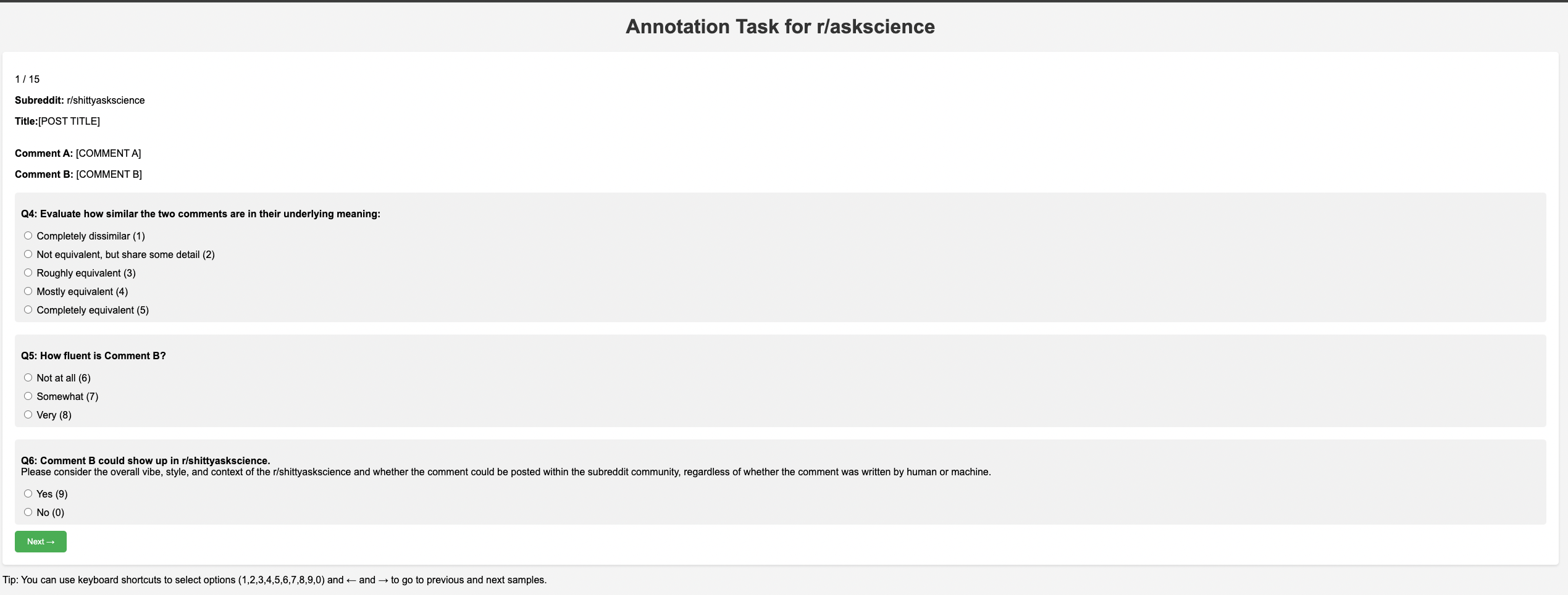}
    \caption{The annotators are presented with two versions of comments on a post: one synthetic and the other the original seed comment. Then, the annotators evaluated these two comments for their qualities, such as fluency and content preservation.}
    \label{fig:external-sample-anno}
\end{figure*}

To evaluate content preservation, we follow \citet{briakou-etal-2021-evaluating-custom} and adopt the Semantic Textual Similarity annotation scheme of \citet{agirre-etal-2016-semeval-custom}, where the original seed comment and its synthetic comment are rated on a scale based on the similarity of their underlying meaning (e.g., \textit{Completely Dissimilar, Not equivalent but share some details, Roughly Equivalent, Mostly Equivalent, Completely Equivalent}). To evaluate the fluency quality of the synthetic comments, we follow \citet{briakou-etal-2021-evaluating-custom} and ask annotators to assess the fluency of the comments (e.g., \textit{Not at all}, \textit{Somewhat}, \textit{Very}). To evaluate the naturalness of the synthetic comments, we employ a Turing Test approach from \citet{mir-etal-2019-evaluating-custom} and ask annotators to predict whether the comment was authored by a \textit{human} or \textit{machine}. Lastly, to evaluate the holistic quality of the synthetic comments, annotators were asked to consider the holistic vibe, style, and context of the subreddit and evaluate whether the comment could show up within the subreddit community (e.g., \textit{Yes}, \textit{No}). See \Fref{fig:external-sample-anno} for the sample questions from our annotation task.

Across 195 annotated examples, we found that 86\% obtained a rating of \textit{roughly equivalent} or better for content preservation between the synthetic and original comments, indicating that much of the underlying meaning was preserved in the synthetic comments (See \Tref{tab:content-pres-results} for the full annotation results on content preservation). Additionally, we found that 96\% of the synthetic comments obtained a fluency rating of ``Somewhat'' or ``Very'', suggesting that nearly all of our synthetic comments are indeed fluent (See \Tref{tab:fluency-results} for the full annotation results on fluency). As shown in \Tref{tab:naturalness}, we found that the expert annotators failed to detect the synthetic comments as machine-generated 50\% of the time, suggesting that much of the synthetic comments appear natural. Most importantly, annotators assessed that 71\% of the synthetic comments could be posted within the subreddit, indicating that the vast majority of the synthetic comments match the overall vibe, style, and context of the community (See \Tref{tab:overall-quality-results} for the full annotation results on the holistic quality). Overall, these results validate the quality of the synthetic data across content preservation, fluency, naturalness, and overall quality.

 \begin{table}[h!]
    \centering
    \setlength{\tabcolsep}{4pt}
    \resizebox{\linewidth}{!}{
    \begin{tabular}{cccccc}
    \toprule
        \multirow{2}{*}{\textbf{Topic}} 
        & \textbf{Completely} & \textbf{Share} & \textbf{Roughly} & \textbf{Mostly} & \textbf{Completely} \\
        & \textbf{Dissimilar} & \textbf{Details} & \textbf{Equiv.} & \textbf{Equiv.} & \textbf{Equiv.} \\\midrule
        Gender & 0.02 & 0.16 & 0.16 & 0.51 & 0.16 \\
        Politics & 0.02 & 0.16 & 0.29 & 0.4 & 0.13 \\
        Science & 0.04 & 0.07 & 0.13 & 0.42 & 0.33 \\
        Finance & 0.03 & 0.1 & 0.12 & 0.38 & 0.37 \\\midrule
        Total & 0.03 & 0.12 & 0.17 & 0.43 & 0.26\\\bottomrule
    \end{tabular}
    }\vspace{-2mm}
    \caption{The distribution of human judgments on content preservation between synthetic and original seed comments. Human annotators were asked to ``Evaluate how similar the two comments are in their underlying meaning.'' ``Comp. Dissimilar'' : Completely Dissimilar, ``Share Details'' : Not equivalent but share some details, ``Roughly Equiv.'' : Roughly Equivalent, ``Mostly Equiv.'' : Mostly Equivalent, and ``Comp. Equivalent'' : Completely Equivalent.}
    \label{tab:content-pres-results}
\end{table}

 \begin{table}[h!]
    \centering
    \resizebox{0.9\linewidth}{!}{
    \begin{tabular}{cccccc}
    \toprule
        \textbf{Topic} & \textbf{Not at all} & \textbf{Somewhat} & \textbf{Very} \\\midrule
        Gender & 0.04 & 0.16 & 0.80\\
        Politics & 0.04 & 0.09 & 0.87\\
        Science & 0.00 & 0.16 & 0.84\\
        Finance & 0.07 & 0.17 & 0.77\\\midrule
        Total & 0.04 & 0.14 & 0.82 \\\bottomrule
    \end{tabular}
    }
    \vspace{-2mm}
    \caption{The distribution of human judgments on the fluency of synthetic comments. The human annotators were asked to evaluate ``How fluent is [comment]?''}
    \label{tab:fluency-results}
\end{table}

 \begin{table}[h!]
    \centering
    \resizebox{0.9\linewidth}{!}{
    \begin{tabular}{ccc}
    \toprule
        \textbf{Topic} & \textbf{High Quality} & \textbf{Not High Quality} \\\midrule
        Gender & 0.67 & 0.33\\
        Politics & 0.58 & 0.42 \\
        Science & 0.91 & 0.09\\
        Finance & 0.70 & 0.30\\\midrule
        Total & 0.71 & 0.29 \\\bottomrule
    \end{tabular}
    }
    \vspace{-2mm}
    \caption{The distribution of human judgments on the holistic quality of synthetic comments. The human annotators were asked to consider the overall vibe, style, and context of the subreddit and evaluate ``[Comment] could show up in r/[subreddit].''}
    \label{tab:overall-quality-results}
\end{table}

 \begin{table}[h!]
    \centering
    \resizebox{\linewidth}{!}{
    \begin{tabular}{ccc}
    \toprule
        \textbf{Topic} & \textbf{Original Comments} & \textbf{Synthetic Comments} \\\midrule
        Gender & 0.96 & 0.43\\
        Politics & 0.73 & 0.13 \\
        Science & 0.75 & 0.78\\
        Finance & 0.42 & 0.78\\\midrule
        Total & 0.81 & 0.50 \\\bottomrule
    \end{tabular}
    }\vspace{-2mm}
    \caption{The percentage of original comments and synthetic comments that were predicted to be written by a human. The human annotators were asked to evaluate whether ``[Comment] was written by.''}
    \label{tab:naturalness}
\end{table}

\begin{table*}[htbp]
    \centering
    \resizebox{\textwidth}{!}{
    \begin{tabular}{cccccccc}
    \toprule
        \textbf{Model} & \textbf{Formality} & \textbf{Politeness} & \textbf{Humor} & \textbf{Supportiveness} & \textbf{Sarcasm} & \textbf{Verbosity} & \textbf{Average}\\ \midrule
        GPT-4o Judge & 0.93 & 0.96 & 0.87 & 0.95 & 0.84 & 0.87 & 0.90\\
        Normness model & 0.88 & 0.84 & 0.76 & 0.75 & 0.64 & 0.91 & 0.80\\
        \bottomrule
    \end{tabular}}
    \caption{Evaluation results on the faithfulness of the community language simulation. We sampled 1,560 pairs of original and Llama3-8b-Instruct generated style-transferred comments (e.g. rewritten to be more sarcastic) and used GPT-4o as a judge to determine whether the comment is, for example, more sarcastic than the original one, finding an average percentage agreement of 90\%. In addition, we checked whether the intended change by the prompt in the style transfer aligned with the normness scale predictor model, finding an average percentage agreement of 80\% across topics and norm dimension.}
    \label{tab:st_faithfulness}
\end{table*}

\subsection{Faithfulness of the Community Language Simulation}\label{app:st_faitfulness}

After conducting human evaluations to assess the content preservation, fluency, naturalness, and overall quality of the generated comments, we evaluated the \textit{faithfulness} of the community language simulation. Specifically, we validated whether the style-transferred comments adopted the intended norm dimension (e.g., more sarcasm) when prompted to. To do this, we sampled 1,560 pairs of original and Llama3-8b-Instruct generated style-transfer comments and conducted two validations. Table \ref{tab:st_faithfulness} contains the validation results.

Our validations demonstrate that the style-transferred comments successfully adopted the intended norm dimension when prompted. First, we employed GPT-4o as a judge to determine whether the generated comment had, for instance, become more sarcastic than the original one, finding an average percentage agreement of 90\%. Across the norm dimensions, we found that GPT-4o agreed with the intended style transfer, with percentage agreement rates ranging from 84\%-96\%. Second, we validated whether the intended change by the prompt in the style transfer aligned with the normness scale predictor model (NSP), finding an average percentage agreement of 80\% across the topics and norm dimensions. These validations collectively indicate that the style-transferred comments effectively captured the intended shifts in the norm dimension (e.g., becoming sarcastic).

\section{Community Preference Prediction}\label{app:upvote_pred}
\subsection{CPP Training Details}\label{app:upvote_training}

The training label for the CPP model is derived from the logarithm of the net upvotes (upvotes minus downvotes) across various subreddits. This approach helps to stabilize the variance and improve the model's performance with skewed distributions of upvote counts. The input is described in \Sref{sec:exp:upvote_pred} to take on 4 variations containing different extents of contextual information. 

The model was trained for five epochs across most subreddits to ensure adequate learning without overfitting. However, for subreddits with larger datasets—specifically AskMen, AskWomen, WallStreetBets, and Libertarian—training was limited to two epochs. This adjustment was made to keep the total number of training steps across all subreddits on the same magnitude, thus enabling fair comparison.

The learning rate was set at $1 \times 10^{-5}$, with a batch size of 128. The Mean Squared Error (MSE) loss function was used, a standard choice for regression models that promotes the minimization of the average squared difference between the estimated values and what is estimated. This choice helps in refining the model's accuracy by adjusting weights based on the gradient of the loss incurred with each epoch.

\subsection{CPP Evaluation Details \& Results}\label{app:upvote_eval}

We use binary accuracy, which measures whether predicted relationship (greater or lesser approval) between any two comments aligns with their actual relationship derived from ground truth data. This metric determines if the model correctly predicts the relative preference between pairs of randomly sampled comments, grounded in their ground truth preference scores. The model's accuracy varied significantly depending on the contextual information provided during training. Specifically, the basic \textbf{comment} only variant averaged an accuracy of 61.8\%, indicating a foundational level of predictability based on comment content alone. With the addition of \textbf{post} context, the accuracy improved to 65.6\%, underscoring the importance of the discussion's broader context in influencing user preferences.

Further enhancements in model input by including \textbf{time} metadata yielded an average accuracy of 73.9\%, reflecting the temporal dynamics of user interactions and preferences. The comprehensive variant, which incorporates \textbf{comment}, \textbf{post}, \textbf{time}, and \textbf{author} information, maintained a similar accuracy, suggesting a marginal gain from including author-specific data. However, this was notably beneficial in subreddits with strong individual influencer effects such as \texttt{r/libertarian}, where the accuracy increased slightly, implying that certain communities benefit more from recognizing individual contribution patterns.

Subreddit-specific analysis revealed that preferences of \texttt{r/askwomen} is the easiest to learn, with an accuracy of 80.8\% for the \textbf{comment+post+time} variant, likely due to its focused content and consistent user engagement patterns. In contrast, politically oriented subreddits like \texttt{r/libertarian}, \texttt{r/democrats}, and \texttt{r/republican} faced lower accuracies, reflecting the challenge of modeling preferences in environments with dynamic, ideologically charged discussions. The impact of rapidly changing topical engagement and the diverse ideological landscape within these communities makes preference prediction particularly challenging. The model's relative struggle in these contexts highlights the complex interplay of content, timing, and participant identity in shaping online discourse and user preferences.

\begin{table}[ht!]
    \centering
    \resizebox{\linewidth}{!}{
    \begin{tabular}{ccccc}
    \toprule
\textbf{Comment}            & \textbf{X}& \textbf{X}& \textbf{X}& \textbf{X} \\
\textbf{Post}               &      -    & \textbf{X}& \textbf{X}& \textbf{X} \\
\textbf{Time}               &      -    &      -    & \textbf{X}& \textbf{X} \\
\textbf{Author}             &      -    &      -    &      -    & \textbf{X} \\\midrule
\texttt{r/askmen}           & 59.3      & 67.9      & 77.3      & 77.2 \\
\texttt{r/askwomen}         & 60.2      & 66.3      & 80.8      & 80.0 \\
\texttt{r/asktransgender}   & 60.6      & 68.9      & 78.3      & 78.3 \\ \midrule
\texttt{r/libertarian}      & 58.9      & 61.4      & 67.3      & 69.8 \\ 
\texttt{r/democrats}        & 60.0      & 66.0      & 75.7      & 70.4 \\
\texttt{r/republican}       & 62.7      & 63.3      & 70.9      & 70.8 \\ \midrule
\texttt{r/askscience}       & 62.9      & 65.1      & 71.9      & 71.9 \\
\texttt{r/shittyaskscience} & 59.8      & 66.3      & 74.6      & 74.5 \\
\texttt{r/akksciencediscussion}& 60.8   & 63.5      & 71.8      & 71.8 \\ \midrule
\texttt{r/wallstreetbets}   & 61.9      & 65.2      & 70.3      & 69.1 \\
\texttt{r/stocks}           & 60.2      & 63.1      & 70.3      & 70.7 \\
\texttt{r/pennystocks}      & 62.8      & 66.0      & 72.5      & 72.2 \\
\texttt{r/wallstreetbetsnew}& 70.8      & 75.8      & 79.1      & 79.1 \\ \midrule
Average    & 61.7 \scriptsize$\pm$3.0\normalsize  & 66.1 \scriptsize$\pm$3.6\normalsize & 73.9 \scriptsize$\pm$4.1\normalsize & 73.5 \scriptsize$\pm$3.8\normalsize \\
        \bottomrule
    \end{tabular}}\vspace{-2mm}
    \caption{Community Preference Prediction model accuracy across four proposed variants.}
    \label{tab:upvote_results}
\end{table}

\section{Point of Maximum Return}\label{app:max_return}
\begin{figure*}[t!]
    \centering
    \begin{subfigure}[b]{0.33\textwidth}
        \centering
        \includegraphics[width=\textwidth]{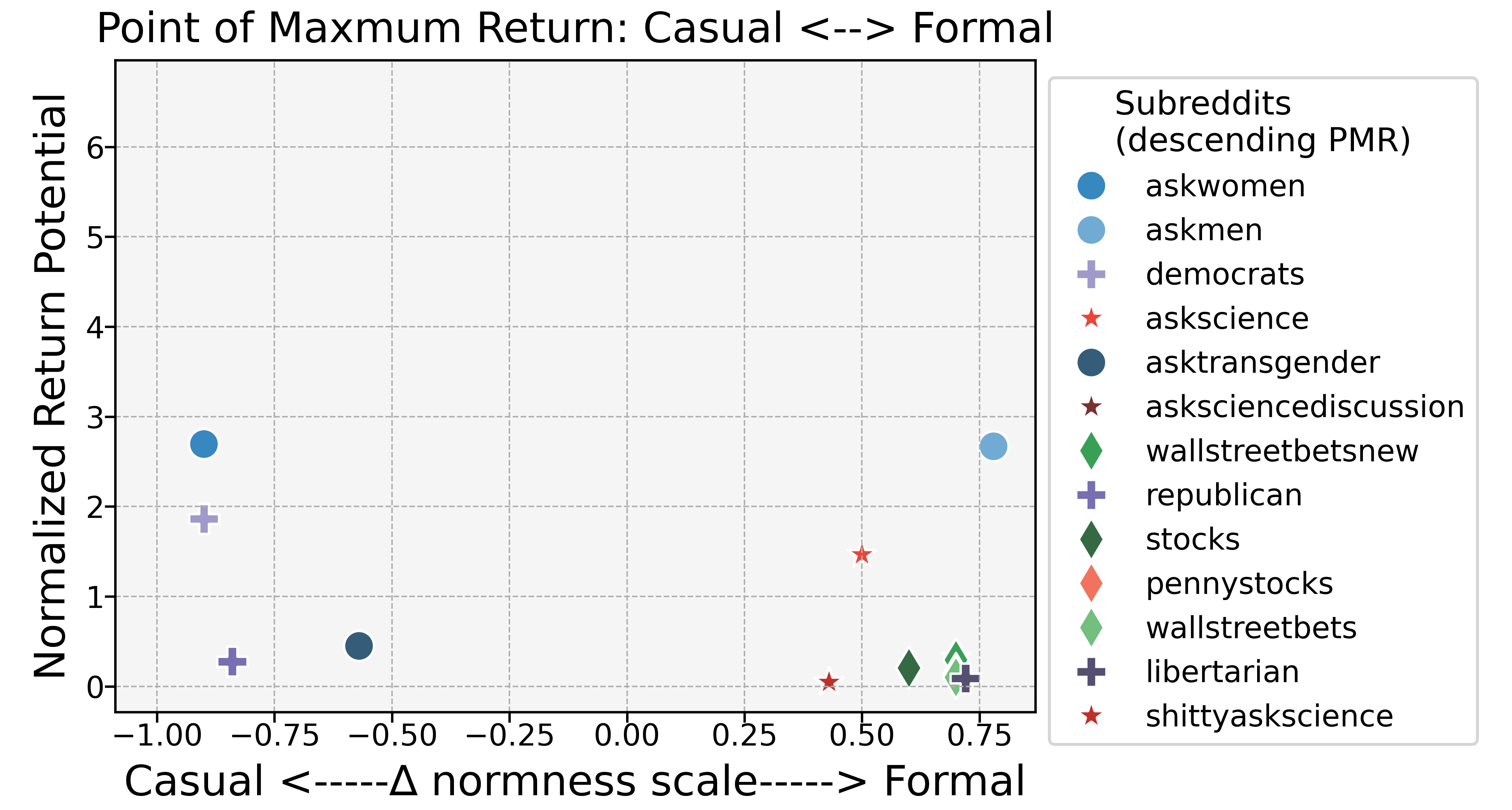}
        \caption{PMR for Formality}
    \end{subfigure}\hfill
    \begin{subfigure}[b]{0.33\textwidth}
        \centering
        \includegraphics[width=\textwidth]{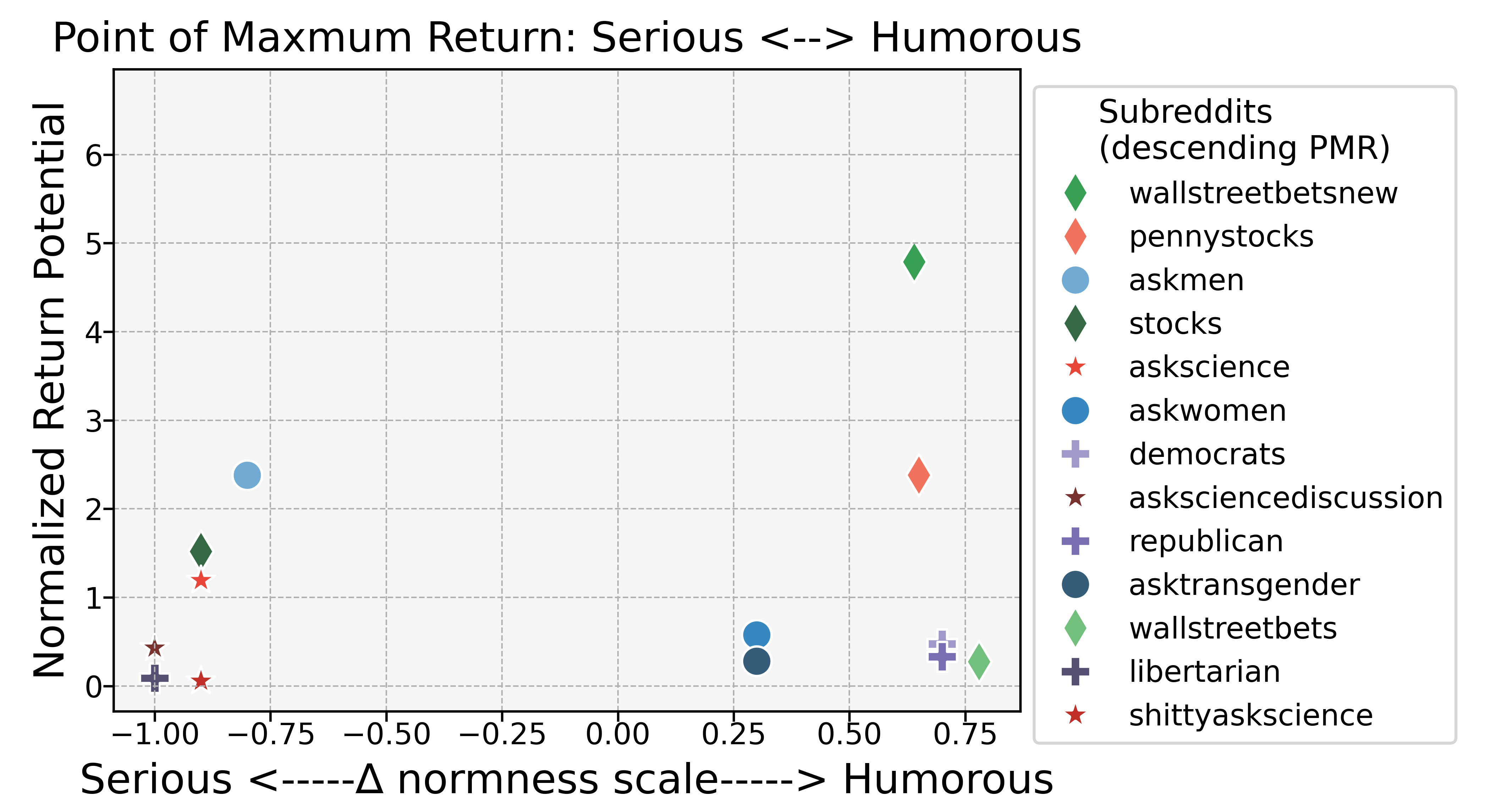}
        \caption{PMR for Humor}
    \end{subfigure}\hfill
    \begin{subfigure}[b]{0.33\textwidth}
        \centering
        \includegraphics[width=\textwidth]{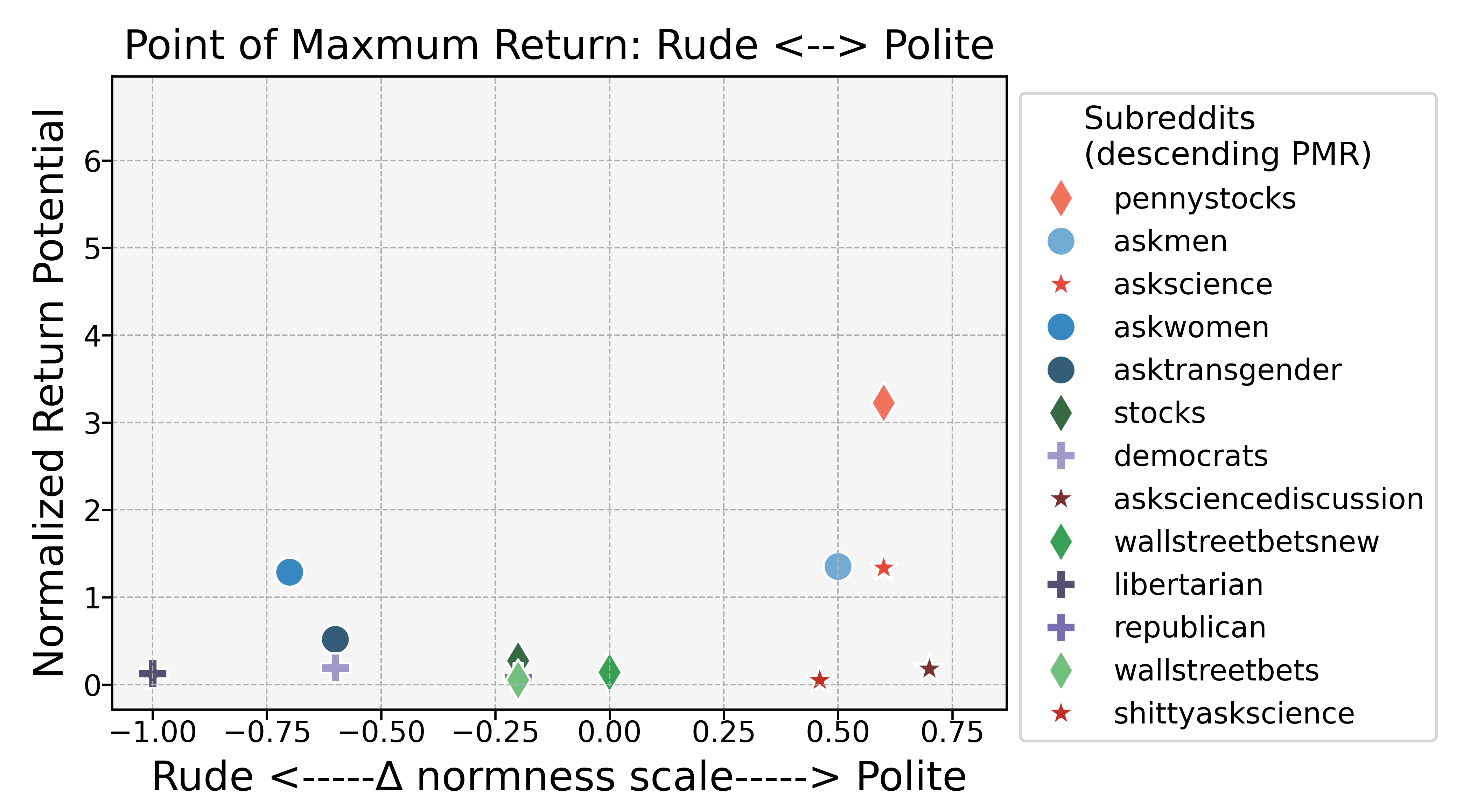}
        \caption{PMR for Politeness}
    \end{subfigure}\\
    \hspace*{\fill}%
    \begin{subfigure}[b]{0.33\textwidth}
        \centering
        \includegraphics[width=\textwidth]{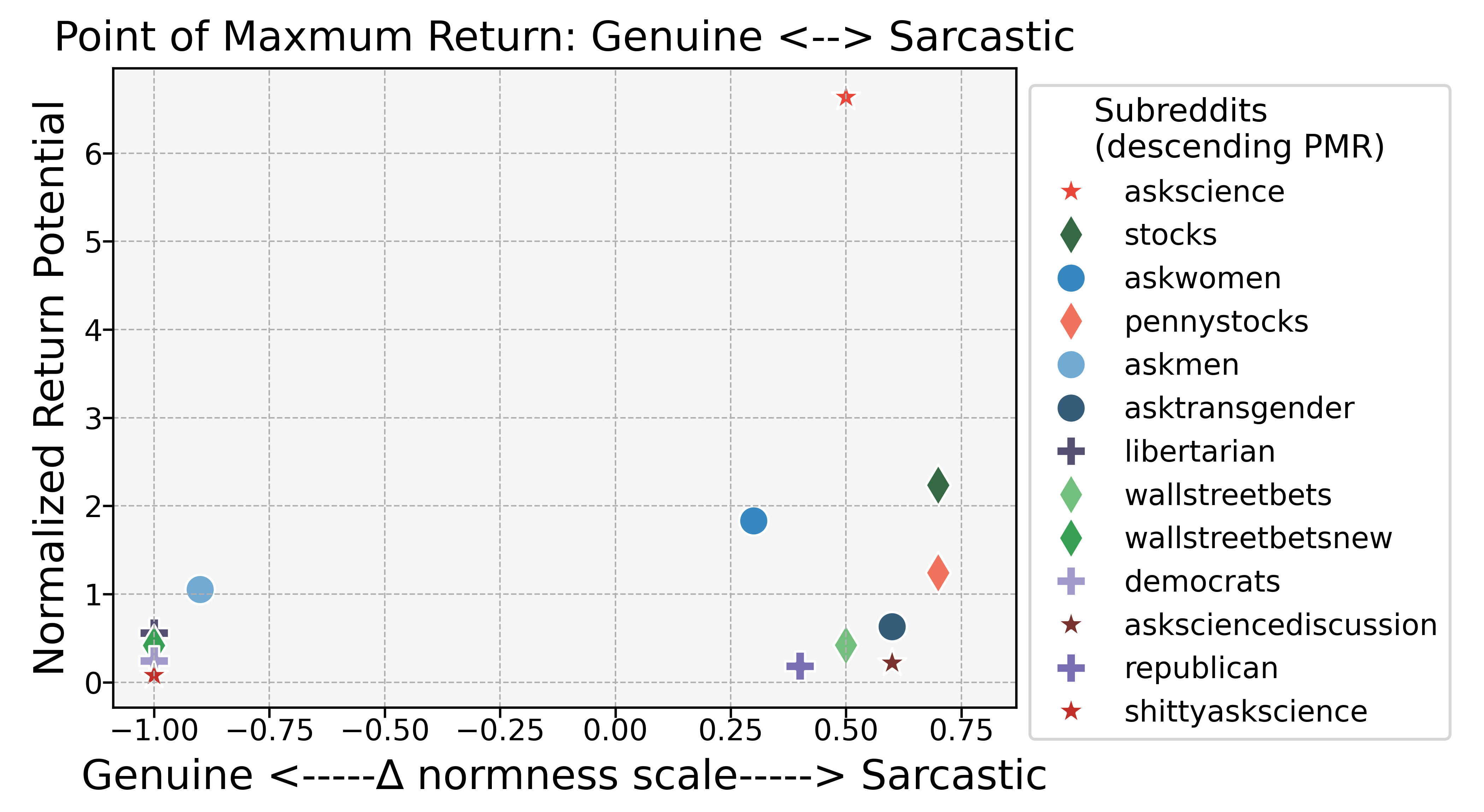}
        \caption{PMR for Sarcasm}
    \end{subfigure}
    \hfill%
    \begin{subfigure}[b]{0.33\textwidth}
        \centering
        \includegraphics[width=\textwidth]{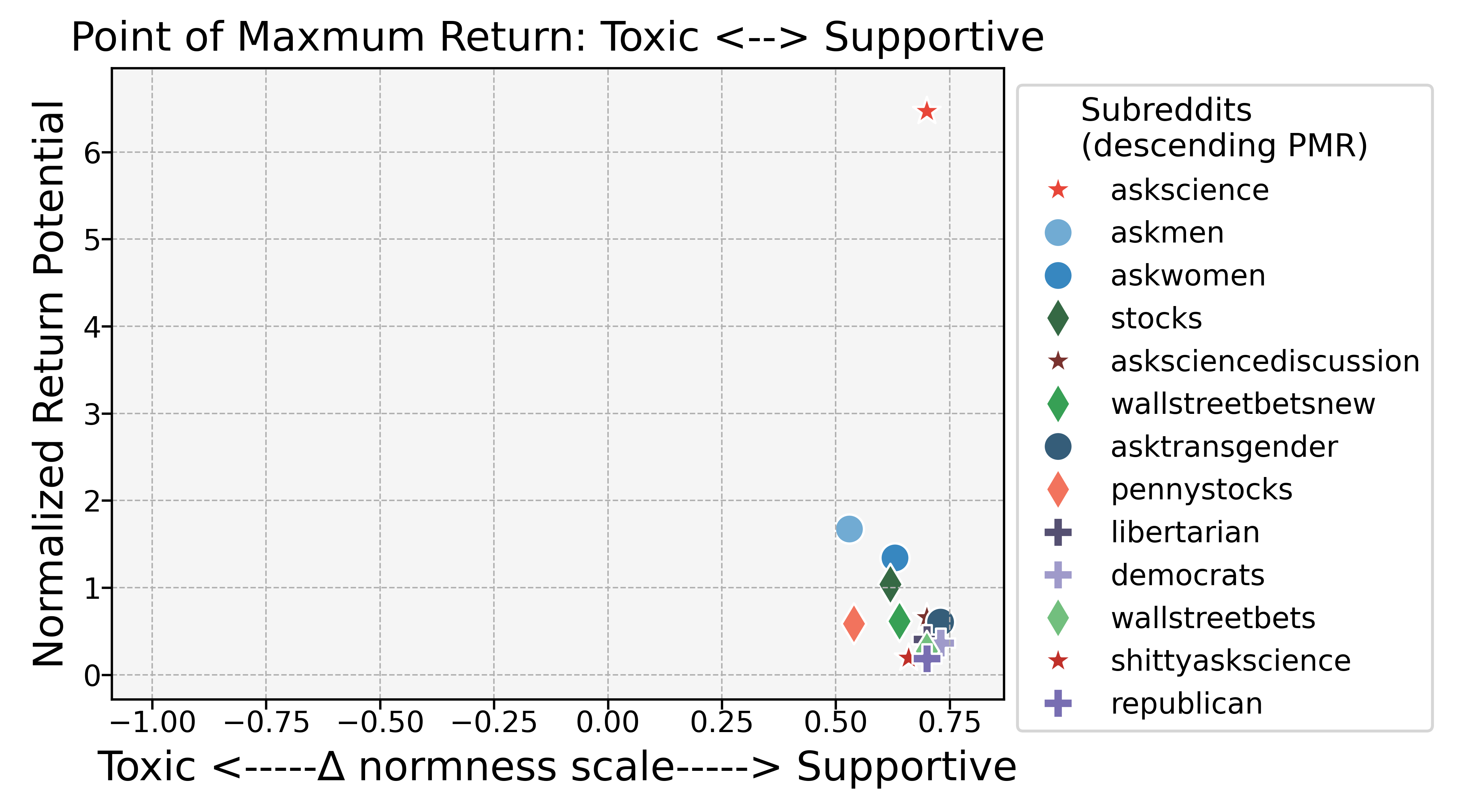}
        \caption{PMR for Supportiveness}
    \end{subfigure}
    \hspace*{\fill}%
    \caption{Maximum Return Potentials for all 13 subreddits along each norm dimension.}\label{fig:max_return_potential_all}
\end{figure*}

\Fref{fig:max_return_potential_all} shows the point of maximum return potential for the top 5 subreddits along each norm dimension. We find that the salient norms shown in the plots correspond to explicit subreddit rules, and report the rules that we refer to at the time of the analysis in our Github repository.

\section{Intensity \& Crystallization}\label{app:crystallization}

For each equidistant bin on the normness dimension, we sample equal number of comments and compute $NI$ as the mean norm intensity and $CR$ as the inverse of variance of norm intensity following \citet{linnan2005norms} as follows: 
\begin{align*}
    NI_{c,\Phi_d^i,t} = \frac{\sum_{a_j\in\mathcal{A}_{c,d,t}^i}\Psi_c(a_j)}{|\mathcal{A}_{c,d,t}^i|},\\
    CR_{c,\Phi_d^i,t} = \frac{|\mathcal{A}_{c,d,t}^{i'}|}{\sum_{a_j\in\mathcal{A}_{c,d,t}^{i'}}(\Psi_c(a_j)-NI_{c,\Phi_d^i,t})^2}
\end{align*}
where $\mathcal{A}_{c,d,t}^i$ is the set of comments posted within the given period $t$ in community $c$ on dimension $d$, and $\mathcal{A}'_{c,d,t}$ is the set of subsampled comments by the number of comments in a bin that has the minimum number of comments, to make the variance across bins comparable.  The dependent variable representing temporal changes in norms is defined as $TC_{c,\Phi_d^i,s_1,s_2}=NI_{c,\Phi_d^i,s_1}-NI_{c,\Phi_d^i,s_2}$, where we set $s_1$ as 2019-2020 and $s_2$ as 2021-2023.

We fit two linear regression models to predict $TC$: one using only $NI$ and another using both $NI$ and $CR$. We then evaluate the models' coefficients and $R^2$ (\Tref{tab:crystallization}). The results show that $NI$ and $CR$ are significant predictors of temporal change. Across all norm dimensions, the coefficients for both variables were statistically significant ($p$<0.01). Additionally,
$R^2$ increased significantly when $CR$ was added as an independent variable. Interestingly, the signs of the coefficients were opposite: positive for $NI$ and negative for $CR$. This suggests that higher norm intensity and less crystallization (i.e. community members have strong opinions about them but less agreed upon) make norms more likely to change over time. Our findings support \citet{jackson1975normative}'s hypothesis that norms with high $NI$ and low $CR$ are prone to generating conflicts within the community, thereby triggering changes in their norms. This demonstrates \methodname's potential for helping moderators identify norms likely to change and proactively address them, such as by setting explicit community rules.

\section{{User Level Community Norm Adaptation}}
\label{appendix:user-norm}

\begin{table*}[ht]
\centering
\resizebox{\linewidth}{!}{
\begin{tabular}{@{}lccccc@{}}
\toprule
\textbf{level} & \textbf{politeness} & \textbf{supportiveness} & \textbf{sarcasm} & \textbf{humor} & \textbf{formality} \\ 
\midrule
\texttt{r/wallstreetbets} $\rightarrow$ \texttt{r/wallstreetbetsnew} (925.6) & \graycell{-0.003} & \graycell{0.013} & \graycell{0.003} & \graycell{0.005} & \greencell{0.018} \\ 
\texttt{r/wallstreetbets} $\rightarrow$ \texttt{r/stocks} (2157.6) & \greencell{0.084} & \greencell{0.092} & \redcell{-0.044} & \redcell{-0.062} & \greencell{0.131} \\ 
\texttt{r/wallstreetbets} $\rightarrow$ \texttt{r/pennystocks} (1052.0) & \greencell{0.091} & \greencell{0.094} & \redcell{-0.023} & \redcell{-0.084} & \greencell{0.063} \\ 
\texttt{r/wallstreetbetsnew} $\rightarrow$ \texttt{r/stocks} (641.4) & \greencell{0.079} & \greencell{0.063} & \redcell{-0.067} & \redcell{-0.049} & \greencell{0.072} \\ 
\texttt{r/wallstreetbetsnew}  $\rightarrow$ \texttt{r/pennystocks} (566.4) & \greencell{0.083} & \greencell{0.080} & \redcell{-0.046} & \redcell{-0.078} & \greencell{0.036} \\ 
\texttt{r/stocks} $\rightarrow$ \texttt{r/pennystocks} (1524.6) & \graycell{-0.005} & \graycell{0.005} & \greencell{0.026} & \graycell{-0.011} & \redcell{-0.049} \\ 
\texttt{r/republican} $\rightarrow$ \texttt{r/libertarian} (497.0) & \greencell{0.027} & \greencell{0.053} & \redcell{-0.028} & \graycell{-0.002} & \greencell{0.036} \\ 
\texttt{r/republican} $\rightarrow$ \texttt{r/democrats} (223.8) & \graycell{0.026} & \graycell{0.016} & \greencell{0.036} & \graycell{0.018} & \graycell{-0.008} \\ 
\texttt{r/libertarian}  $\rightarrow$ \texttt{r/democrats}  (243.8) & \graycell{-0.007} & \graycell{-0.023} & \graycell{0.026} & \graycell{0.013} & \graycell{0.003} \\ 
\texttt{r/askscience} $\rightarrow$ \texttt{r/shittyaskscience} (133.4) & \redcell{-0.275} & \redcell{-0.308} & \greencell{0.292} & \greencell{0.326} & \redcell{-0.274} \\ 
\texttt{r/askscience} $\rightarrow$ \texttt{r/asksciencediscussion} (367.2) & \redcell{-0.054} & \graycell{-0.048} & \greencell{0.057} & \greencell{0.071} & \redcell{-0.089} \\ 
\texttt{r/shittyaskscience} $\rightarrow$ \texttt{r/asksciencediscussion} (94.2) & \greencell{0.174} & \greencell{0.186} & \redcell{-0.177} & \redcell{-0.202} & \greencell{0.146} \\ 
\texttt{r/askwomen} $\rightarrow$ \texttt{r/askmen} (717.4) & \graycell{-0.015} & \graycell{-0.022} & \graycell{0.026} & \greencell{0.036} & \graycell{0.004} \\ 
\texttt{r/askwomen}  $\rightarrow$ \texttt{r/asktransgender} (132.4) & \graycell{0.026} & \greencell{0.037} & \graycell{0.007} & \redcell{-0.073} & \greencell{0.053} \\ 
\texttt{r/askmen} $\rightarrow$ \texttt{r/asktransgender} (47.0) & \graycell{0.014} & \greencell{0.085} & \graycell{-0.086} & \redcell{-0.128} & \graycell{0.040} \\ 
\bottomrule
\end{tabular}
}
\caption{Norm differences and p-values across various subreddit transitions. Gray cells indicate changes that are insignificant (p > 0.05) according to a paired t-test. Red and green~cells represent significant negative and positive changes. In the row ``republican $\rightarrow$ libertarian,'' users posted more polite, more supportive, more formal, and less sarcastic comments in \texttt{r/libertarian} than in \texttt{r/republican}.}\label{appendix-table:norm-diff-user}
\end{table*}

In this section, we examine how individual users modify their language and interaction styles based on the community they are interacting with.
\Tref{appendix-table:norm-diff-user} presents the average change in norms for common users between two subreddits. We define user norm behavior in a community as the average $NI$ of comments left by the specific user in the community. 
For each subreddit, we only included users who had written at least two comments included in our analysis, ensuring we had a reliable measure of their behavior.
We present the averages across users in the table, and we conducted a paired two-tailed t-test to determine if these differences are statistically significant from 0. The results indicate whether users' language changes more positively (green cells), negatively (red cells), or does not change significantly (gray cells). For instance, green cells indicate that the users adapt their behavior to exhibit \textit{more} of the norm dimension (e.g. politeness) between subreddits. 

Our observations provide valuable insights into the adaptive mechanisms of online communities, revealing how community norms are not static but evolve in response to internal dynamics and external sociopolitical events. Understanding these variations can aid in managing community dynamics, which is vital for platform administrators and content creators to foster positive and inclusive communities.

\section{RPM Plots}\label{appendix:rpm-plots}
We applied \methodname{} to four different topics (gender, finance, politics, science) across six norm dimensions (supportiveness, formality, politeness, sarcasm, humor, verbosity). The resulting RPM plots illustrate how community approval (i.e., preference) changes as the normness scale of the comment varies. Our results, in turn, provide insights into the norms of each community.

\subsection{Gender Subreddits}
\begin{figure}[!h]
    \centering
    \includegraphics[width=\columnwidth]{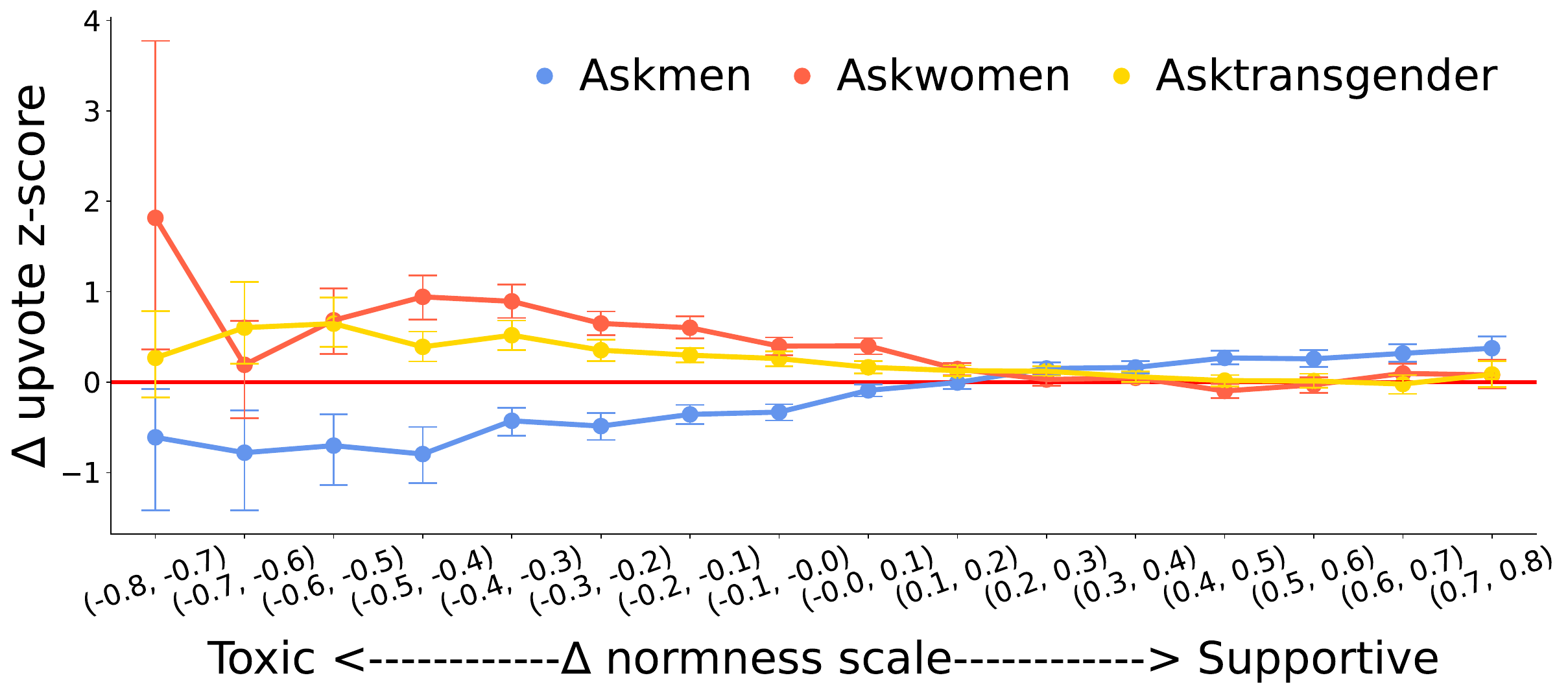}
    \caption{RPM plots for gender subreddits on the supportiveness dimension.}
    \label{fig:rpm-gender-supportiveness}
\end{figure}

\begin{figure}[!h]
    \centering
    \includegraphics[width=\columnwidth]{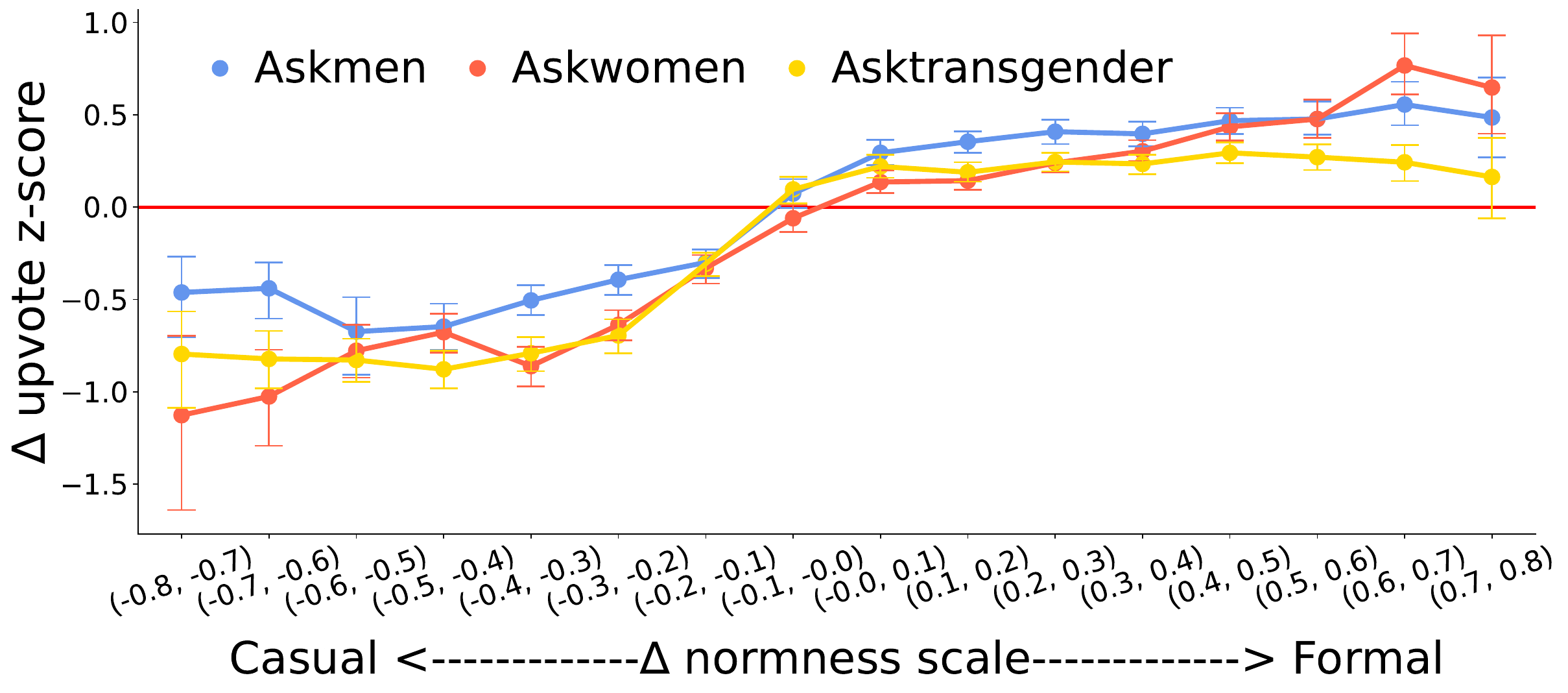}
    \caption{RPM plots for gender subreddits on the formality dimension.}
    \label{fig:rpm-gender-formality}
\end{figure}

\begin{figure}[!h]
    \centering
    \includegraphics[width=\columnwidth]{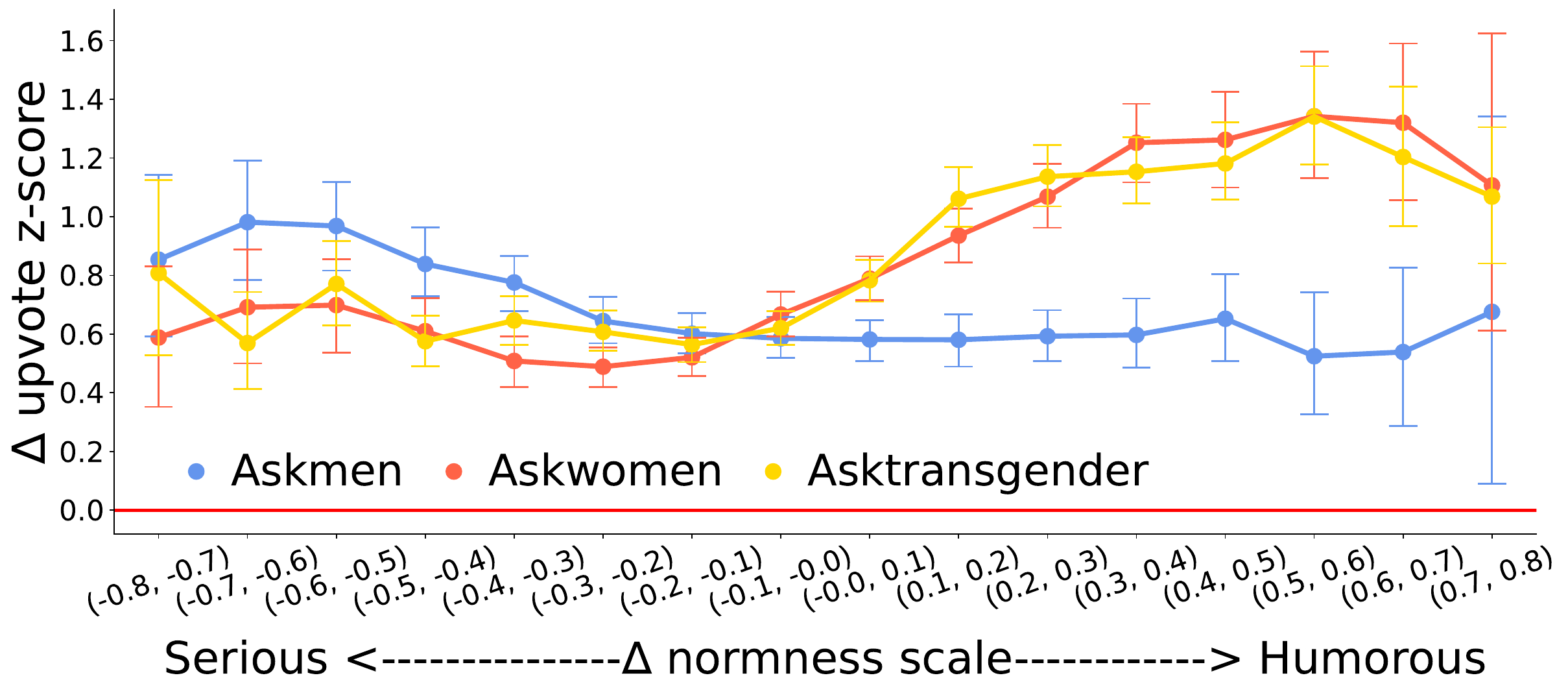}
    \caption{RPM plots for gender subreddits on the humor dimension.}
    \label{fig:rpm-gender-humor}
\end{figure}

We examined three subreddits related to gender: \texttt{r/askmen}, \texttt{r/askwomen}, and \texttt{r/asktransgender}. In terms of formality, sarcasm, and verbosity, there are no significant differences across the three subreddits. However, supportiveness, politeness, and humor show distinct variations. While \texttt{r/askwomen} and \texttt{r/asktransgender} exhibit similar trends, \texttt{r/askmen} notably disapproves of toxic (Figure \ref{fig:rpm-gender-supportiveness}) and rude comments, prefers polite comments (Figure \ref{fig:rpm-gender-formality}), and reacts less to humorous comments (Figure \ref{fig:rpm-gender-humor}) compared to the other two subreddits.
Our findings suggest that \texttt{r/askwomen} and \texttt{r/asktransgender} share similar norms and values, whereas \texttt{r/askmen} appears to be a relatively more polite and serious community.

\begin{figure}[!h]
    \centering
    \includegraphics[width=\columnwidth]{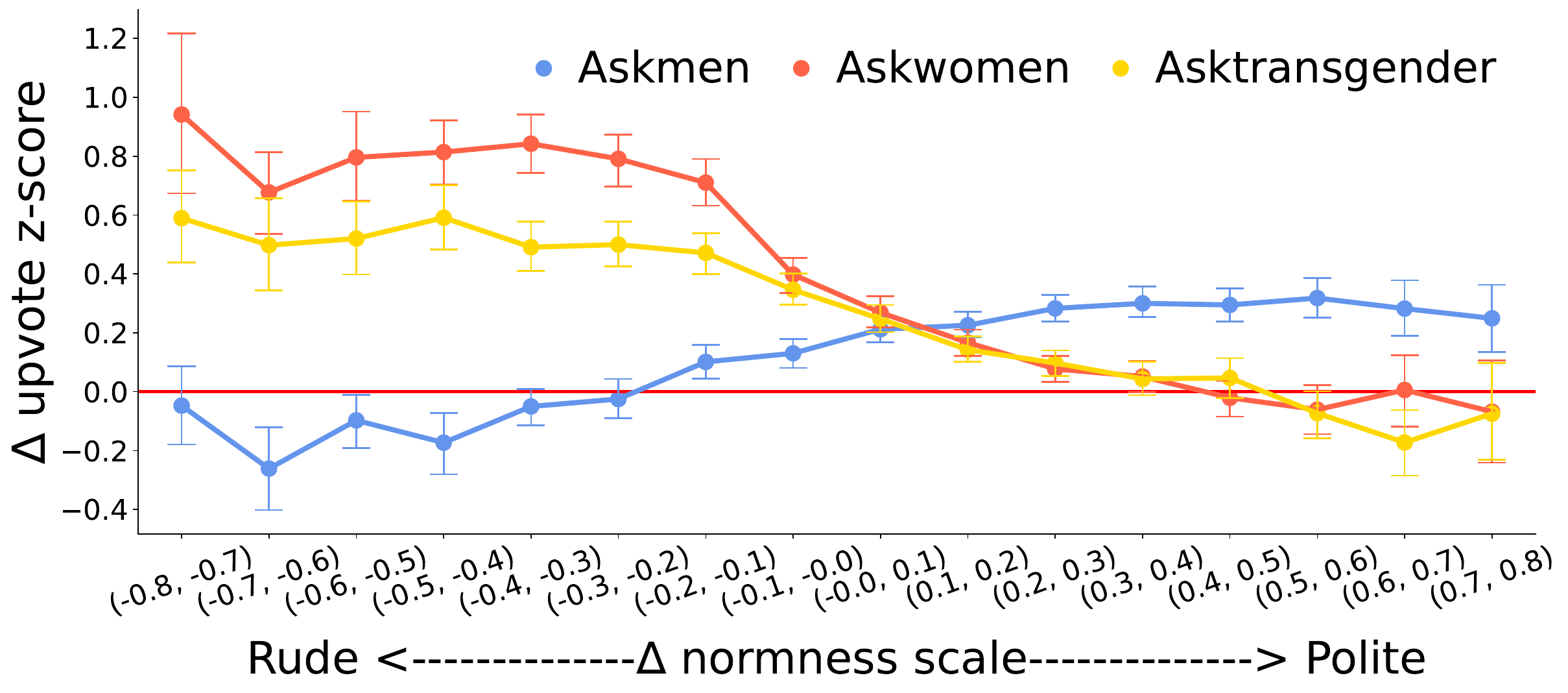}
    \caption{RPM plots for gender subreddits on the politeness dimension.}
    \label{fig:rpm-gender-politeness}
\end{figure}

\begin{figure}[!h]
    \centering
    \includegraphics[width=\columnwidth]{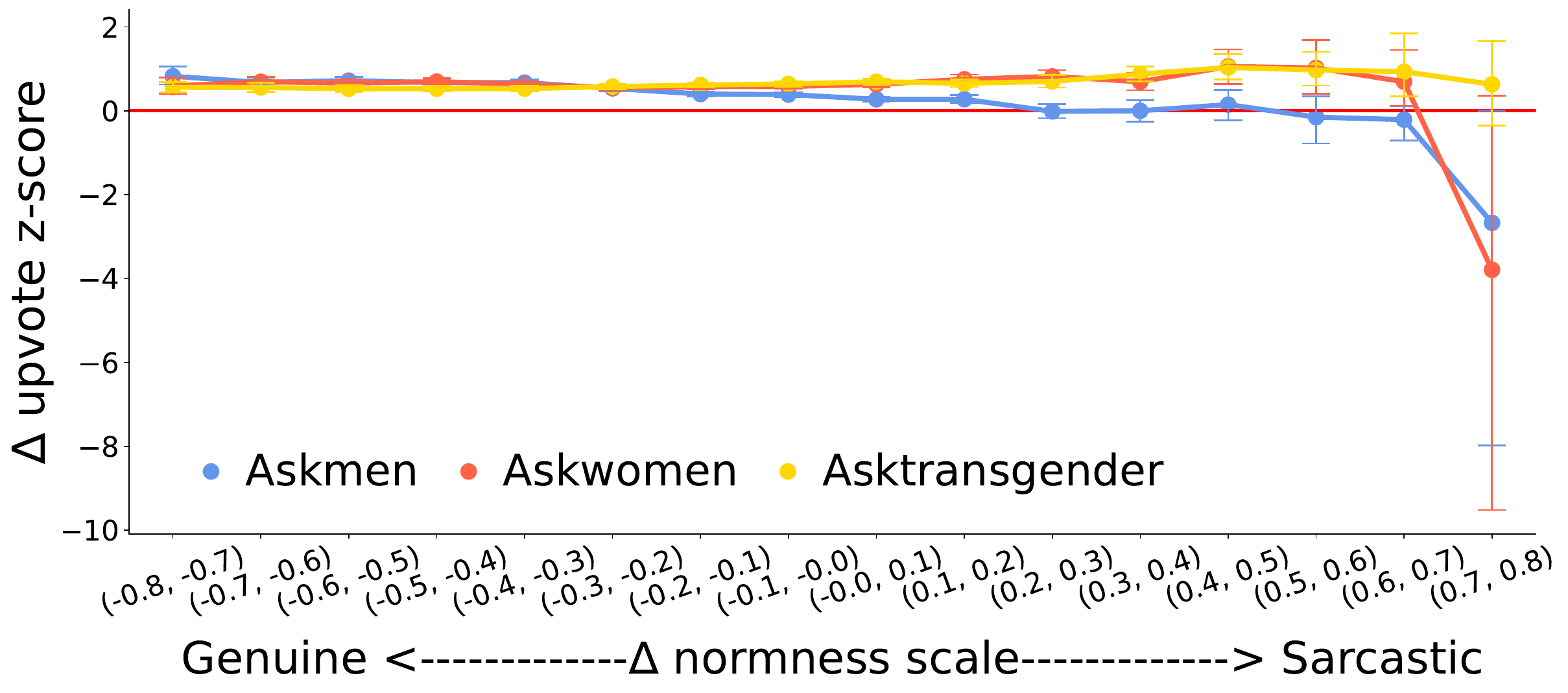}
    \caption{RPM plots for gender subreddits on the sarcasm dimension.}
    \label{fig:rpm-gender-sarcasm}
\end{figure}

\begin{figure}[!h]
    \centering
    \includegraphics[width=\columnwidth]{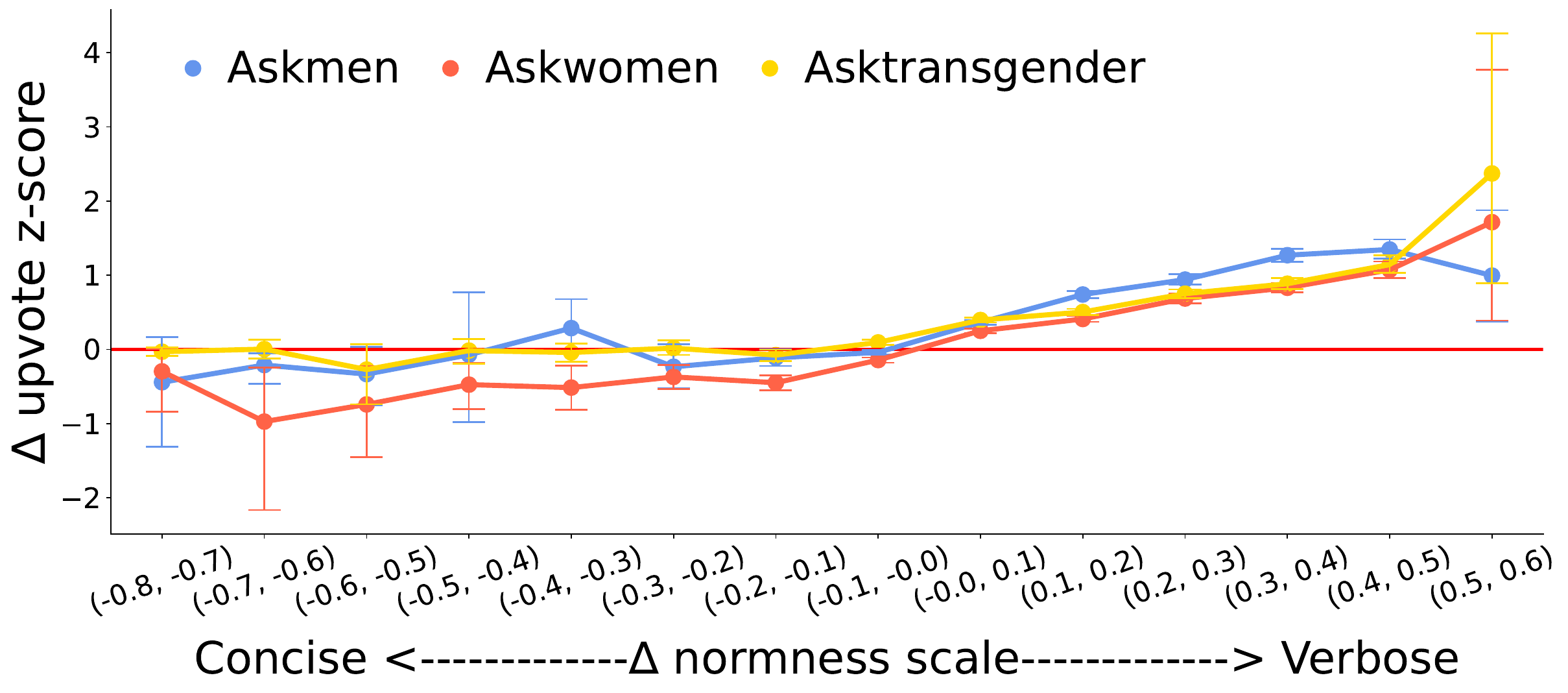}
    \caption{RPM plots for gender subreddits on the verbosity dimension.}
    \label{fig:rpm-gender-verbosity}
\end{figure}

\subsection{Finance Subreddits}
\begin{figure}[h!]
    \centering
    \includegraphics[width=\columnwidth]{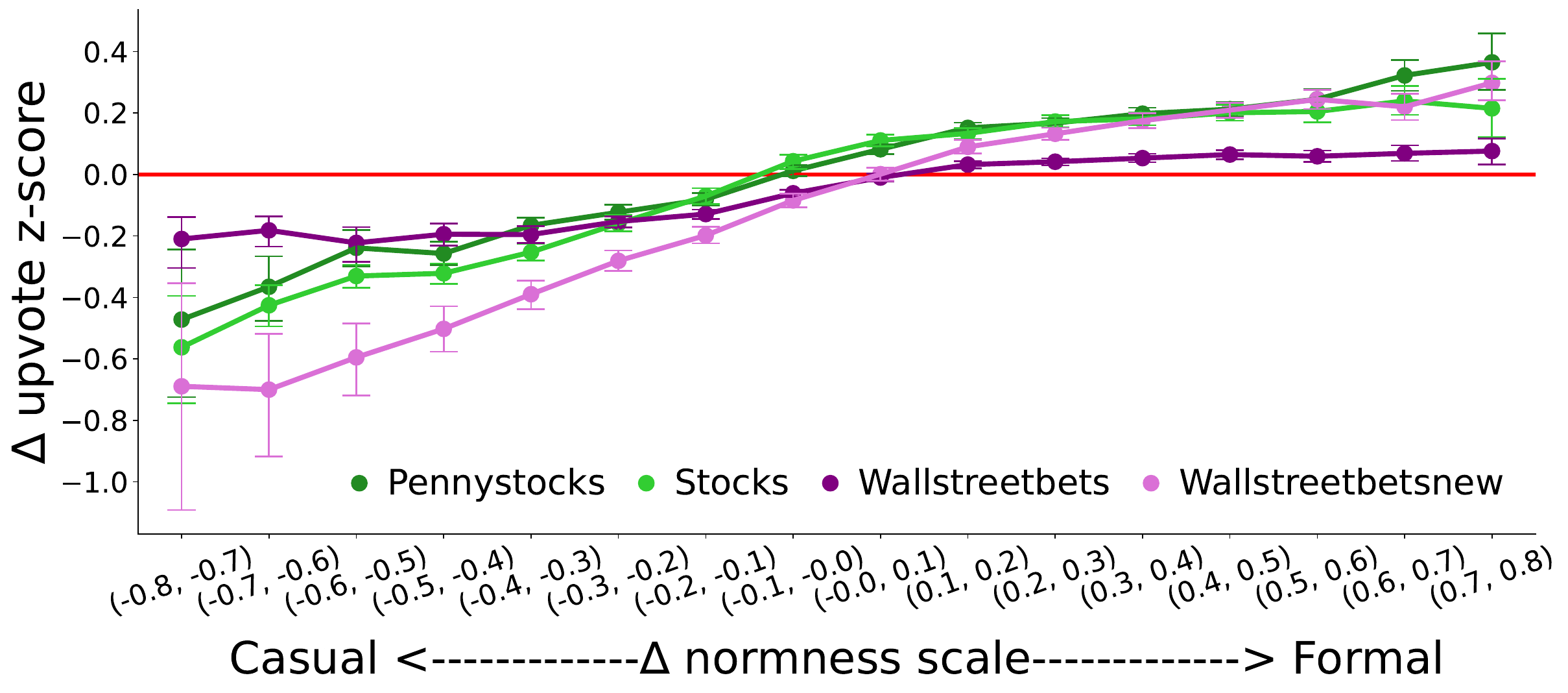}
    \caption{RPM plots for finance subreddits on the formality dimension.}
    \label{fig:rpm-finance-formality}
\end{figure}

\begin{figure}[h!]
    \centering
    \includegraphics[width=\columnwidth]{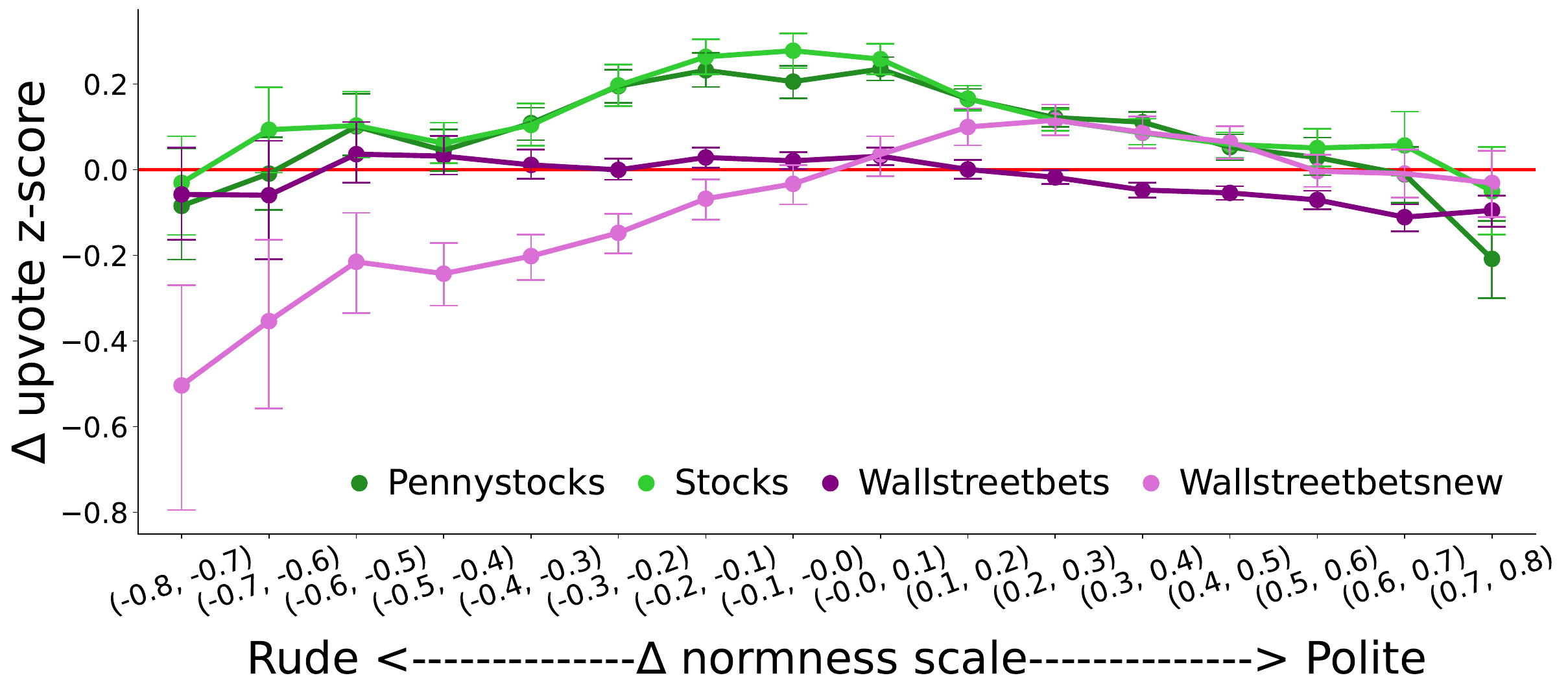}
    \caption{RPM plots for finance subreddits on the politeness dimension.}
    \label{fig:rpm-finance-politeness}
\end{figure}

\begin{figure}[h!]
    \centering
    \includegraphics[width=\columnwidth]{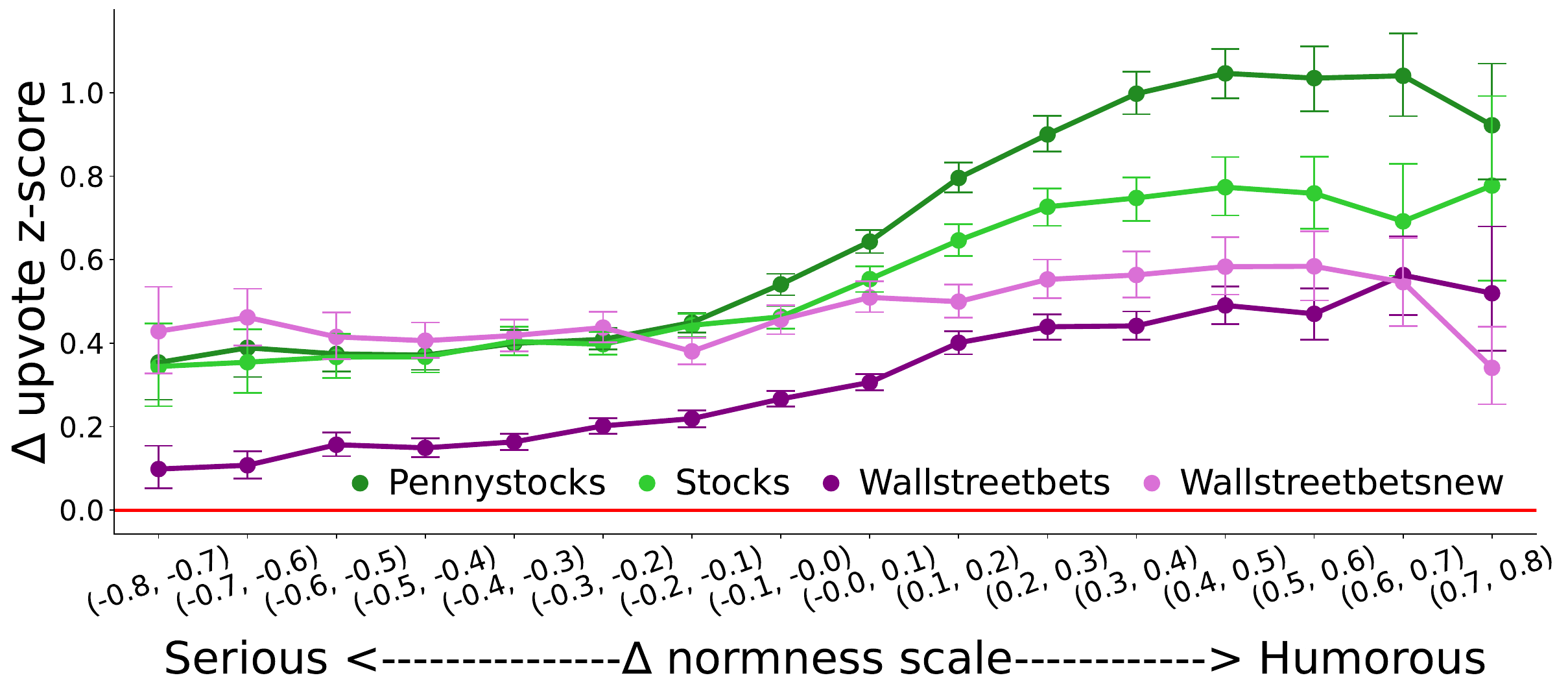}
    \caption{RPM plots for finance subreddits on the humor dimension.}
    \label{fig:rpm-finance-humor}
\end{figure}

We examined four subreddits related to finance: \texttt{r/pennystocks}, \texttt{r/stocks}, \texttt{r/wallstreetbets}, and \texttt{r/wallstreetbetsnew}. Unlike the gender subreddits, which showed distinct patterns in some dimensions, the finance subreddits exhibit similar trends overall, differing primarily in the degree of their preferences. For example, all four subreddits disapprove of overly casual and rude comments, with \texttt{r/wallstreetbets} showing the strongest disapproval (Figure \ref{fig:rpm-finance-formality} and \ref{fig:rpm-finance-politeness}). Additionally, we find that all finance subreddits prefer humorous comments over serious ones, with \texttt{r/wallstreetbets} displaying a significant dislike for serious comments (Figure \ref{fig:rpm-finance-humor}).

\begin{figure}[!h]
    \centering
    \includegraphics[width=\columnwidth]{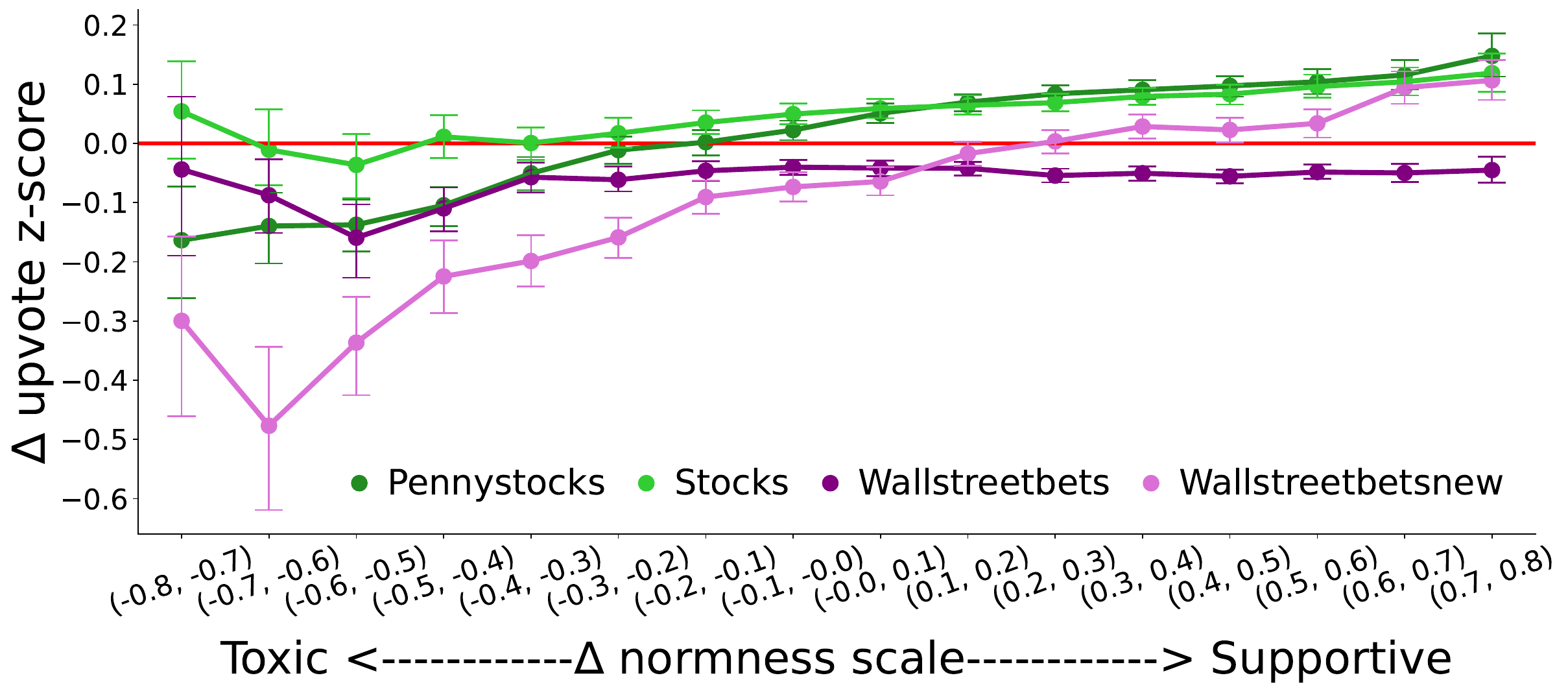}
    \caption{RPM plots for finance subreddits on the supportiveness dimension.}
    \label{fig:rpm-finance-supportiveness}
\end{figure}

\begin{figure}[h!]
    \centering
    \includegraphics[width=\columnwidth]{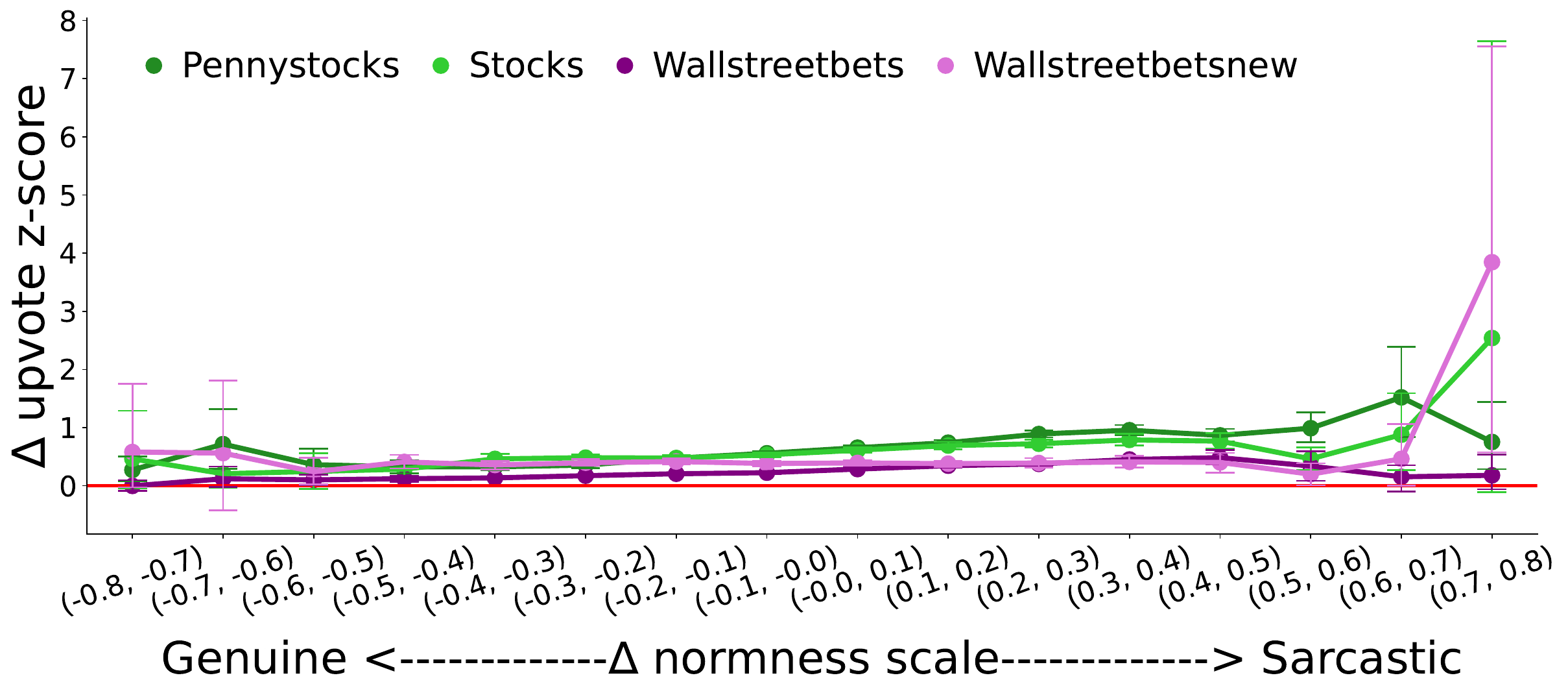}
    \caption{RPM plots for finance subreddits on the sarcasm dimension.}
    \label{fig:rpm-finance-sarcasm}
\end{figure}

\begin{figure}[h!]
    \centering
    \includegraphics[width=\columnwidth]{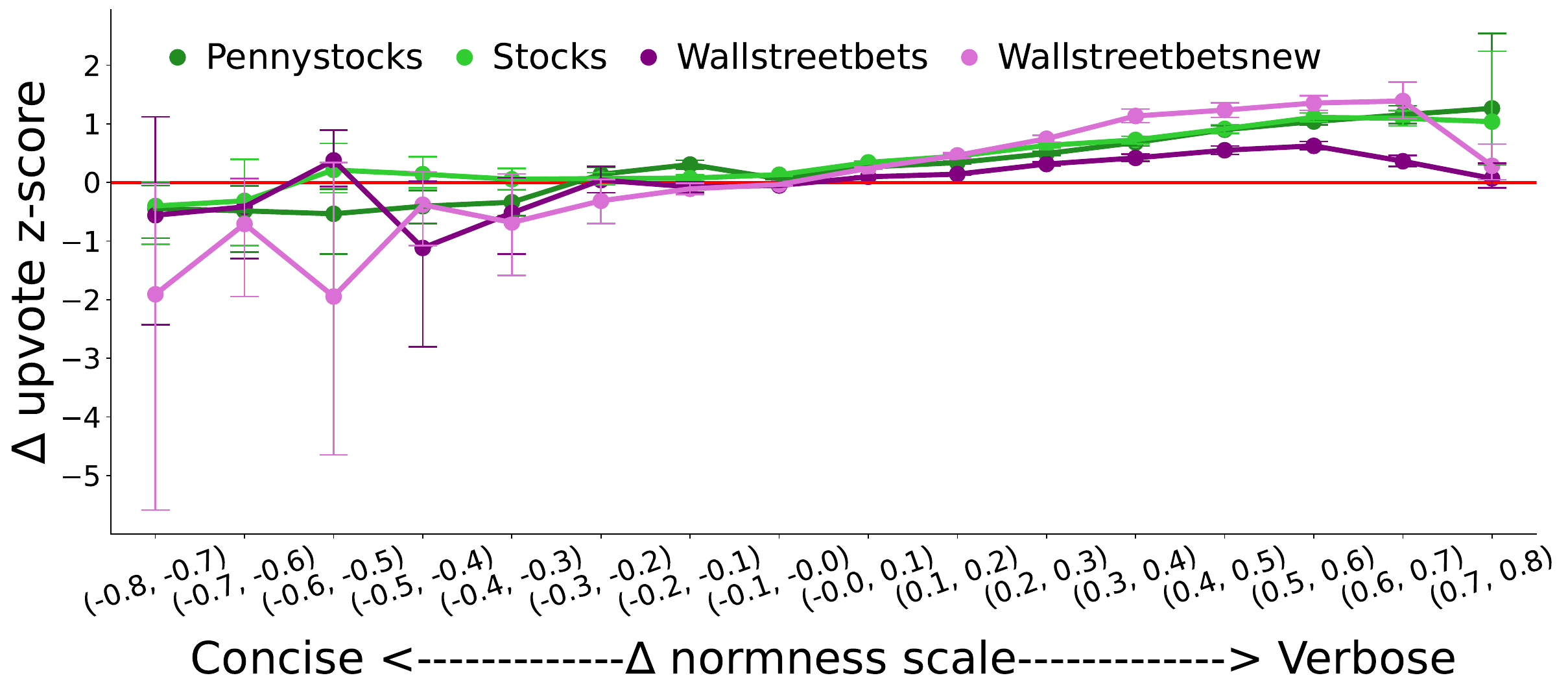}
    \caption{RPM plots for finance subreddits on the verbosity dimension.}
    \label{fig:rpm-finance-verbosity}
\end{figure}

\subsection{Politics Subreddits}
\begin{figure}[h!]
    \centering
    \includegraphics[width=\columnwidth]{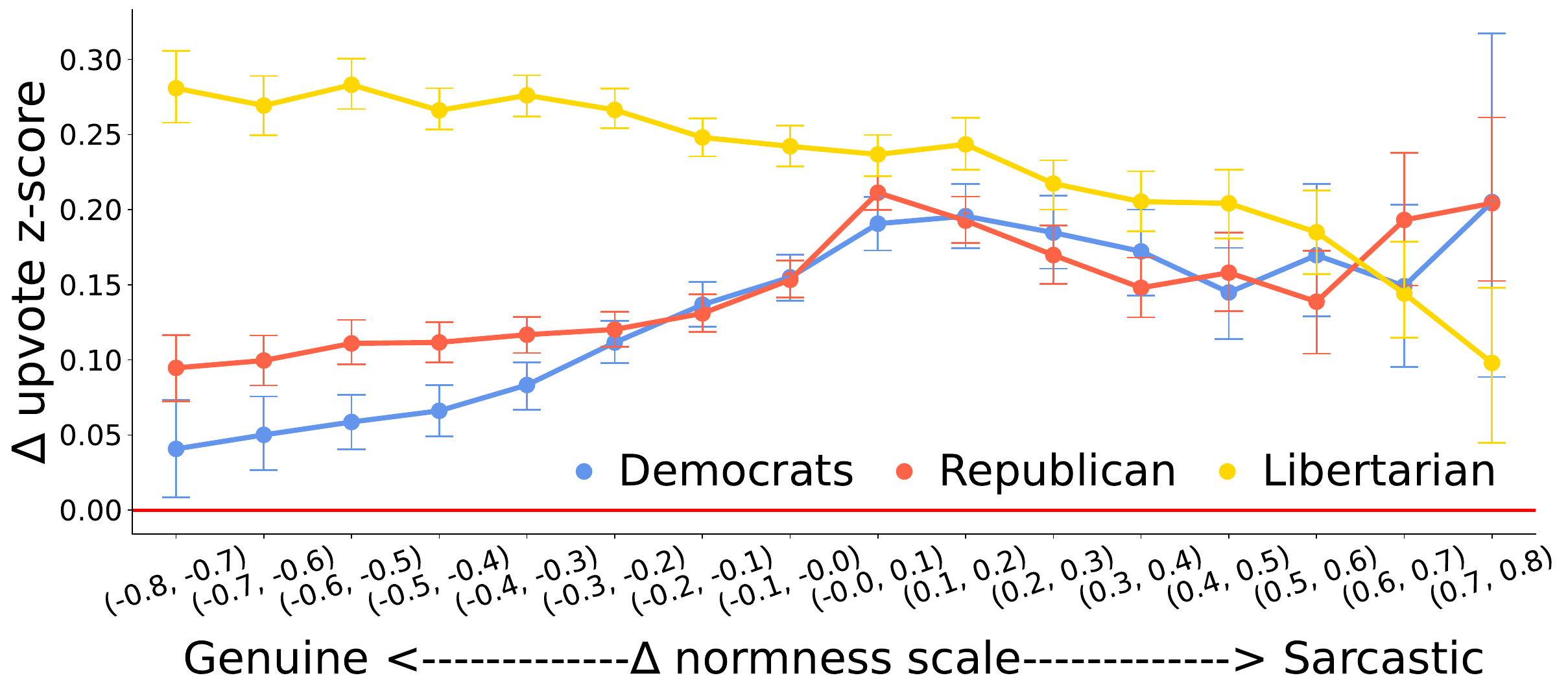}
    \caption{RPM plots for politics subreddits on the sarcasm dimension.}
    \label{fig:rpm-politics-sarcasm}
\end{figure}

\begin{figure}[h!]
    \centering
    \includegraphics[width=\columnwidth]{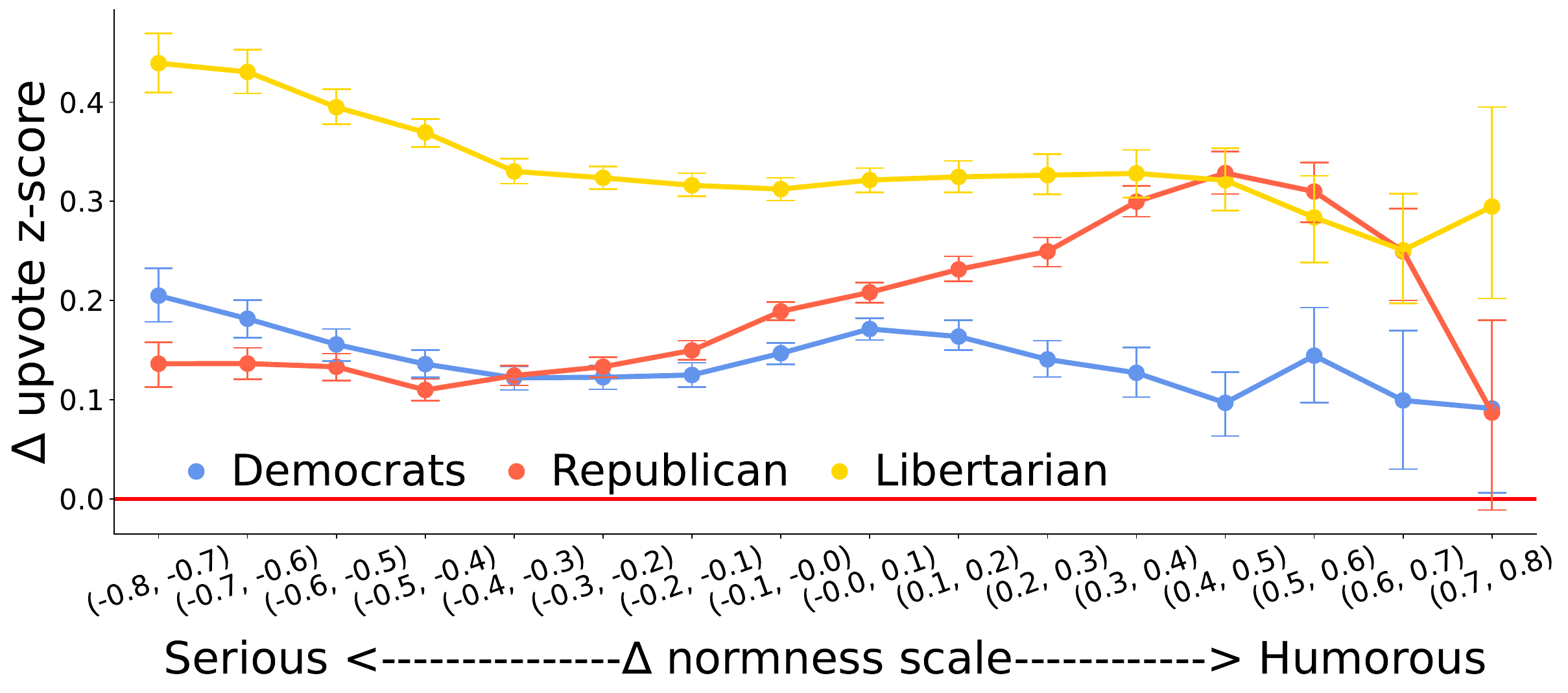}
    \caption{RPM plots for politics subreddits on the humor dimension.}
    \label{fig:rpm-politics-humor}
\end{figure}

We examined three subreddits related to politics: \texttt{r/democrats}, \texttt{r/republican}, and \texttt{r/libertarian}. The RPM plots reveal that \texttt{r/democrats} and \texttt{r/republican} exhibit similar preferences across multiple dimensions. Especially in genuine--sarcastic and serious--humorous, \texttt{r/democrats} and \texttt{r/republican} share high degrees of similarity while \texttt{r/libertarian} has its unique preferences (Figures \ref{fig:rpm-politics-sarcasm}--\ref{fig:rpm-politics-humor}). This similarity suggests that despite political differences, there are shared norms regarding the tone and style of discourse in these communities. Both subreddits also demonstrate moderate preferences for politeness and formality, indicating a mutual appreciation for respectful and well-mannered discussions.

In contrast, \texttt{r/libertarian} stands out with a strong preference for supportive, formal, and verbose comments, indicating a community that values thorough and well-structured discourse. This unique preference set suggests that \texttt{r/libertarian} places a higher emphasis on detailed and supportive interactions compared to the other two subreddits. These findings highlight the nuanced differences and similarities in community norms within political subreddits, with \texttt{r/democrats} and \texttt{r/republican} sharing many conversational norms, while \texttt{r/libertarian} adopts a distinctively detailed and supportive approach to ore structured discourse.

\begin{figure}[h!]
    \centering
    \includegraphics[width=\columnwidth]{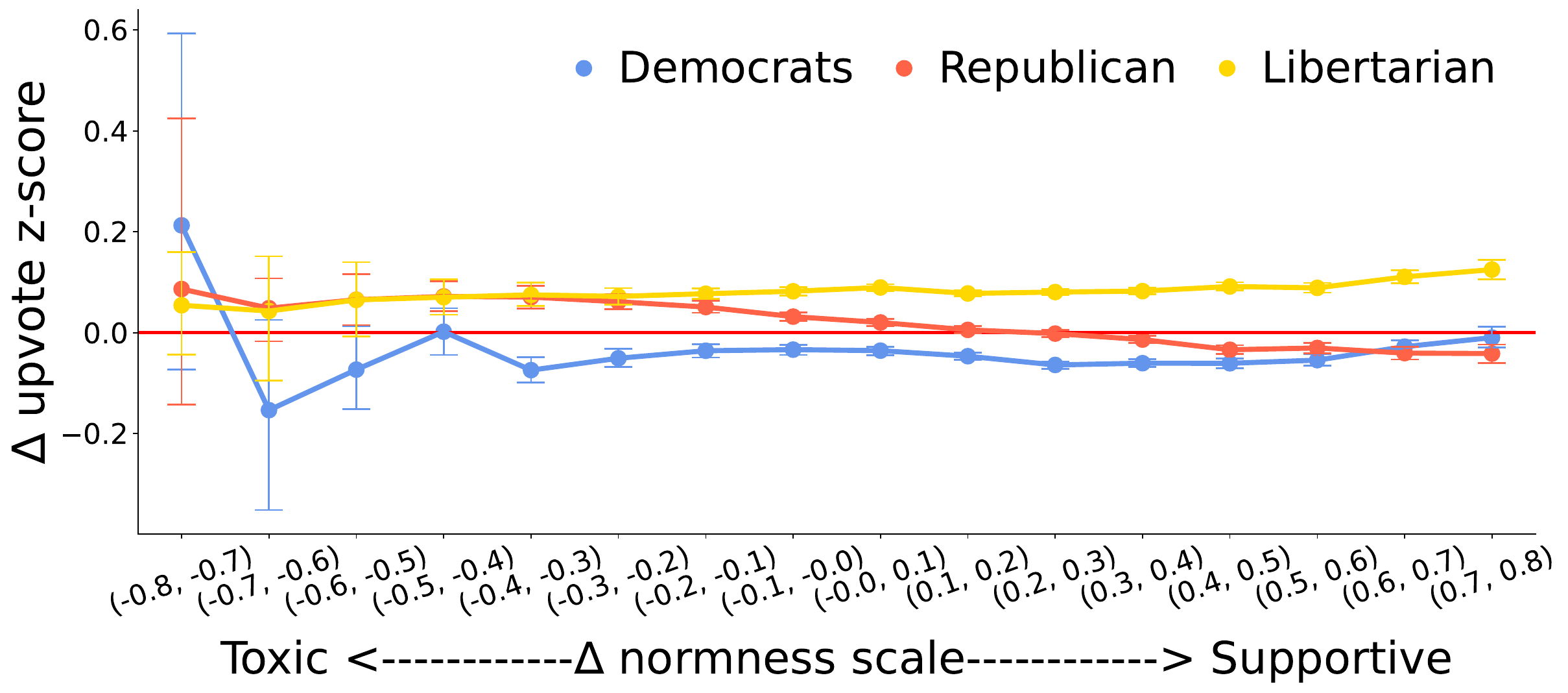}
    \caption{RPM plots for politics subreddits on the supportiveness dimension.}
    \label{fig:rpm-politics-supportiveness}
\end{figure}

\begin{figure}[h!]
    \centering
    \includegraphics[width=\columnwidth]{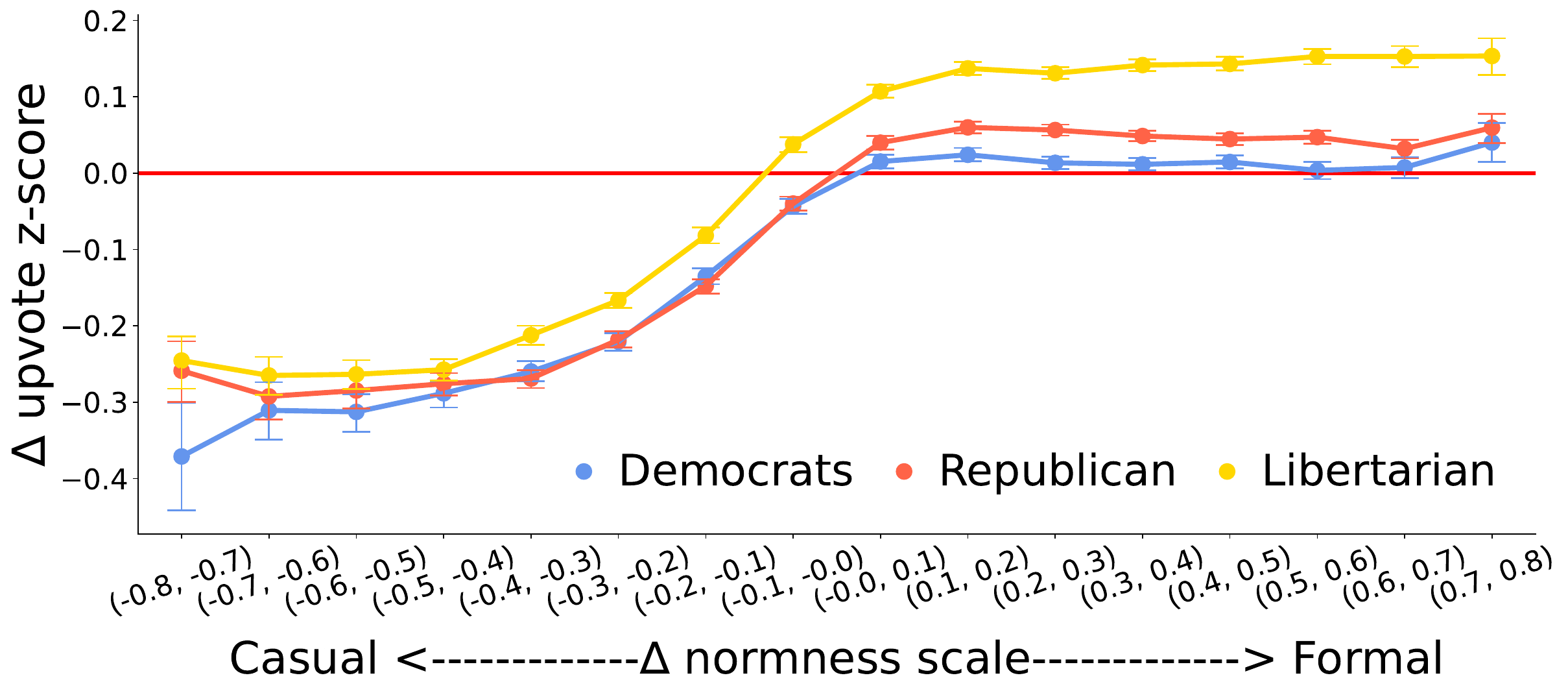}
    \caption{RPM plots for politics subreddits on the formality dimension.}
    \label{fig:rpm-politics-formality}
\end{figure}

\begin{figure}[h!]
    \centering
    \includegraphics[width=\columnwidth]{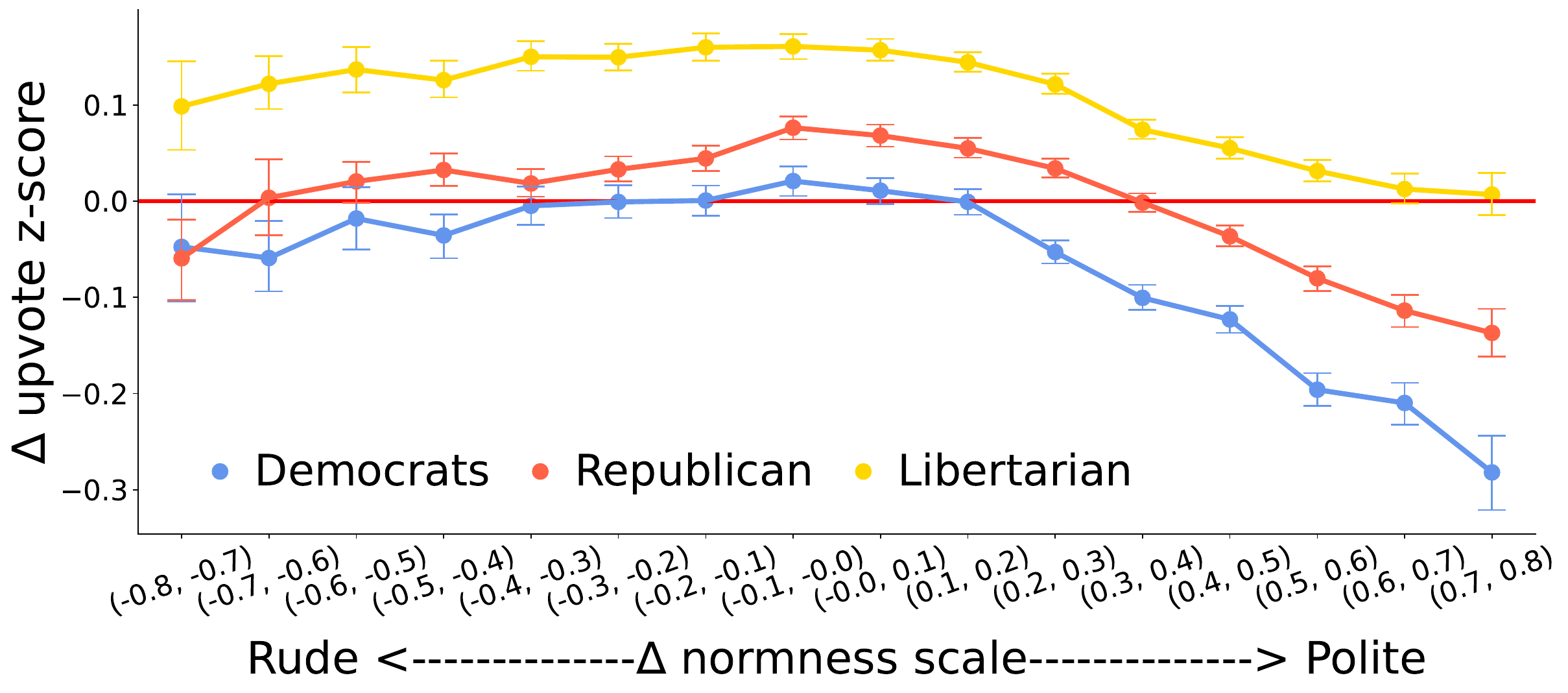}
    \caption{RPM plots for politics subreddits on the politeness dimension.}
    \label{fig:rpm-politics-politeness}
\end{figure}

\begin{figure}[h!]
    \centering
    \includegraphics[width=\columnwidth]{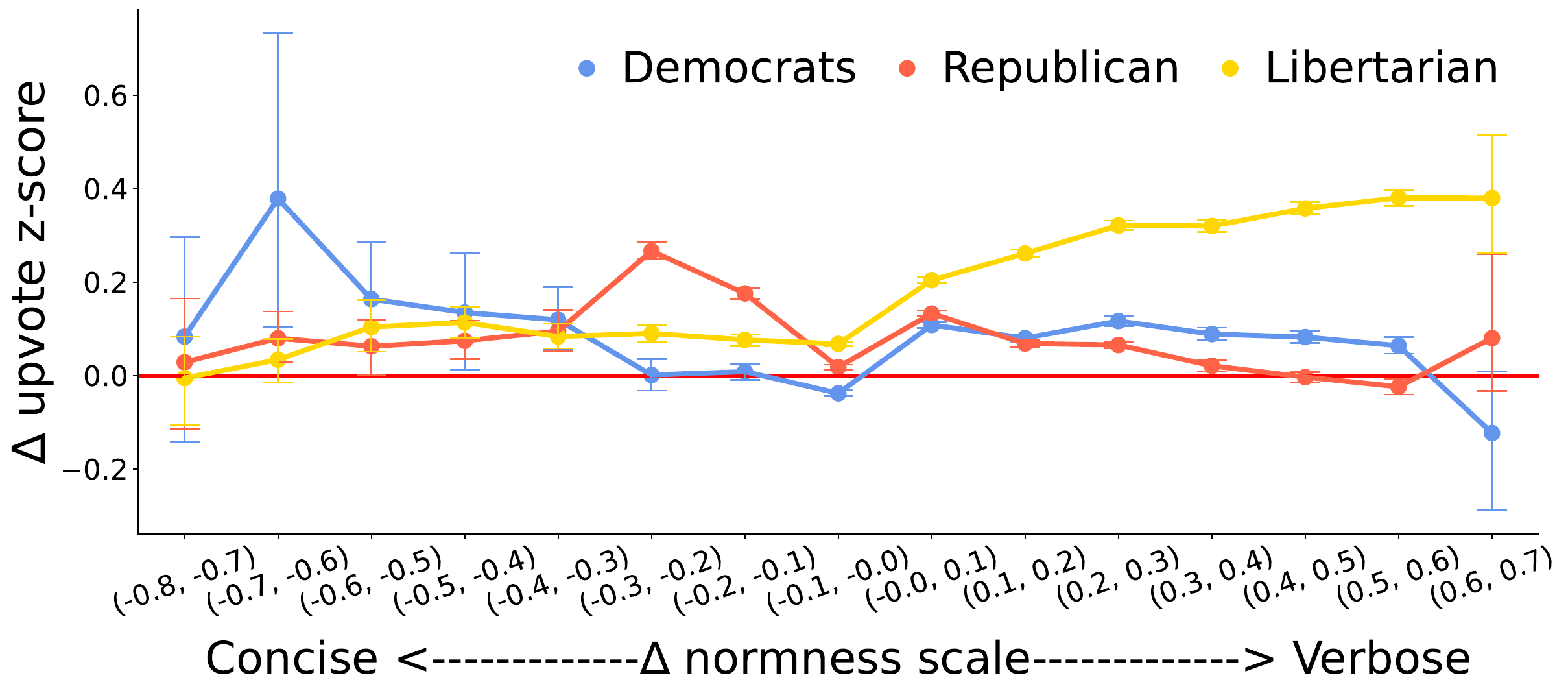}
    \caption{RPM plots for politics subreddits on the verbosity dimension.}
    \label{fig:rpm-politics-verbosity}
\end{figure}

\newpage
\subsection{Science Subreddits}
\begin{figure}[h!]
    \centering
    \includegraphics[width=\columnwidth]{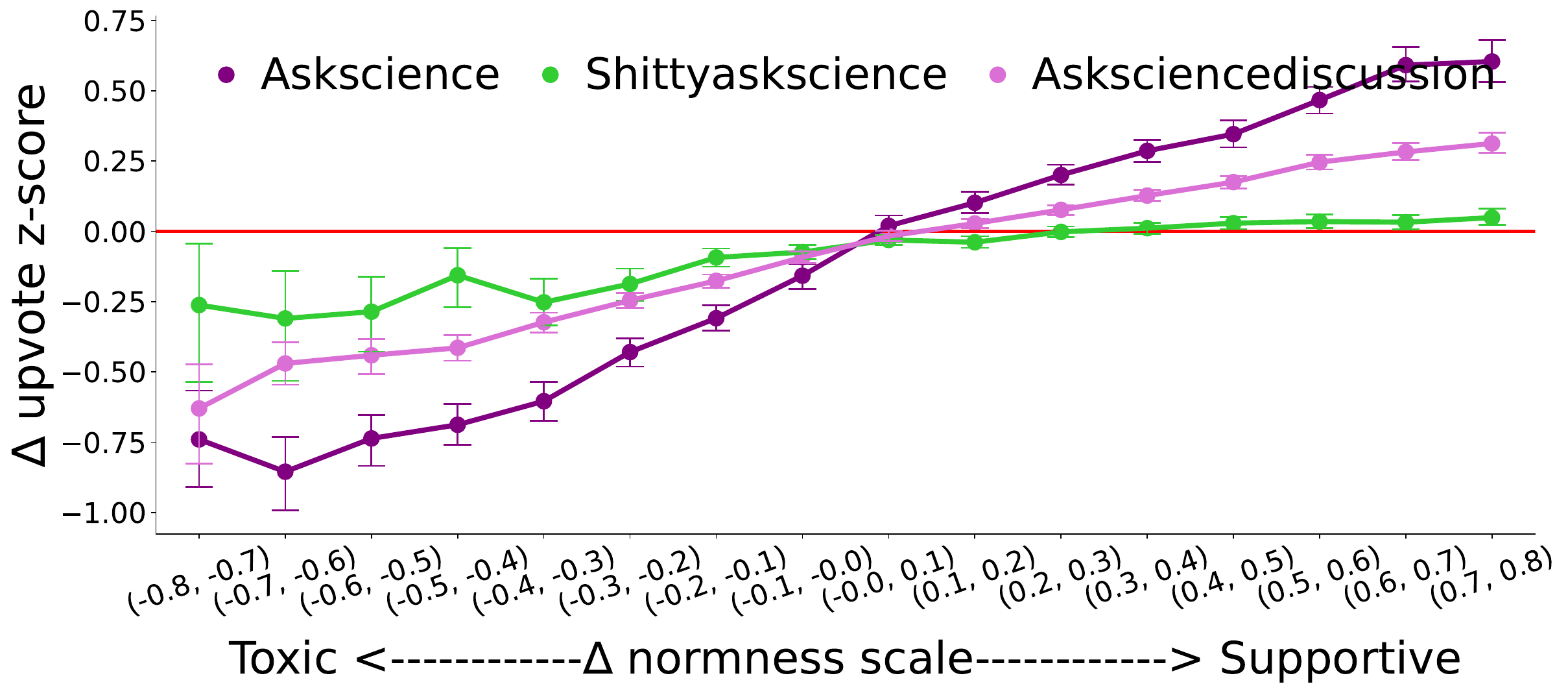}
    \caption{RPM plots for science subreddits on the supportiveness dimension.}
    \label{fig:rpm-science-supportiveness}
\end{figure}

\begin{figure}
    \centering
    \includegraphics[width=\columnwidth]{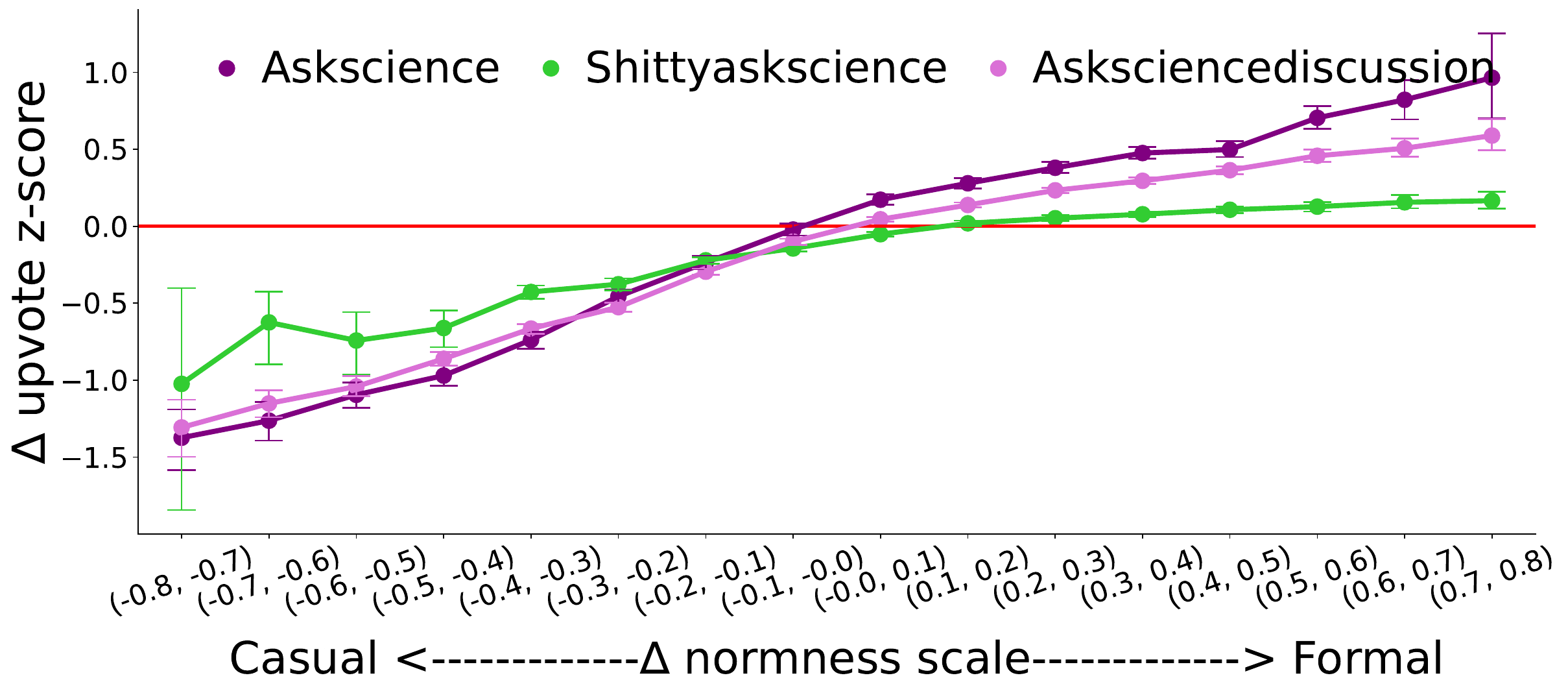}
    \caption{RPM plots for science subreddits on the formality dimension.}
    \label{fig:rpm-science-formality}
\end{figure}

\begin{figure}
    \centering
    \includegraphics[width=\columnwidth]{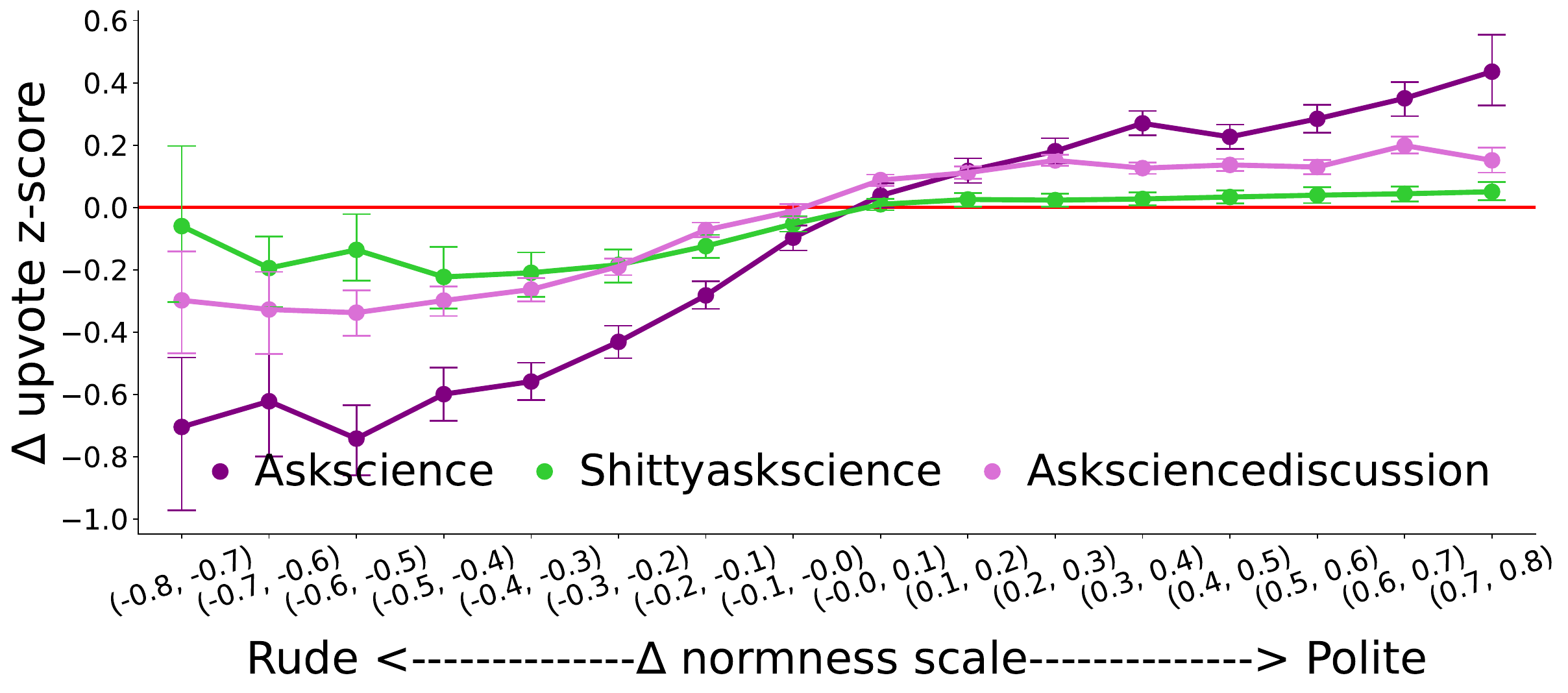}
    \caption{RPM plots for science subreddits on the politeness dimension.}
    \label{fig:rpm-science-politeness}
\end{figure}

\begin{figure}
    \centering
    \includegraphics[width=\columnwidth]{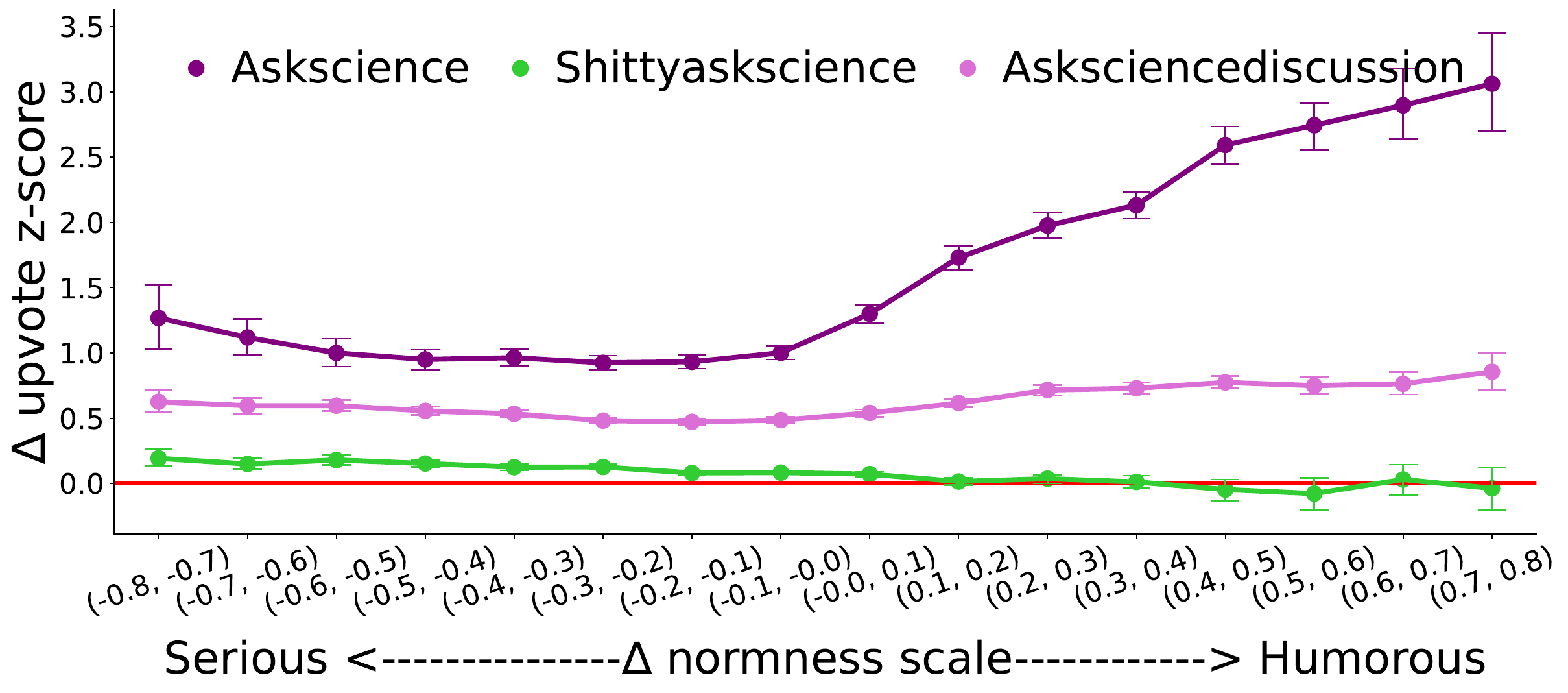}
    \caption{RPM plots for science subreddits on the humor dimension.}
    \label{fig:rpm-science-humor}
\end{figure}

We examined three subreddits related to science: \texttt{r/askscience}, \texttt{r/shittyaskscience}, and \texttt{r/asksciencediscussion}. First, it is notable that \texttt{r/shittyaskscience} shows very weak preference across all norm dimensions, and relatively weaker disapproval than the other two subreddits. \texttt{r/shittyaskscience} disproves of toxic, casual and rude comments to a lesser extent than \texttt{r/askscience} and \texttt{r/asksciencediscussion} (Figures \ref{fig:rpm-science-supportiveness}--\ref{fig:rpm-science-politeness}). This could be attributed to the fact that it is a spin-off subreddit created to mock \texttt{r/askscience}, which makes it more tolerant to toxic, casual or rude comments. On the other hand, the three subreddits have similar overall preference patterns despite different magnitudes, except for \texttt{r/askscience} in the serious--humorous norm dimension. As shown in \Fref{fig:rpm-science-humor}, \texttt{r/askscience} exhibits significantly stronger preference for humorous comments compared to the other two subreddits. While seemingly counter-intuitive, since \texttt{r/askscience} is a tightly moderateed subreddit, but its preference for humorous data implies that comments that both adhere to the subreddit rules \emph{and} humorous would typically be a high quality comment preferred by community members.

\begin{figure}
    \centering
    \includegraphics[width=\columnwidth]{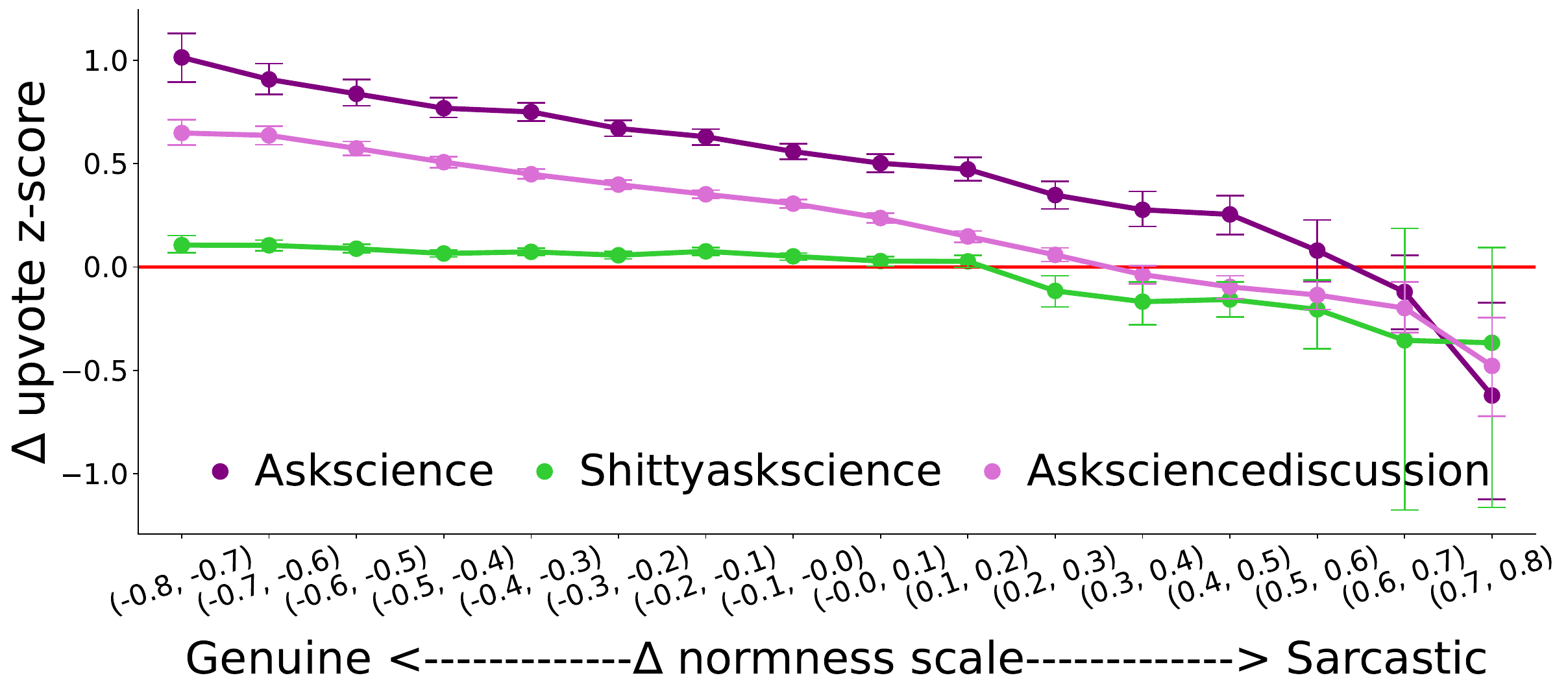}
    \caption{RPM plots for science subreddits on the sarcasm dimension.}
    \label{fig:rpm-science-sarcasm}
\end{figure}

\begin{figure}
    \centering
    \includegraphics[width=\columnwidth]{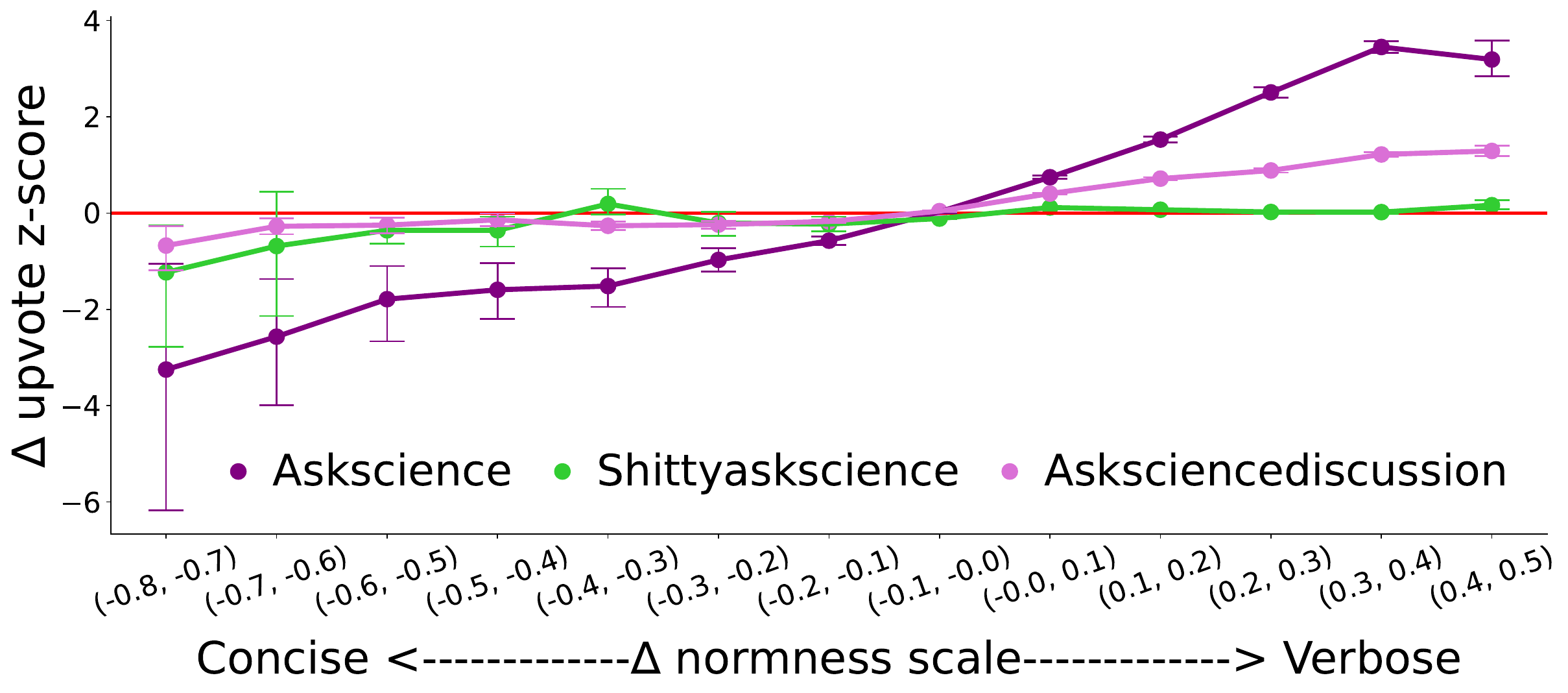}
    \caption{RPM plots for science subreddits on the verbosity dimension.}
    \label{fig:rpm-science-verbosity}
\end{figure}

\section{RPM Plots with Original Comments}\label{appendix:rpm-plots-original}
RPM plots that feature only the original comments. Unlike standard RPM plots, which display the difference in normness scale and z-score between original and style-transferred comments, these plots use the absolute values: normness scale on the x-axis and z-score on the y-axis.

Although these RPM plots don't show how changes in normness influence community approval, they do provide insight into the average normness of comments across various communities.

\subsection{Gender Subreddits}

\begin{figure}[!h]
    \centering
    \includegraphics[width=\columnwidth]{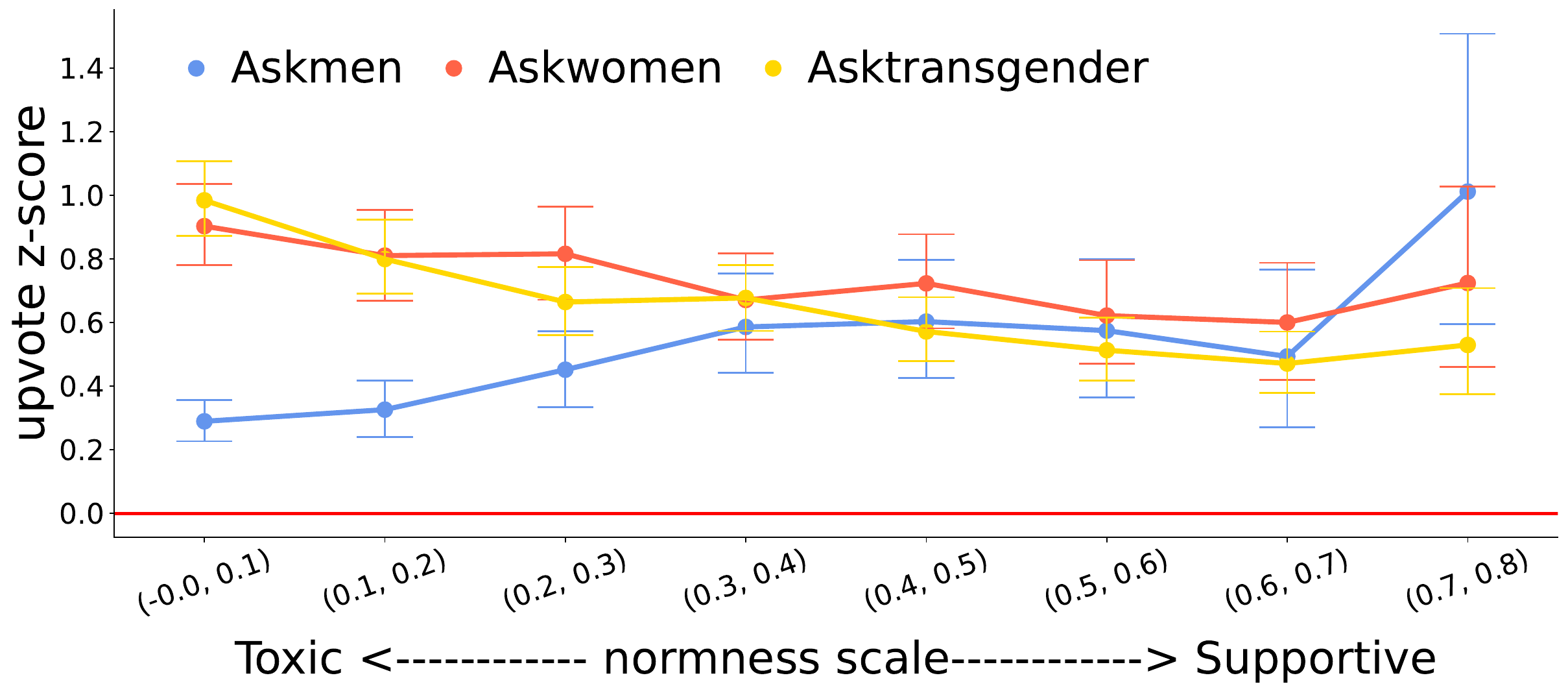}
    \caption{RPM plots for gender subreddits on the supportiveness dimension.}
    \label{fig:rpm-gender-supportiveness-original}
\end{figure}

\begin{figure}[!h]
    \centering
    \includegraphics[width=\columnwidth]{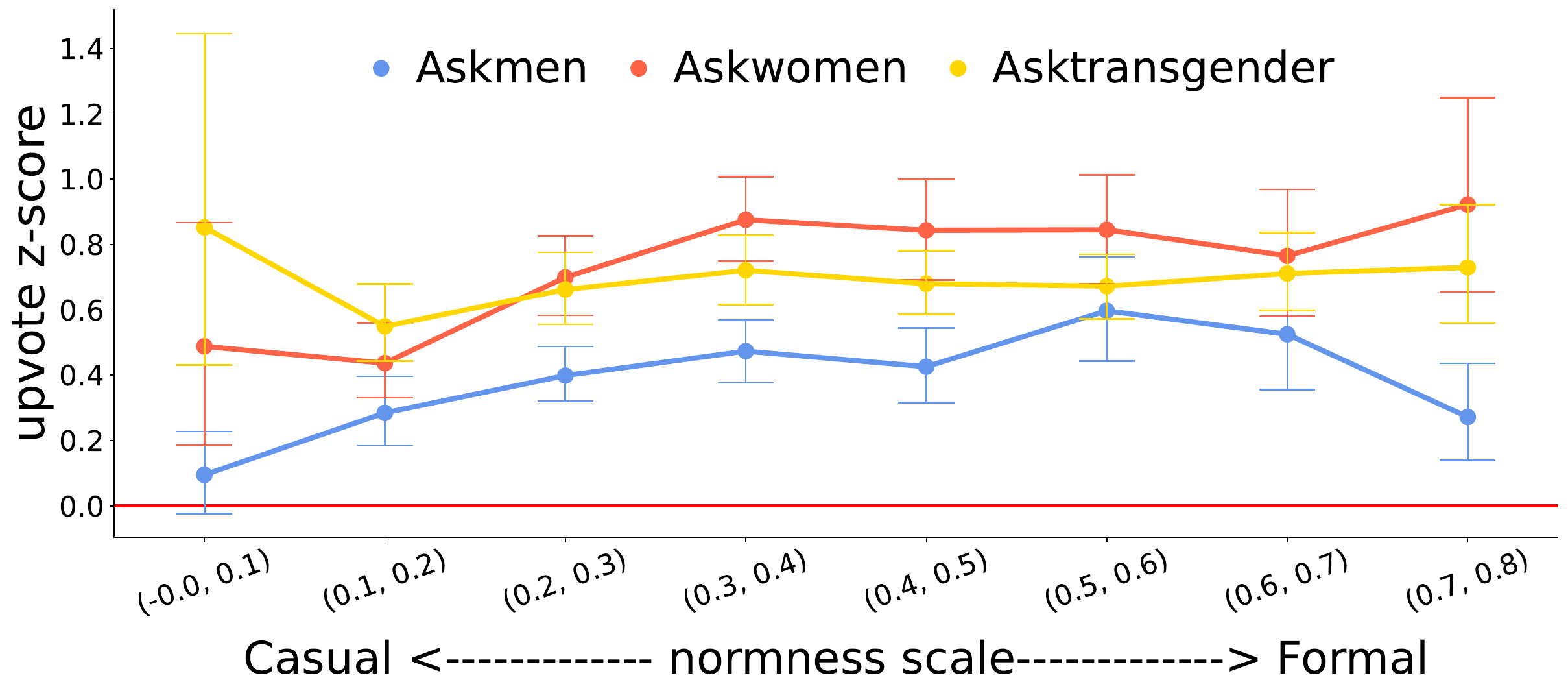}
    \caption{RPM plots for gender subreddits on the formality dimension.}
    \label{fig:rpm-gender-formality-original}
\end{figure}

\begin{figure}[!h]
    \centering
    \includegraphics[width=\columnwidth]{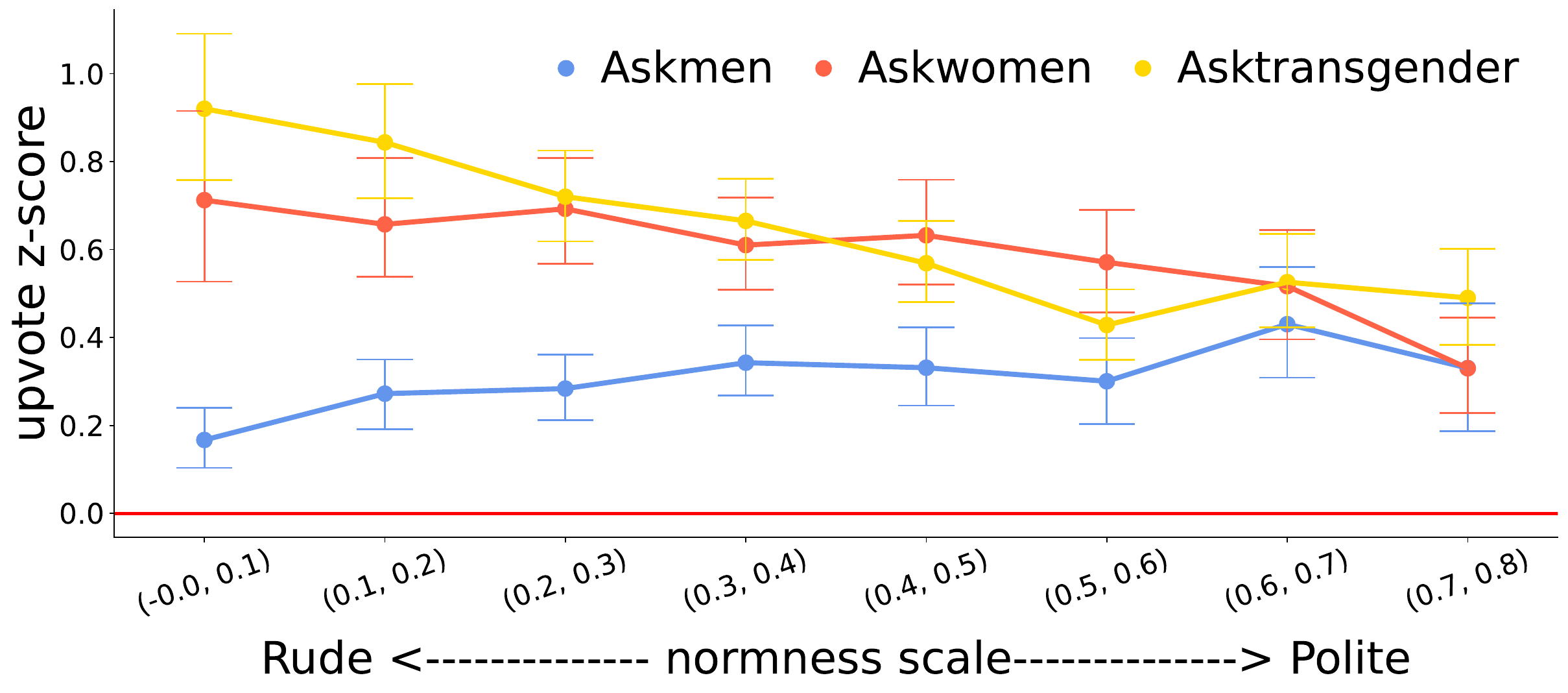}
    \caption{RPM plots for gender subreddits on the politeness dimension.}
    \label{fig:rpm-gender-politeness-original}
\end{figure}

\begin{figure}[!h]
    \centering
    \includegraphics[width=\columnwidth]{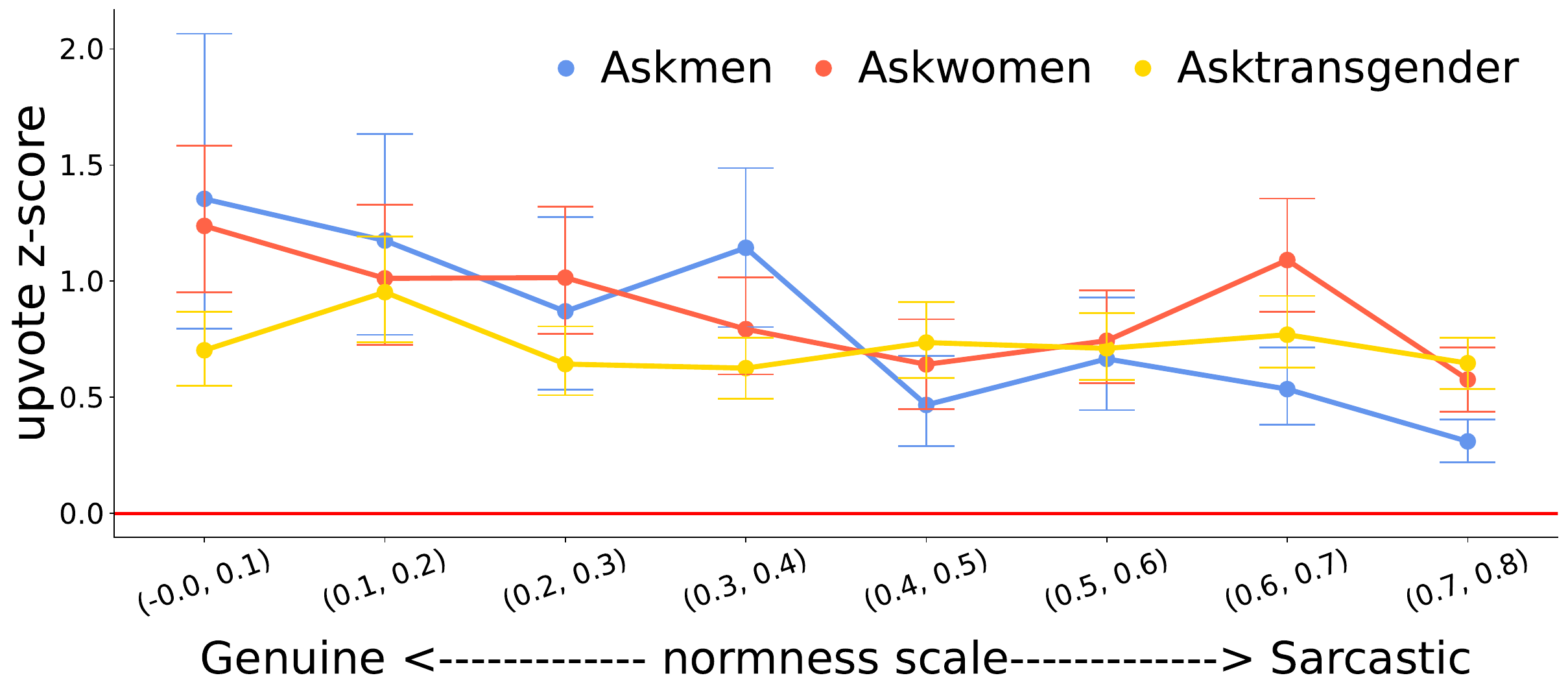}
    \caption{RPM plots for gender subreddits on the sarcasm dimension.}
    \label{fig:rpm-gender-sarcasm-original}
\end{figure}

\begin{figure}[!h]
    \centering
    \includegraphics[width=\columnwidth]{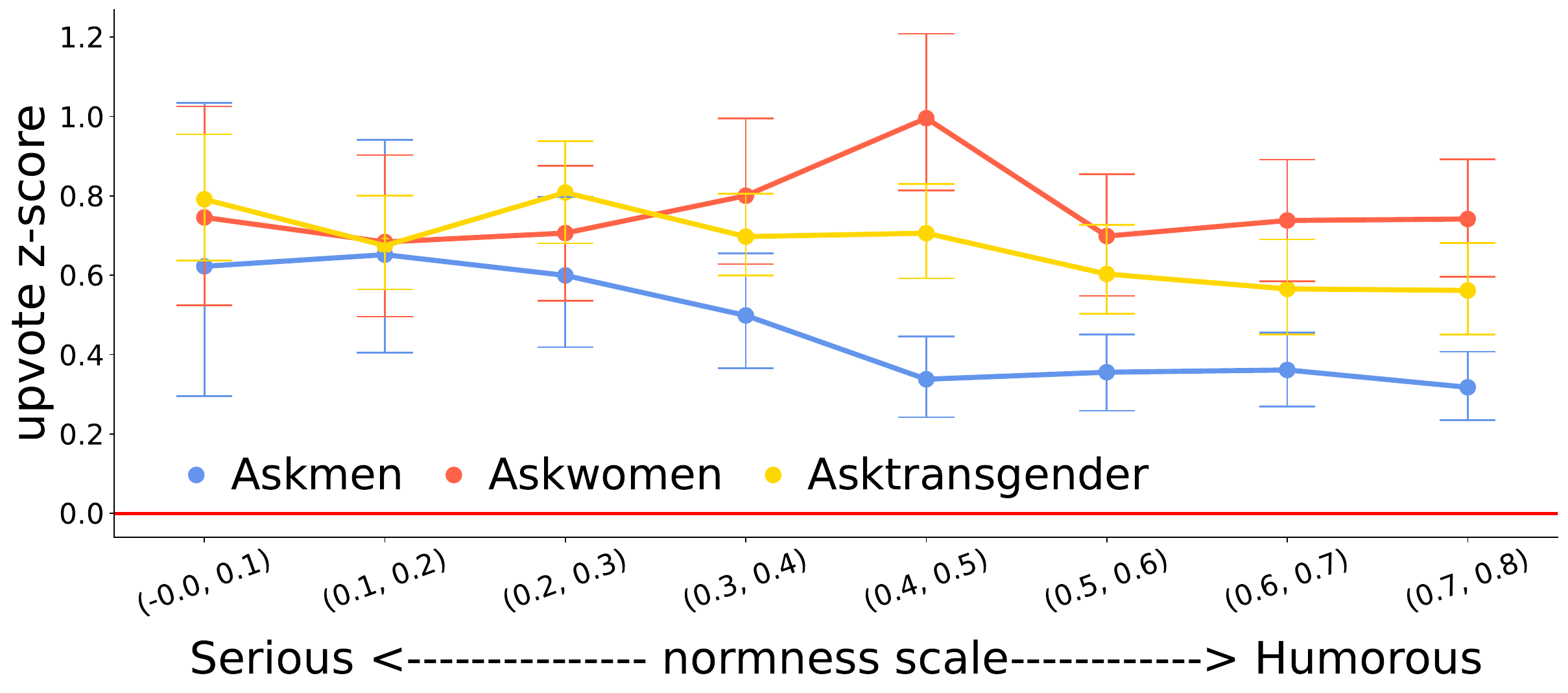}
    \caption{RPM plots for gender subreddits on the humor dimension.}
    \label{fig:rpm-gender-humor-original}
\end{figure}

\begin{figure}[!h]
    \centering
    \includegraphics[width=\columnwidth]{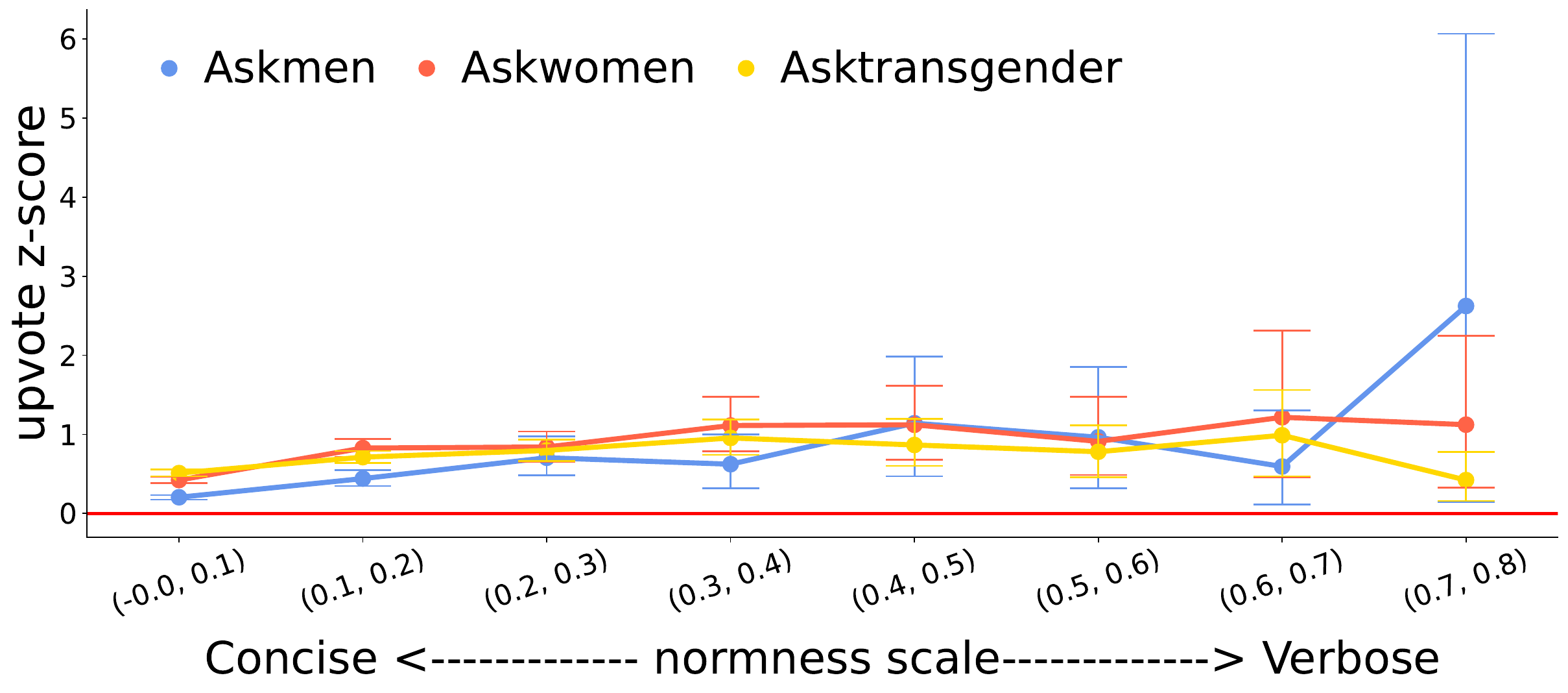}
    \caption{RPM plots for gender subreddits on the verbosity dimension.}
    \label{fig:rpm-gender-verbosity-original}
\end{figure}

\newpage
\subsection{Finance Subreddits}

\begin{figure}[!h]
    \centering
    \includegraphics[width=\columnwidth]{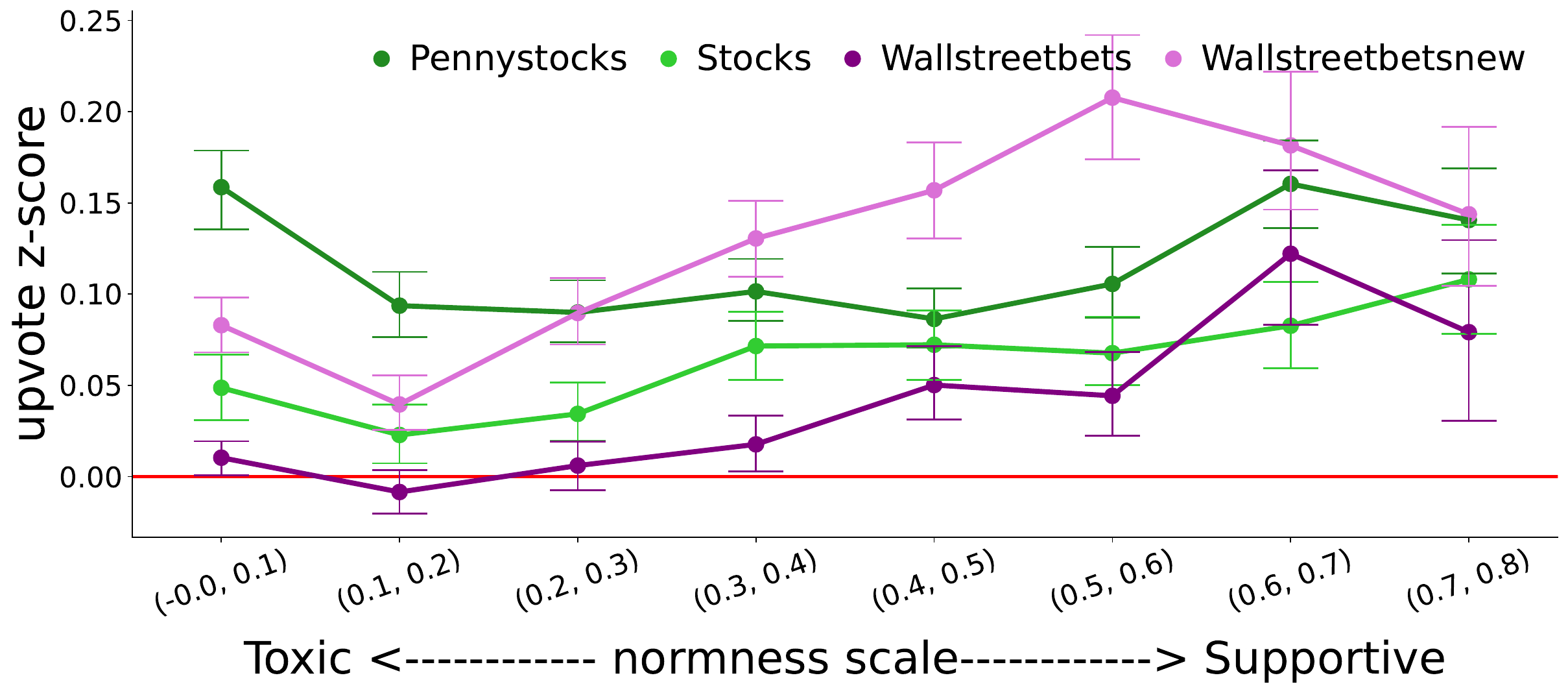}
    \caption{RPM plots for finance subreddits on the supportiveness dimension.}
    \label{fig:rpm-finance-supportiveness-original}
\end{figure}
\newpage

\begin{figure}[h!]
    \centering
    \includegraphics[width=\columnwidth]{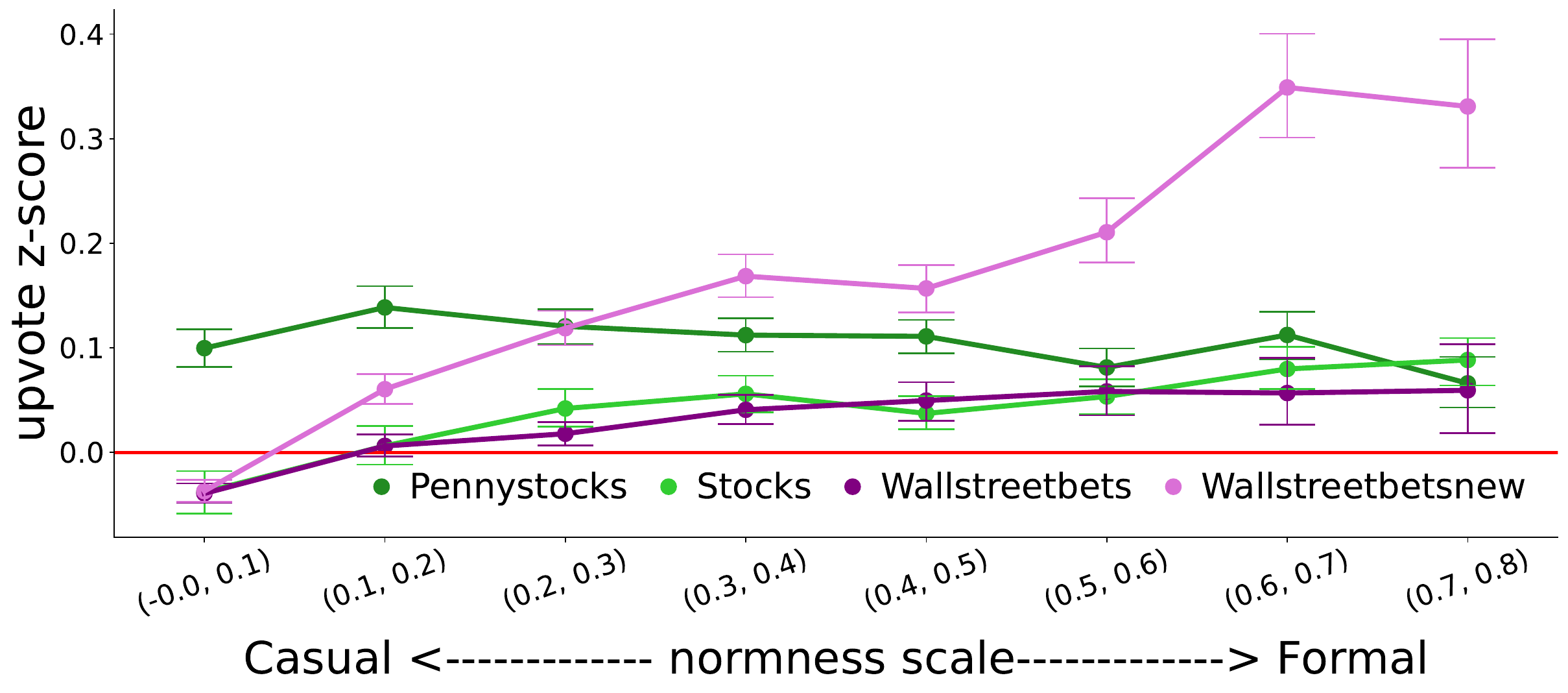}
    \caption{RPM plots for finance subreddits on the formality dimension.}
    \label{fig:rpm-finance-formality-original}
\end{figure}

\begin{figure}[h!]
    \centering
    \includegraphics[width=\columnwidth]{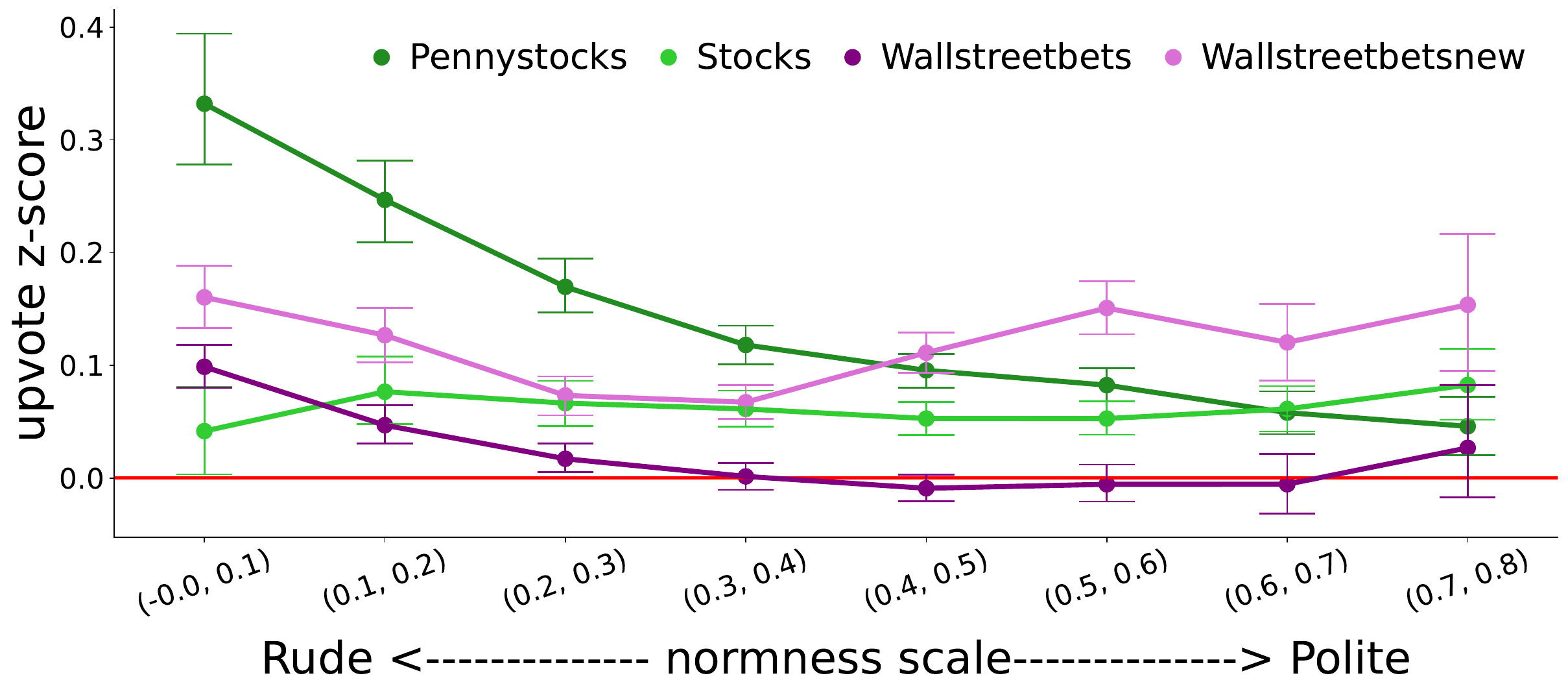}
    \caption{RPM plots for finance subreddits on the politeness dimension.}
    \label{fig:rpm-finance-politeness-original}
\end{figure}

\begin{figure}[h!]
    \centering
    \includegraphics[width=\columnwidth]{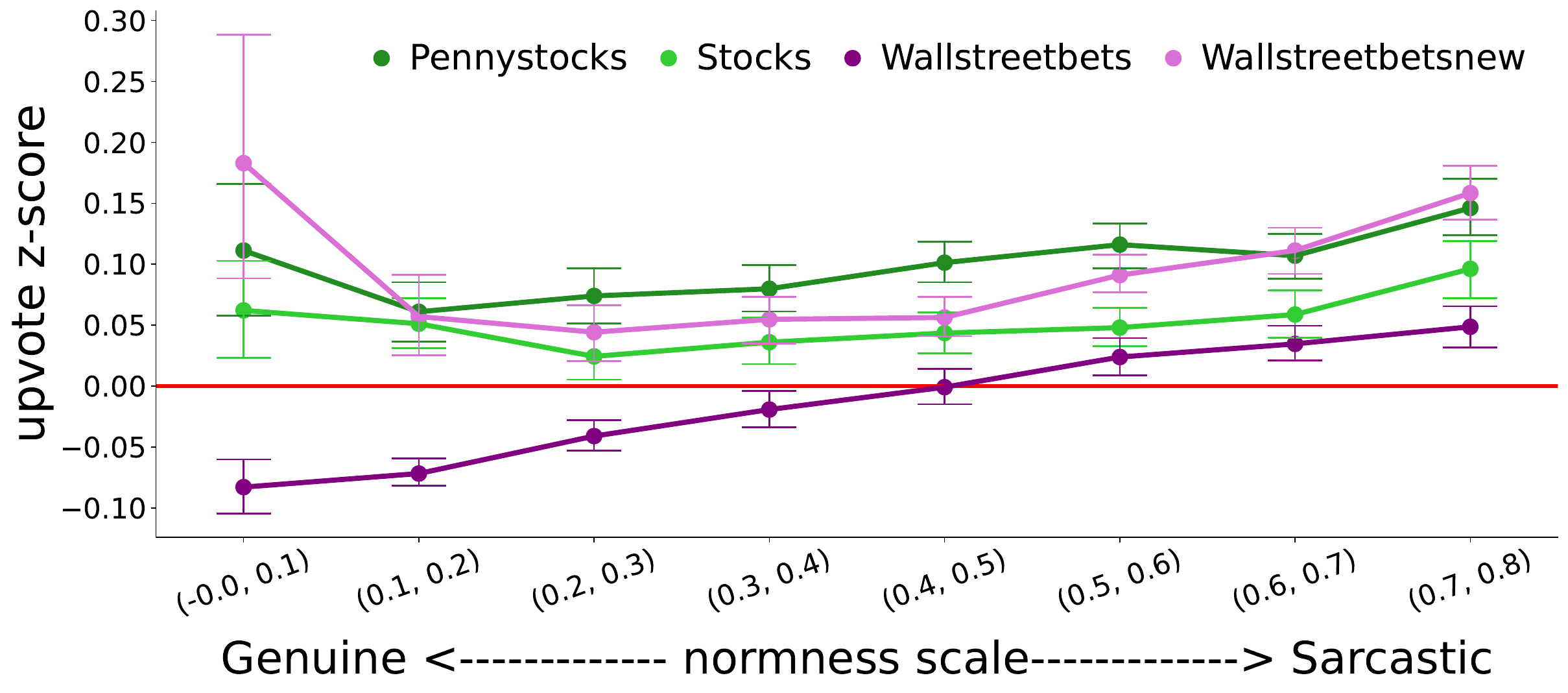}
    \caption{RPM plots for finance subreddits on the sarcasm dimension.}
    \label{fig:rpm-finance-sarcasm-original}
\end{figure}

\begin{figure}[h!]
    \centering
    \includegraphics[width=\columnwidth]{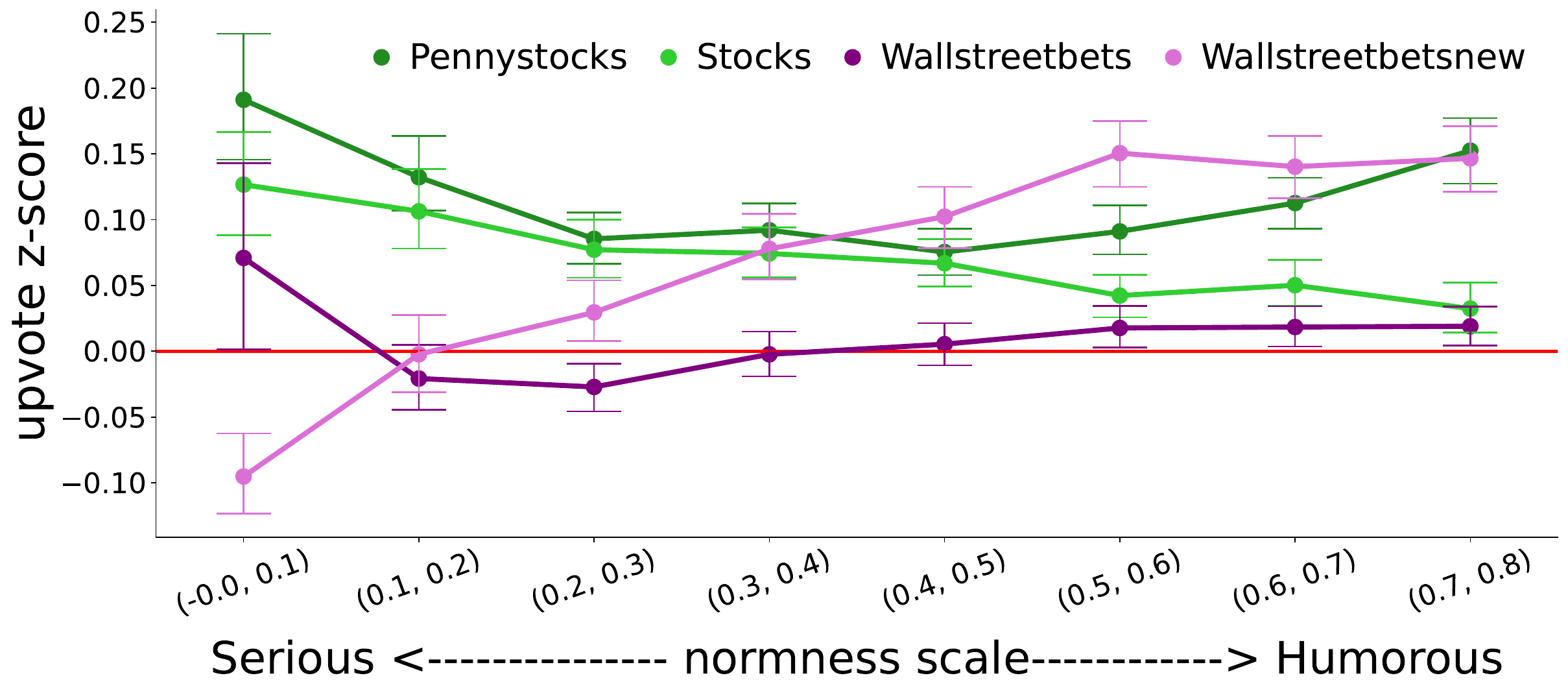}
    \caption{RPM plots for finance subreddits on the humor dimension.}
    \label{fig:rpm-finance-humor-original}
\end{figure}

\begin{figure}[h]
    \centering
    \includegraphics[width=\columnwidth]{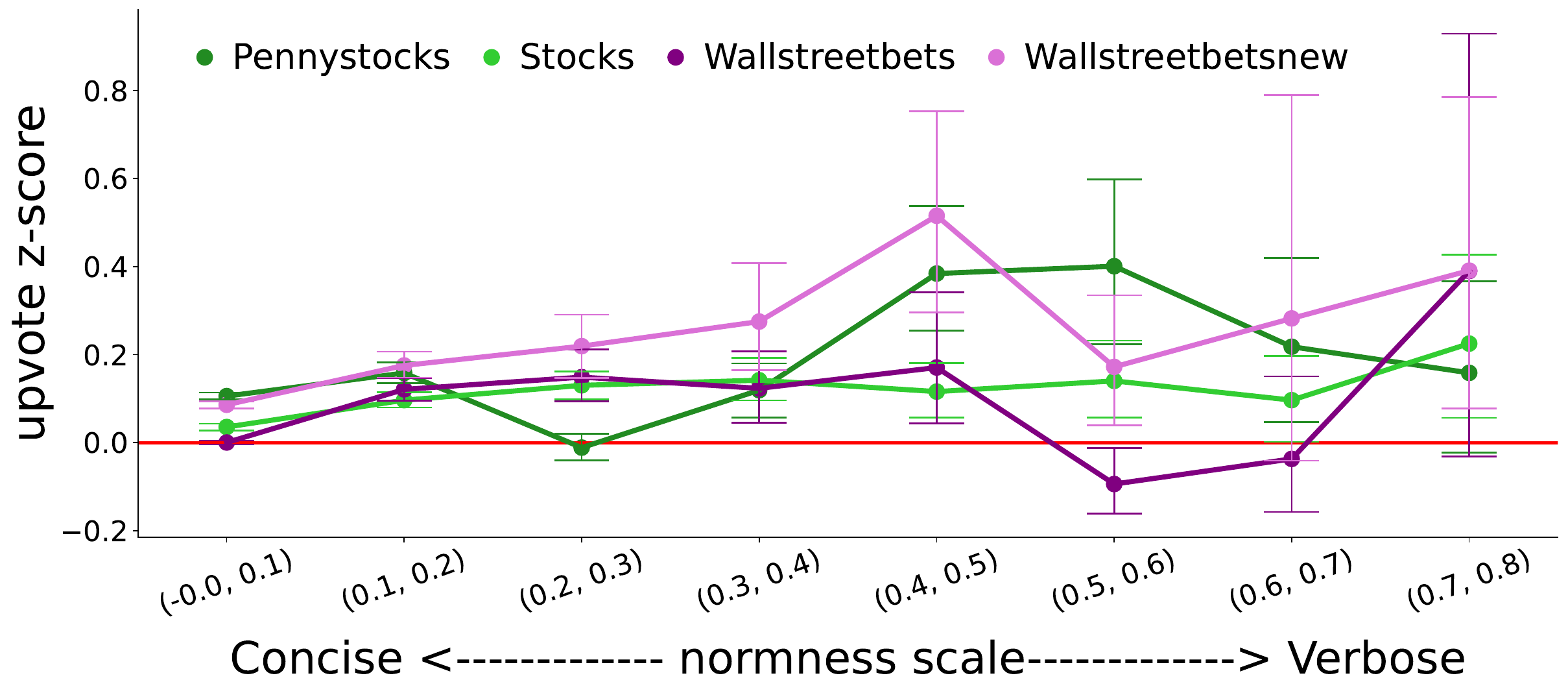}
    \caption{RPM plots for finance subreddits on the verbosity dimension.}
    \label{fig:rpm-finance-verbosity-original}
\end{figure}

\newpage
\subsection{Politics Subreddits}

\begin{figure}[h]
    \centering
    \includegraphics[width=\columnwidth]{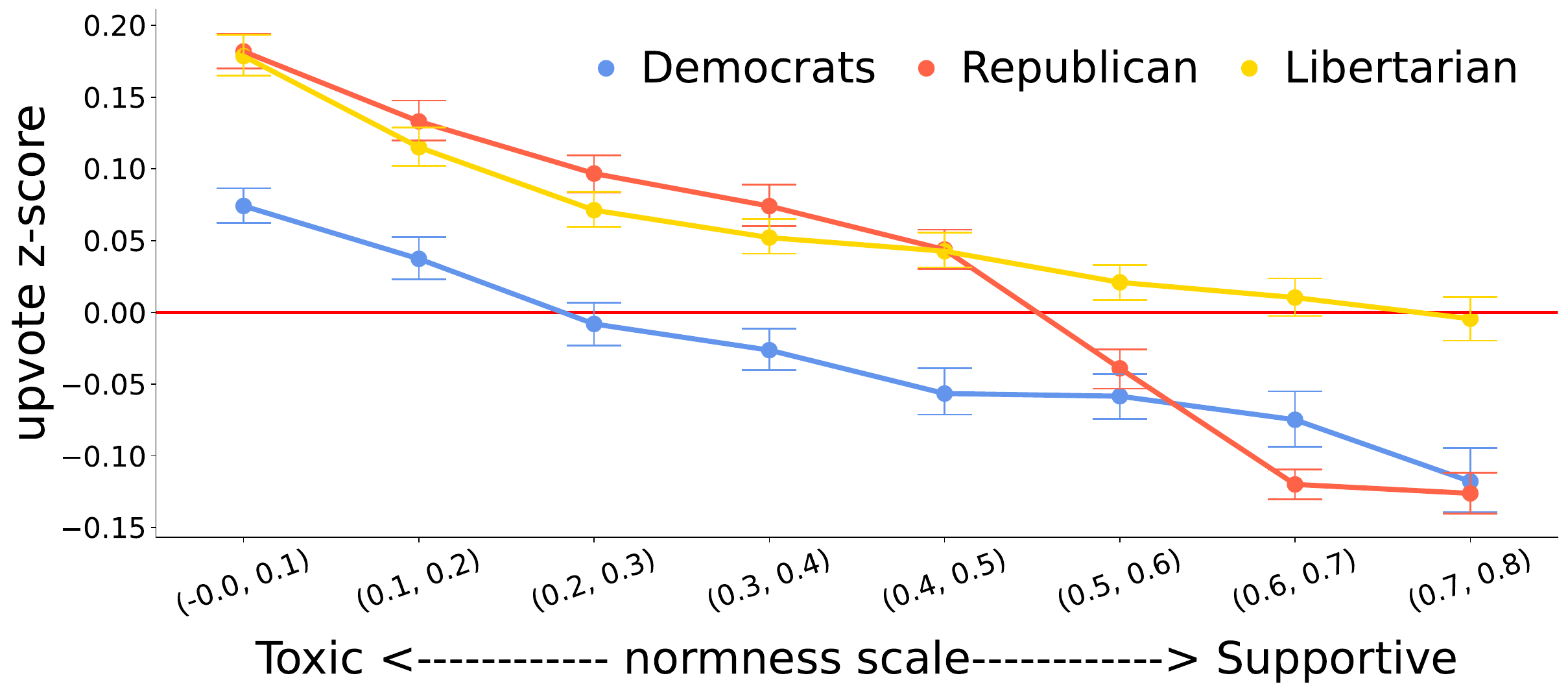}
    \caption{RPM plots for politics subreddits on the supportiveness dimension.}
    \label{fig:rpm-politics-supportiveness-original}
\end{figure}

\begin{figure}[h!]
    \centering
    \includegraphics[width=\columnwidth]{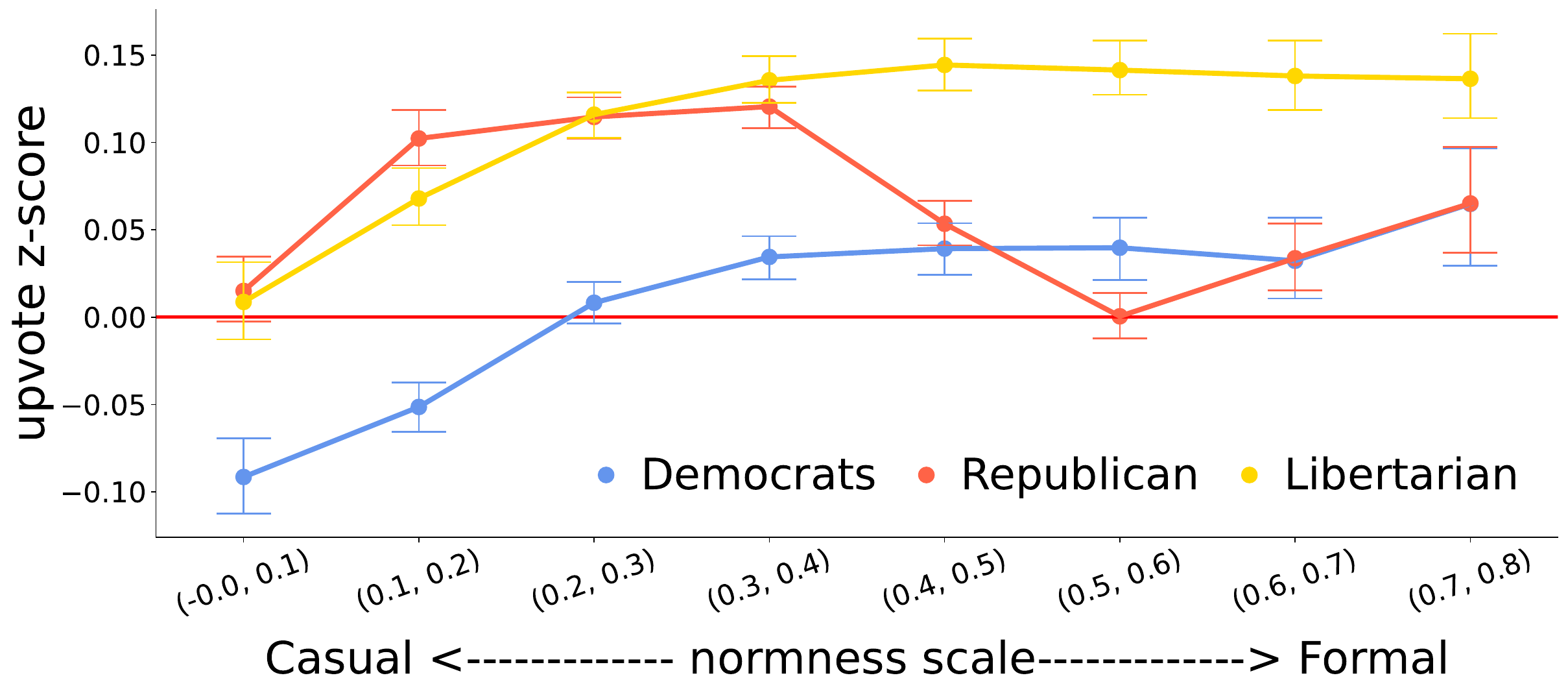}
    \caption{RPM plots for politics subreddits on the formality dimension.}
    \label{fig:rpm-politics-formality-original}
\end{figure}

\begin{figure}[h!]
    \centering
    \includegraphics[width=\columnwidth]{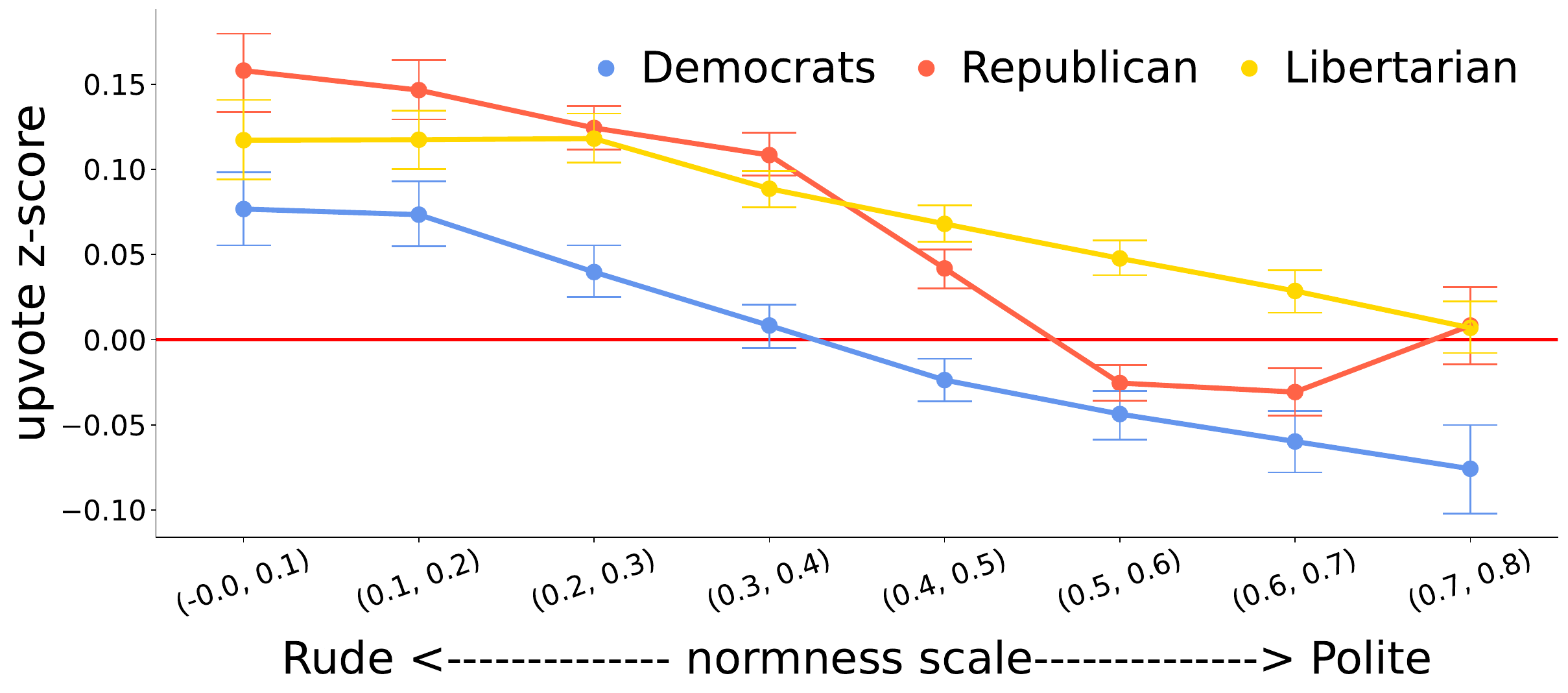}
    \caption{RPM plots for politics subreddits on the politeness dimension.}
    \label{fig:rpm-politics-politeness-original}
\end{figure}
\newpage

\begin{figure}[h!]
    \centering
    \includegraphics[width=\columnwidth]{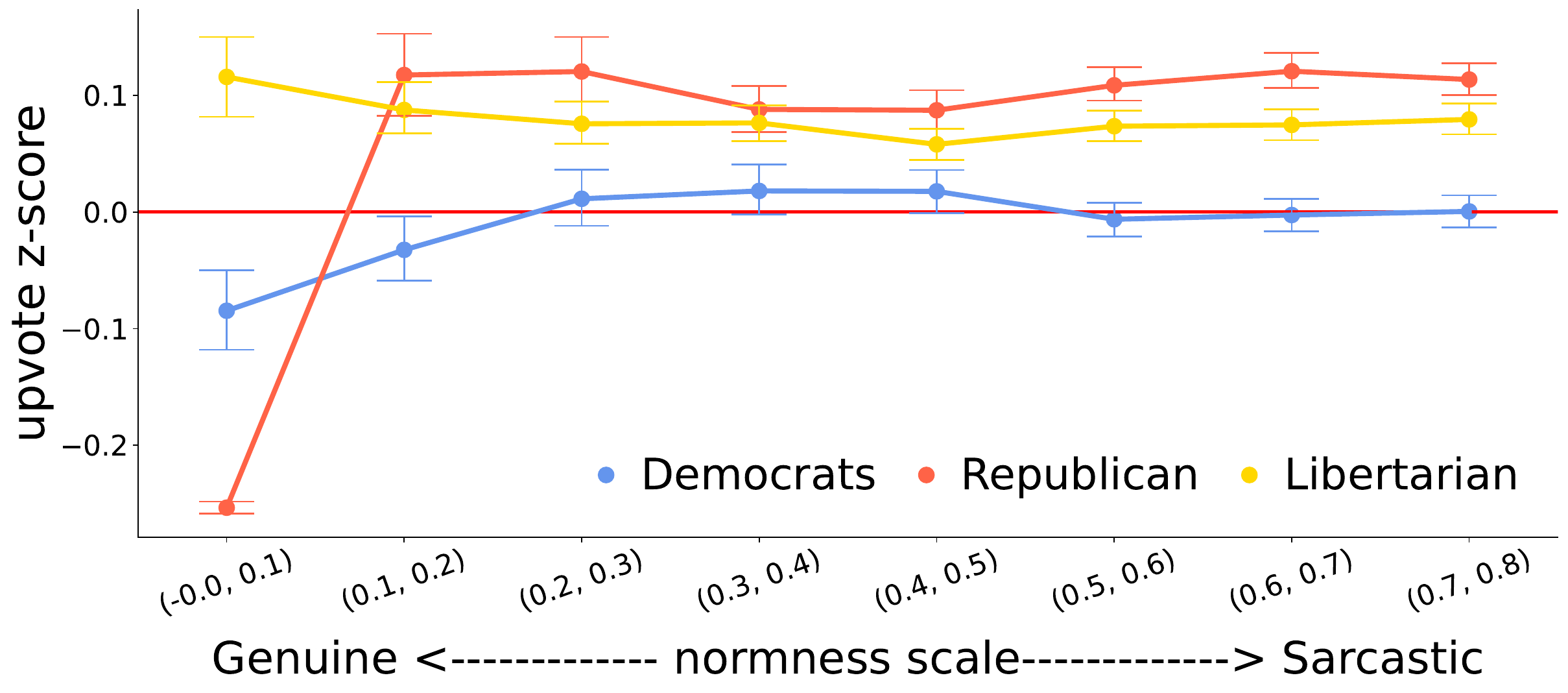}
    \caption{RPM plots for politics subreddits on the sarcasm dimension.}
    \label{fig:rpm-politics-sarcasm-original}
\end{figure}

\begin{figure}[h!]
    \centering
    \includegraphics[width=\columnwidth]{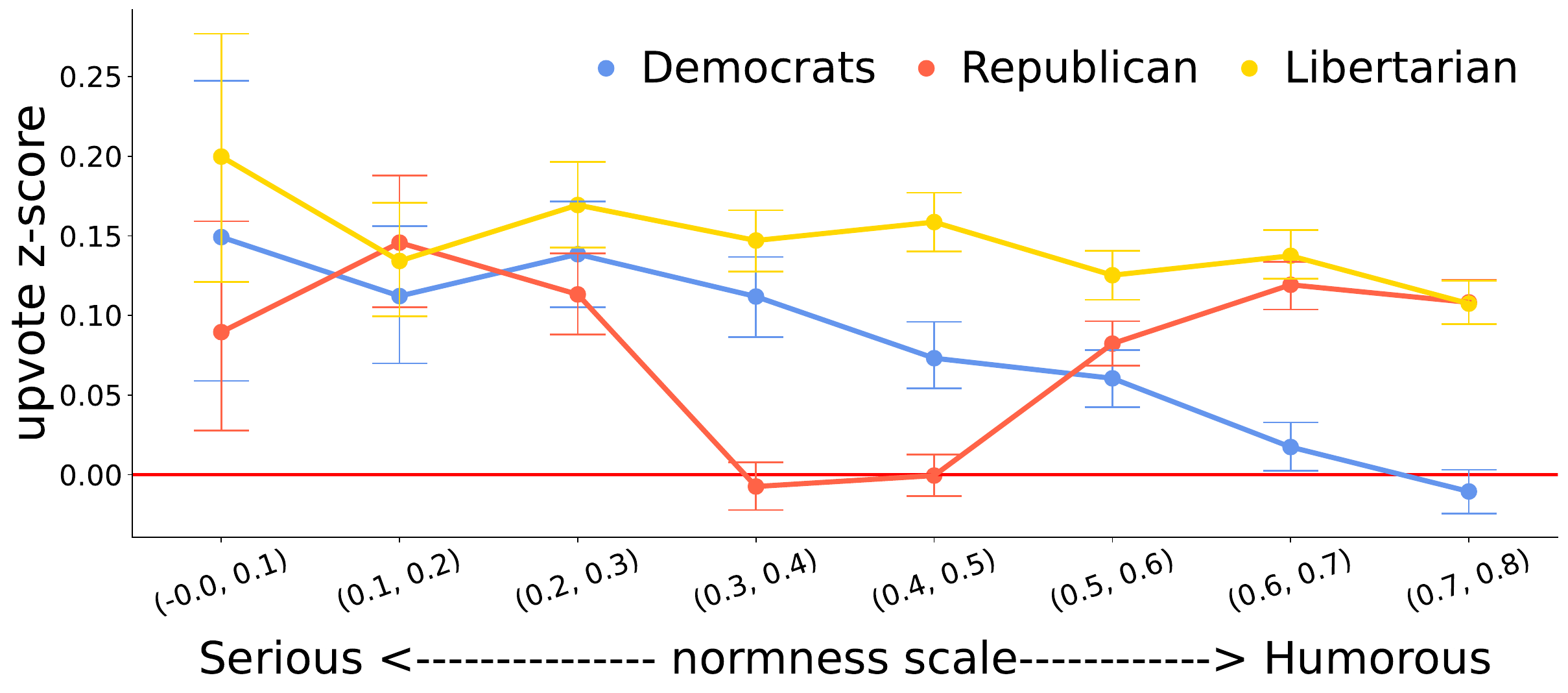}
    \caption{RPM plots for politics subreddits on the humor dimension.}
    \label{fig:rpm-politics-humor-original}
\end{figure}

\begin{figure}[h!]
    \centering
    \includegraphics[width=\columnwidth]{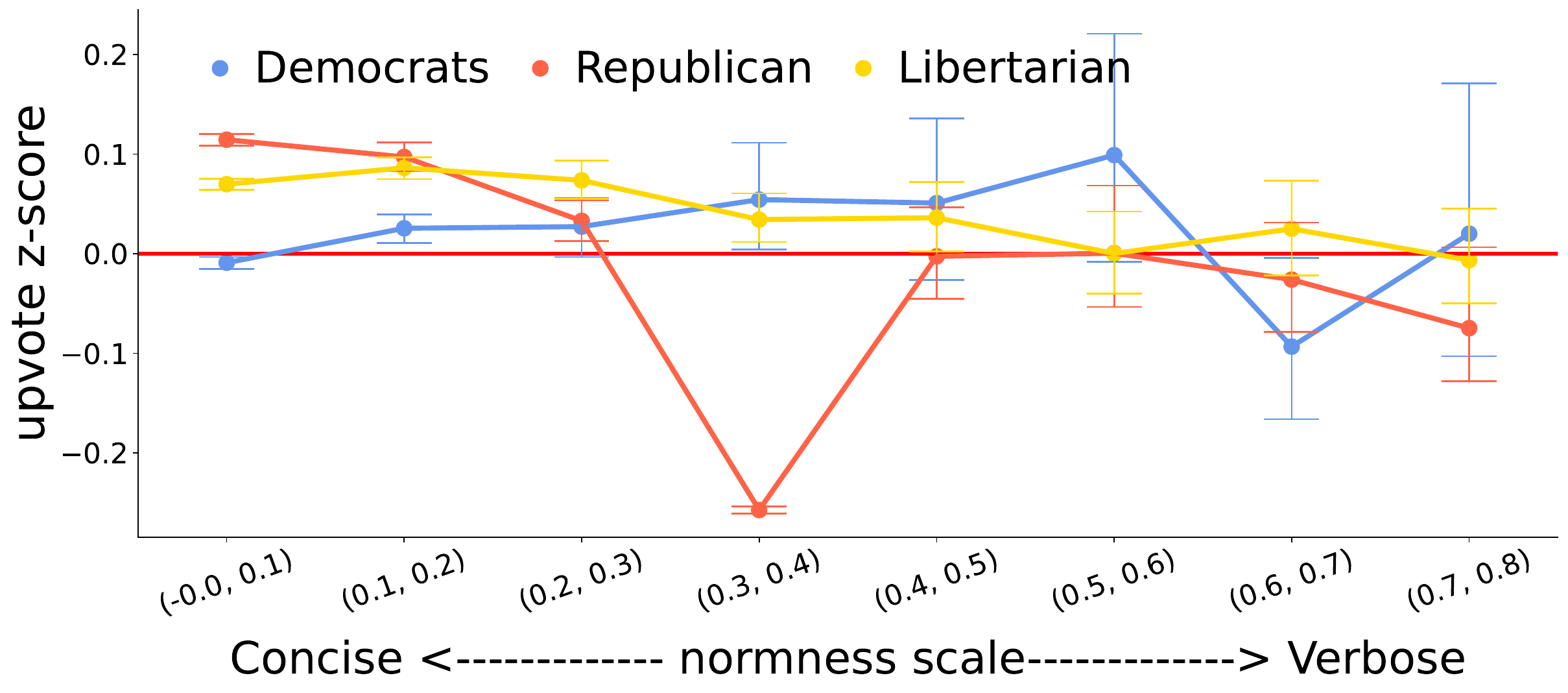}
    \caption{RPM plots for politics subreddits on the verbosity dimension.}
    \label{fig:rpm-politics-verbosity-original}
\end{figure}

\subsection{Science Subreddits}
\begin{figure}[h!]
    \centering
    \includegraphics[width=\columnwidth]{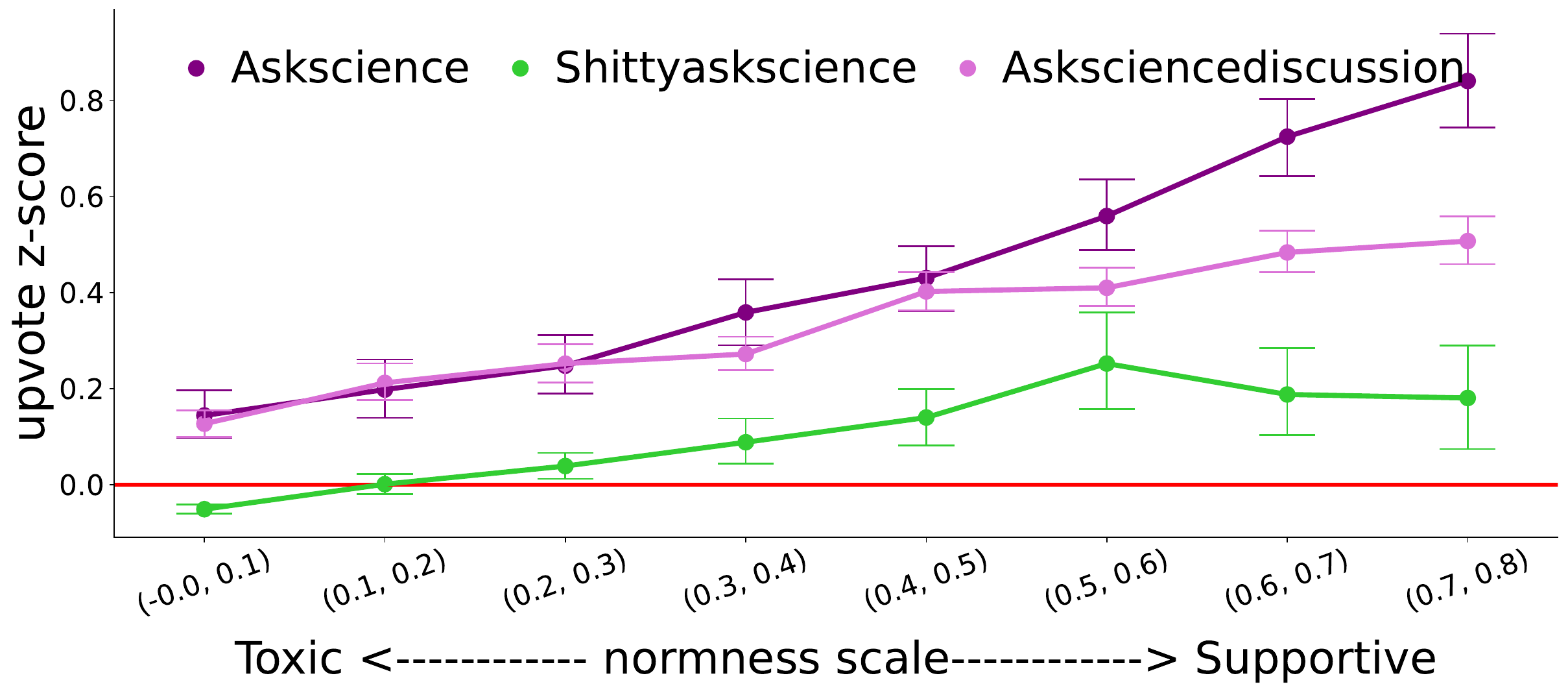}
    \caption{RPM plots for science subreddits on the supportiveness dimension.}
    \label{fig:rpm-science-supportiveness-original}
\end{figure}

\begin{figure}
    \centering
    \includegraphics[width=\columnwidth]{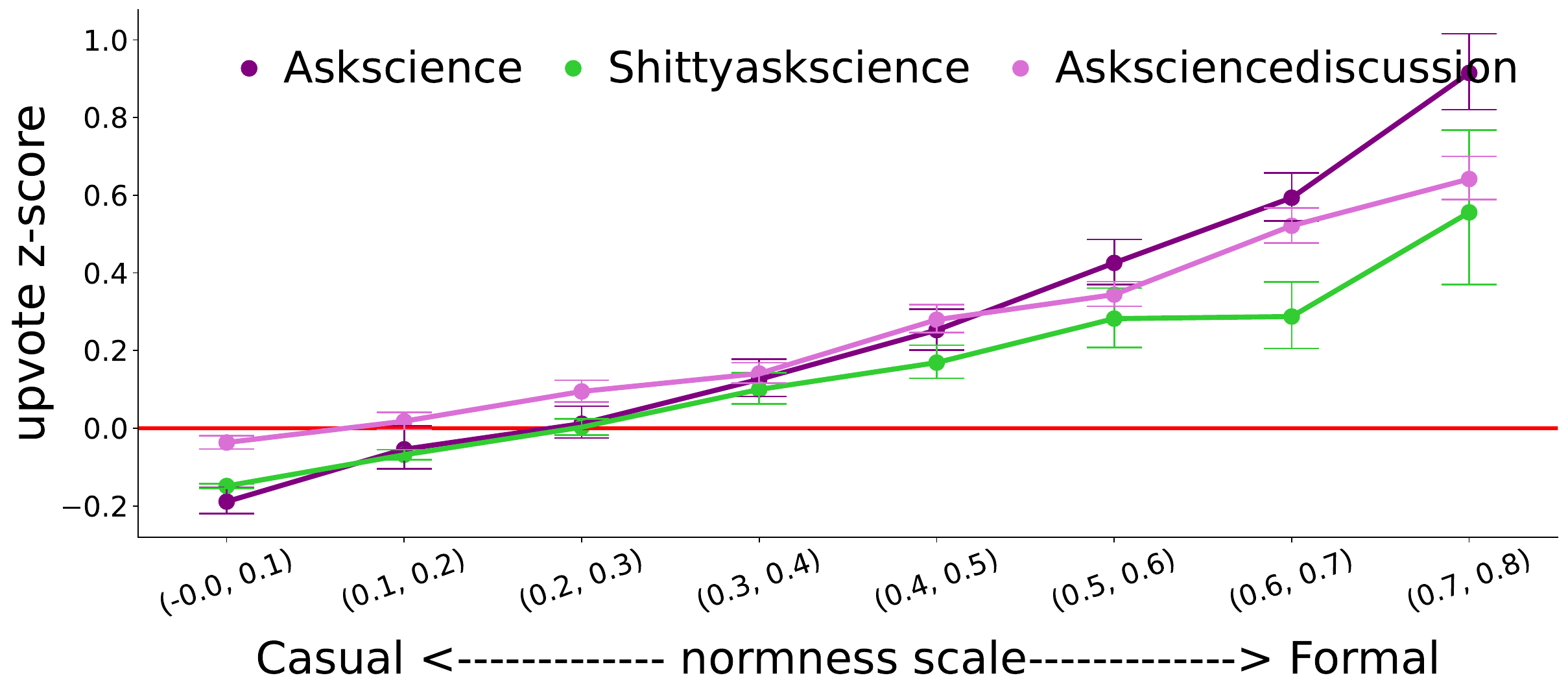}
    \caption{RPM plots for science subreddits on the formality dimension.}
    \label{fig:rpm-science-formality-original}
\end{figure}

\begin{figure}
    \centering
    \includegraphics[width=\columnwidth]{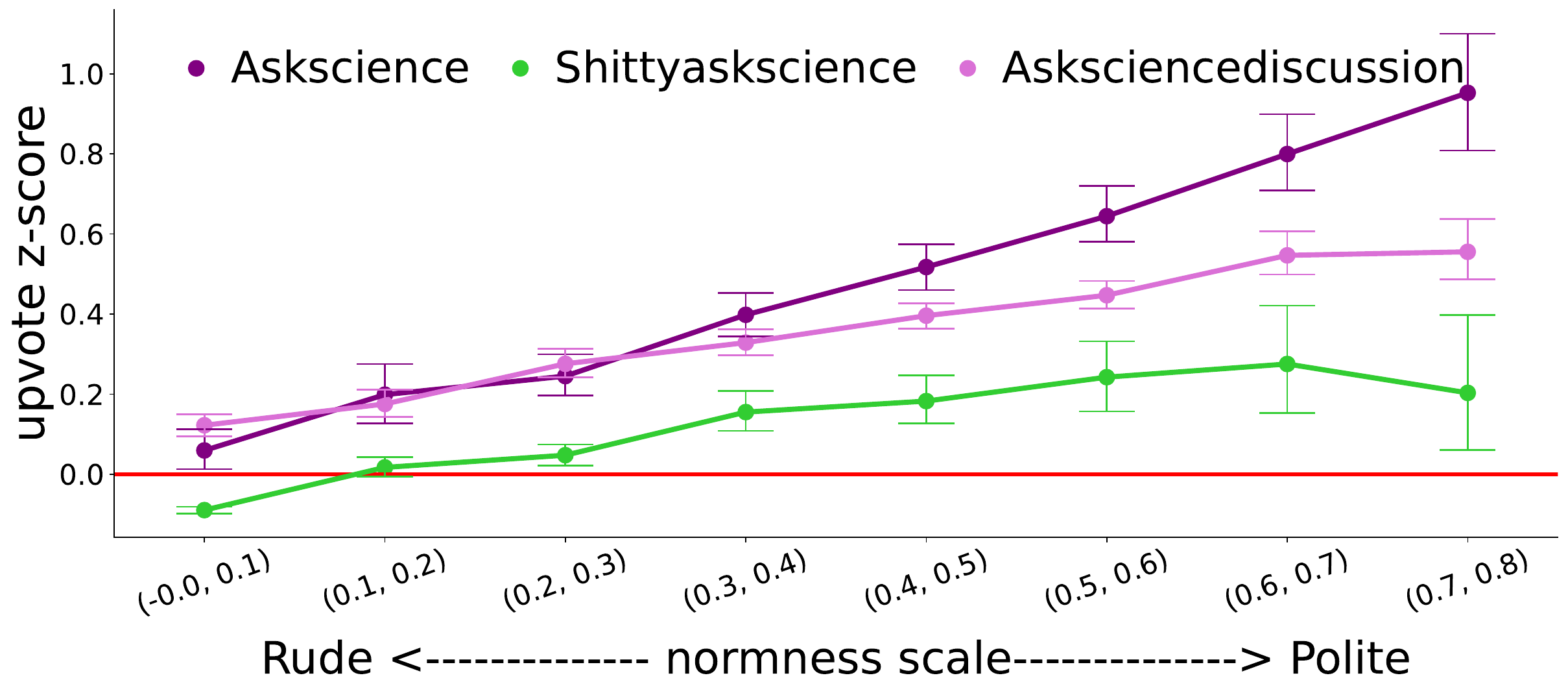}
    \caption{RPM plots for science subreddits on the politeness dimension.}
    \label{fig:rpm-science-politeness-original}
\end{figure}

\begin{figure}
    \centering
    \includegraphics[width=\columnwidth]{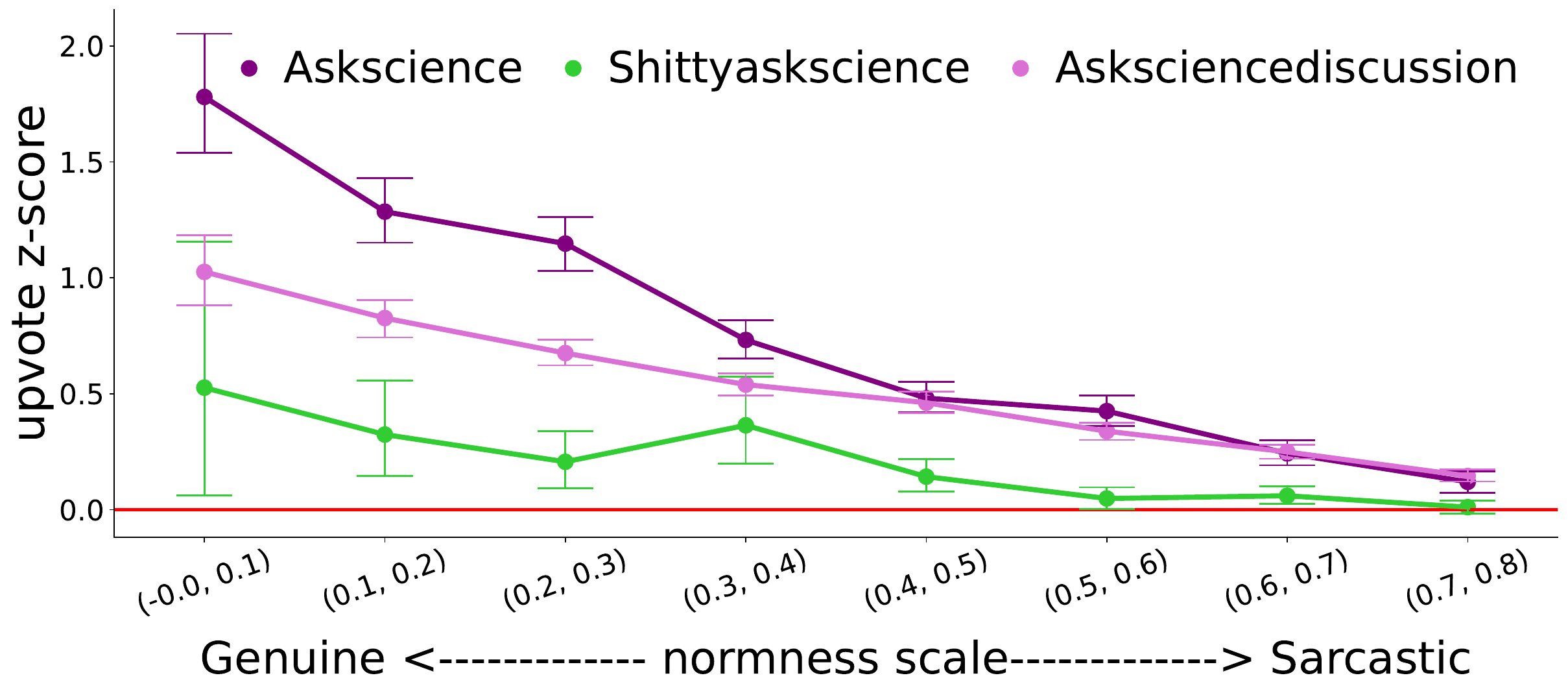}
    \caption{RPM plots for science subreddits on the sarcasm dimension.}
    \label{fig:rpm-science-sarcasm-original}
\end{figure}

\begin{figure}
    \centering
    \includegraphics[width=\columnwidth]{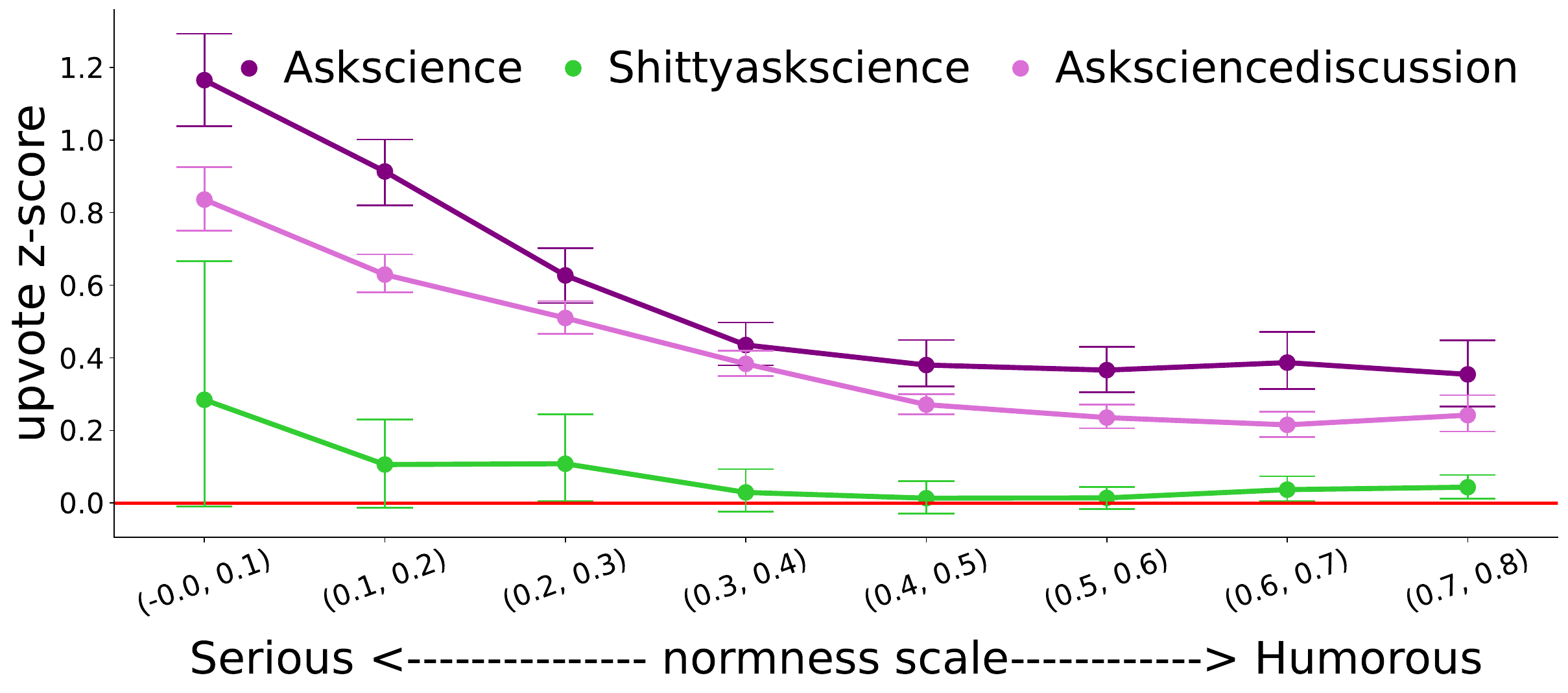}
    \caption{RPM plots for science subreddits on the humor dimension.}
    \label{fig:rpm-science-humor-original}
\end{figure}

\begin{figure}
    \centering
    \includegraphics[width=\columnwidth]{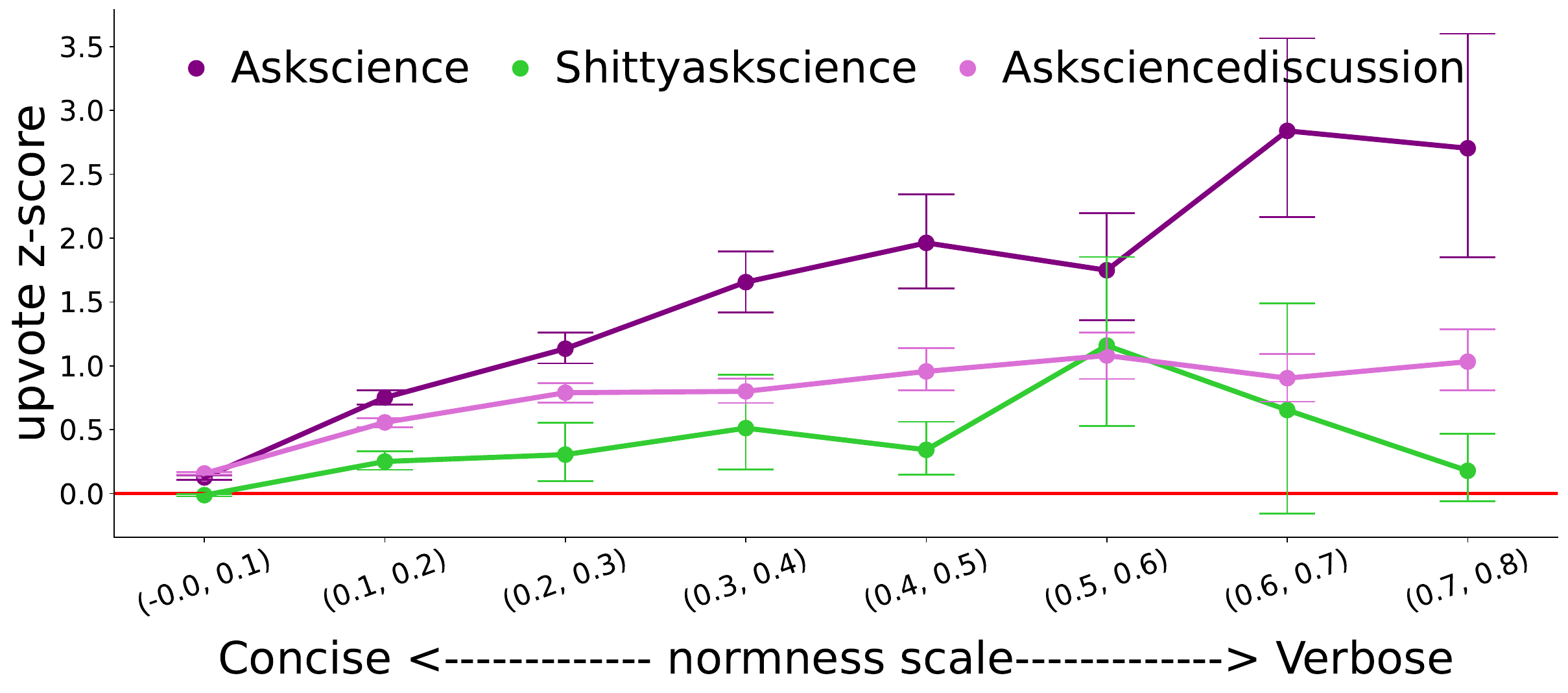}
    \caption{RPM plots for science subreddits on the verbosity dimension.}
    \label{fig:rpm-science-verbosity-original}
\end{figure}

\end{document}